\documentclass{article}

\usepackage{bbm}
\usepackage{makecell}
\usepackage{multirow}
\usepackage{pifont}
\newcommand{\cmark}{\ding{51}}
\newcommand{\xmark}{\ding{55}}
\usepackage{xcolor}
\usepackage[table]{xcolor}

\usepackage{tikz}
\usetikzlibrary{patterns}

\pgfdeclarepatternformonly{diagonal stripes nw}
{\pgfpoint{-3pt}{-3pt}}{\pgfpoint{17pt}{17pt}}
{\pgfpoint{14pt}{14pt}}%
{
  \pgfsetlinewidth{5pt}
  \pgfpathmoveto{\pgfpoint{-3pt}{17pt}}
  \pgfpathlineto{\pgfpoint{17pt}{-3pt}}
  \pgfusepath{stroke}
}

\definecolor{darcyflow}{RGB}{52, 138, 189}
\definecolor{helmholtz}{RGB}{251, 193, 94}
\definecolor{navierstokes}{RGB}{142, 186, 66}
\definecolor{poisson}{RGB}{226, 74, 51}
\definecolor{background}{RGB}{217, 217, 255}

\usepackage{microtype}
\usepackage{graphicx}
\usepackage{subcaption}
\usepackage{booktabs} 

\usepackage{hyperref}

\usepackage[preprint]{icml2026}

\usepackage{amsmath}
\usepackage{amssymb}
\usepackage{mathtools}
\usepackage{amsthm}

\usepackage[capitalize,noabbrev]{cleveref}

\theoremstyle{plain}

\theoremstyle{definition}

\theoremstyle{remark}

\usepackage[textsize=tiny]{todonotes}

\icmltitlerunning{Ambient Physics}

\begin{document}

\twocolumn[
  \icmltitle{Ambient Physics: Training Neural PDE Solvers with Partial Observations}



  \icmlsetsymbol{equal}{*}

  \begin{icmlauthorlist}
    \icmlauthor{Harris Abdul Majid}{equal,UoE}
    \icmlauthor{Giannis Daras}{MIT}
    \icmlauthor{Francesco Tudisco}{UoE}
    \icmlauthor{Steven McDonagh}{UoE}
  \end{icmlauthorlist}

  \icmlaffiliation{UoE}{The University of Edinburgh, Edinburgh, Scotland}
  \icmlaffiliation{MIT}{Massachusetts Institute of Technology, Cambridge, Massachusetts, United States}

  \icmlcorrespondingauthor{Harris Abdul Majid}{h.abdulmajid@ed.ac.uk}

  \icmlkeywords{Machine Learning, ICML}

  \vskip 0.3in
]



\printAffiliationsAndNotice{}  

\begin{abstract}
    In many scientific settings, acquiring complete observations of PDE coefficients and solutions can be expensive, hazardous, or impossible. Recent diffusion-based methods can reconstruct fields given partial observations, but require complete observations for training. We introduce Ambient Physics, a framework for learning the joint distribution of coefficient-solution pairs directly from partial observations, without requiring a single complete observation. The key idea is to randomly mask a subset of already-observed measurements and supervise on them, so the model cannot distinguish ``truly unobserved" from ``artificially unobserved", and must produce plausible predictions everywhere. Ambient Physics achieves state-of-the-art reconstruction performance. Compared with prior diffusion-based methods, it achieves a 62.51\% reduction in average overall error while using 125\(\times\) fewer function evaluations. We also identify a ``one-point transition'': masking a single already-observed point enables learning from partial observations across architectures and measurement patterns. Ambient Physics thus enables scientific progress in settings where complete observations are unavailable.
\end{abstract}

\section{Introduction}
\label{introduction}

Partial differential equations are fundamental to science and engineering \cite{evans2022partial, higham2015princeton}. They are mathematical models describing some physical phenomena, in which inputs are coefficients and outputs are solutions, with a wide range of applications from atmospheric modeling to automotive optimization. Problems are often categorized as either the \textit{forward problem}: to predict the solution from a given set of coefficients, or the \textit{inverse problem}: to make deductions about the coefficients from a given solution. In practice, measurements of coefficients and solutions are often incomplete. Climate scientists approximate atmospheric states from sparse weather stations \cite{lynch2006emergence}, and geophysicists determine properties of the Earth's subsurface from limited borehole data \cite{aki2002quantitative}.

Partial observations present significant challenges. Classical numerical solvers require defined coefficients and boundary conditions, and inverse methods require regularization or strong priors; so predicting the solution from partial observations of the coefficients or making deductions about the coefficients from partial observations of the solution can elicit large uncertainties. To tackle this challenge, researchers have proposed learning-based methods to reconstruct partial observations into physically-plausible complete observations. In particular, founded on Diffusion Posterior Sampling (DPS) \cite{chung2022diffusion}, DiffusionPDE and FunDPS make use of diffusion models to learn the joint distribution of coefficient-solution pairs \(p(a, u)\), where \(a\) and \(u\) denote the coefficients and solution respectively, and then apply DPS to sample from the posterior \(p(a, u|A_a a, A_u u)\), where \(A_a\) and \(A_u\) denote measurement operators, and \(A_a a\) and \(A_u u\) are the corresponding partial observations \cite{huang2024diffusionpde, yao2025guided}.

\begin{figure*}[t]
    \centering
    \includegraphics[width=1.0\linewidth]{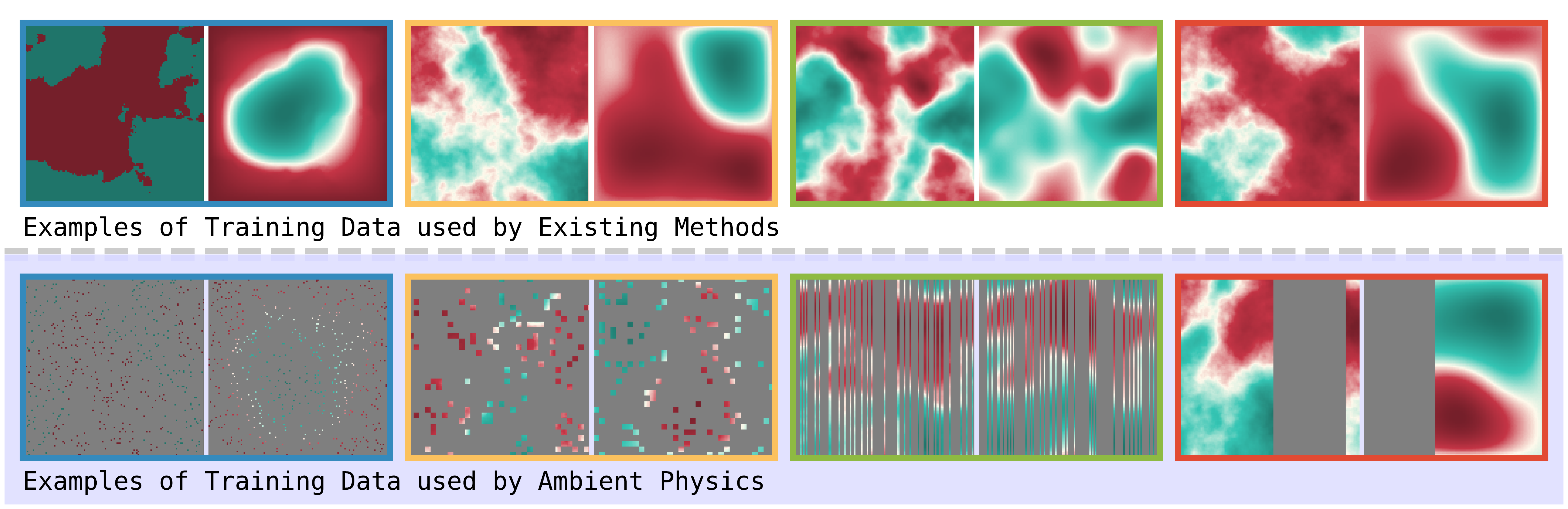}
    \caption{\textbf{Examples of Training Data used by Existing Methods vs Ambient Physics.} \textbf{Top:} Existing diffusion-based methods require access to complete observations to learn the joint distribution of coefficient-solution pairs. \textbf{Bottom:} \setlength{\fboxsep}{0pt}\colorbox{background!75}{Ambient Physics} learns the joint distribution directly from partial observations, without requiring a single complete observation. \textbf{Left to Right:} \setlength{\fboxsep}{0pt}\colorbox{darcyflow!50}{Darcy flow} with point observations, \setlength{\fboxsep}{0pt}\colorbox{helmholtz!50}{Helmholtz} with patch observations, \setlength{\fboxsep}{0pt}\colorbox{navierstokes!50}{Navier-Stokes} with slice observations, and \setlength{\fboxsep}{0pt}\colorbox{poisson!50}{Poisson} with window observations. Within each panel, the left field represents the coefficients (or initial condition) and the right field represents the solution (or final state).}
    \label{fig:plot_training_data_naive_vs_ambient}
\end{figure*}

Both DiffusionPDE and FunDPS exhibit exceptional reconstruction performance. With partial observations of only 3\%, these diffusion-based methods can sample physically plausible complete observations from the posterior distribution; and across diverse PDEs, achieve relative \(L_2\) errors in the range of 2--20\%. In contrast, more traditional neural PDE solvers (DeepONet \cite{lu2019deeponet}, FNO \cite{li2020fourier}, PINN \cite{raissi2019physics}, PINO \cite{li2024physics}) struggle with partial observations and achieve errors that are often an order of magnitude larger. Still, there is a fundamental motivational misalignment: DiffusionPDE and FunDPS require complete observations to learn the joint distribution of coefficient-solution pairs, having been pre-trained on 50,000 complete observations. For many real-world problems---especially those that would most benefit from reconstruction---acquiring 50,000 complete measurements is infeasible. Naive training directly on partial observations fails catastrophically (see \autoref{fig:plot_predictions_naive_vs_ambient}). In that setting, a diffusion model can learn only the distribution of partial observations \(p(A_a a, A_u u)\), predicting accurately only at observed locations and arbitrarily elsewhere. 

To address the motivational misalignment described above, we propose Ambient Physics, a framework for learning the joint distribution of coefficient-solution pairs directly from partial observations. The key idea is to randomly mask a subset of already-observed measurements and supervise on them, so that the model cannot distinguish ``truly unobserved" from ``artificially unobserved" and must produce physically plausible predictions everywhere to minimize error. Concretely, our contributions are as follows:
\begin{itemize}
    \item We formulate the \textit{reconstruction problem}, which treats both the classical forward and inverse problems as special cases.
    \item We show that naive training on partial observations fails catastrophically: models predict accurately only at observed locations and arbitrarily elsewhere.
    \item We demonstrate that, across Darcy flow, Helmholtz, Navier-Stokes, and Poisson, Ambient Physics achieves a 62.51\% reduction in average overall error while using 125\(\times\) fewer function evaluations., compared with prior diffusion-based methods. We emphasize that Ambient Physics is trained entirely on partial observations, whereas prior methods require complete observations.
    \item We demonstrate that Ambient Physics is not tied to diffusion or flow matching by adapting traditional neural PDE solvers to learn directly from partial observations.
    \item We identify a ``one-point transition": masking a single already-observed point enables learning from partial observations across architectures and measurement patterns. We also evaluate robustness under different observation patterns, and evaluate super-resolution without any high-resolution observations.
\end{itemize}

\section{Background}
\label{background}

\paragraph{Partial Differential Equations.}
We consider partial differential equations of the form: 
\begin{align*}
    \mathcal{L}_a u &=0, \quad \text{ on } \Omega, \\
    u_{|\partial \Omega} &= g,
\end{align*}
where \(\mathcal{L}_a\) represents a differential operator, \(a: \Omega \rightarrow \mathbb{R}^n\) represents the coefficients, \(u: \Omega \rightarrow \mathbb{R}^n\) represents the solution, and \(g: \partial \Omega \rightarrow \mathbb{R}^{n_b}\) represents the boundary conditions. For time-dependent PDEs, we may view \(a\) as the initial condition \(u(0)\) and \(u\) as the final state \(u(T)\) at time \(T\).

The forward problem is to map a given set of coefficients (or initial conditions) \(a\) to the solution \(u\), while the inverse problem is to map the solution \(u\) to a given set of coefficients \(a\). These problems induce corresponding operators: the forward operator \(\mathcal{M}_a: a \rightarrow u\), and the inverse operator \(\mathcal{M}_u: u \rightarrow a\).

\paragraph{Neural PDE Solvers.}
A neural PDE solver approximates the forward or inverse operator \(\mathcal{M}\) with a learned model \(\mathcal{M}^\theta\). These methods are often trained by minimizing the data loss
\[\lVert x_\text{true} - x_\text{pred} \rVert_2,\]
where \(x\) denotes the coefficients \(a\) or solution \(u\), using data from high-fidelity numerical solvers or real-world observations, and sometimes the PDE loss which penalizes violations of the governing equation, or a combination of both. However, in many cases, incorporating the PDE loss only yields marginal improvements \cite{krishnapriyan2021characterizing, wang2022and}, and with incomplete inputs, the required derivatives become difficult to compute \cite{huang2024diffusionpde}.

Neural PDE solvers are evaluated by measuring the error between ground truth and predictions. In practice, specific requirements of the problem or task may dictate the choice of the metric. For example, climate models may prioritize long-term accuracy and stability over pointwise accuracy, while engineering designs may prioritize maximum tolerances in critical regions over average performance. Following prior work, we report the \textit{relative \(L_2\) error}
\[\frac{\lVert x_\text{true} - x_\text{pred}\rVert_2}{\lVert x_\text{true}\rVert_2}\]
for continuous-valued fields, which is particularly useful for comparing performance across fields with different magnitudes and across different PDEs. For discrete-valued fields, we report the \textit{misclassification rate}
\[\frac{1}{N}\sum_{i=1}^{N} \mathbbm{1} \left[x_\text{true}^{(i)} \neq x_\text{pred}^{(i)} \right].\]

Neural PDE solvers are commonly grouped into two categories: physics-informed methods and operator-learning methods. Physics-Informed Neural Networks (PINNs) incorporate physical laws, typically described by partial differential equations, into the loss functions to guide training \cite{raissi2019physics}. Several extensions improve parallelization and scalability: conservative PINNs (cPINNs) apply separate neural networks on subdomains and enforce flux continuity at subdomain interfaces \cite{jagtap2020conservative}; eXtended PINNs (XPINNs) generalize this domain decomposition approach to arbitrary domains, geometries, and PDE types \cite{jagtap2020extended}. In contrast, neural operators learn mappings between infinite-dimensional function spaces \cite{kovachki2023neural}. Deep Operator Networks (DeepONets) use two sub-networks that combine via a dot product to produce predictions at queried locations \cite{lu2019deeponet}. Fourier Neural Operators (FNOs) parameterize the solution operator via learnable integral kernels in Fourier space \cite{li2020fourier}. Many extensions improve expressivity and transferability: Factorized-FNO (F-FNO) uses separable spectral layers and modern residual connections and training strategies \cite{tran2021factorized}; Geometry-aware FNO (Geo-FNO) uses message passing to adapt operator-learning to irregular geometries \cite{li2023fourier}. Physics-Informed Neural Operator (PINO) incorporates PDE residuals into the loss function to bridge the gap between physics-informed and operator-learning methods \cite{li2024physics}.
%
These methods have been applied across a wide range of domains, including heat transfer \cite{zobeiry2021physics, cai2021physics, lu2024surrogate, cao2024laplace}, electromagnetic fields \cite{baldan2023physics, piao2024domain, peng2022rapid, augenstein2023neural}, and fluid mechanics \cite{jin2021nsfnets, eivazi2022physics, luo2023cfdbench, azizzadenesheli2024neural}.

\paragraph{Reconstruction Under Partial Observations.}
Several works study reconstruction under partial observations, particularly in fluid mechanics where real-world data are often collected by a limited number of probes, yielding observations only at sparse locations or along a small number of slices, and sometimes at low-resolution. For instance, transformer-based architectures have been used to reconstruct two- and three-dimensional flows from sparse locations \cite{zhang2025operator}, and convolutional neural networks have been used to reconstruct three-dimensional flows from two-dimensional cross-sectional slices \cite{matsuo2024reconstructing}. More recently, diffusion models have been adopted for this problem because they generate high-quality samples and support uncertainty quantification. A number of works modify diffusion training or sampling 
by incorporating PDE residuals, and demonstrate applications to field reconstruction and super-resolution tasks for two- and three-dimensional flows \cite{yousif2021high, shu2023physics, jacobsen2025cocogen, amoros2026guiding}. DiffusionPDE~\cite{huang2024diffusionpde} learns the joint distribution of coefficient-solution pairs \(p(a, u)\) and then applies Diffusion Posterior Sampling \cite{chung2022diffusion} to guide the denoising process using partial observations and PDE residuals, while FunDPS \cite{yao2025guided} extends this framework to function-space diffusion models. Importantly, \textbf{while these works are motivated by scientific settings in which complete observations are unavailable, the experimental setups exclusively use complete observations for training and validation}, with datasets ranging from 1,000--50,000 complete observations. This motivates methods that can learn directly from partial observations.

\paragraph{Diffusion Models for Inverse Problems.} 
Diffusion models \cite{sohl2015deep, ho2020denoising} are effective generative priors for inverse problems, enabling reconstruction by generating samples that are consistent with a learned prior and the available observations. This is often done by training an unconditional diffusion model on complete observations and then performing approximate posterior sampling by guiding the denoising process with a measurement-consistency objective, optionally augmented with additional constraints. Examples include DDRM for linear inverse problems \cite{kawar2022denoising} and DPS for more general inverse problems \cite{chung2022diffusion}. However, these methods assume access to fully observed training samples. In contrast, ``ambient'' methods aim to learn a generative model of the underlying data distribution given only incomplete or noisy measurements.
AmbientGAN shows that learning the target distribution from partially observed training samples is possible \cite{bora2018ambientgan}, and Ambient Diffusion extends this perspective to diffusion models \cite{daras2023ambient}. This enables learning generative priors directly from partial observations, without requiring a single complete observation.

\section{Method}
\label{method}

\paragraph{The Reconstruction Problem.} 
The PDE literature typically distinguishes between the \textit{classical forward problem} \(a \rightarrow u\) and the \textit{classical inverse problem} \(u \rightarrow a\). In this work, we propose a generalized formulation that treats both as special cases and explicitly addresses real-world partial observations. 

Let \(A_a: \mathbb{R}^n \rightarrow \mathbb{R}^n\) and \(A_u: \mathbb{R}^n \rightarrow \mathbb{R}^n\) represent (possibly non-invertible) measurement operators (e.g., binary masks) that act on coefficients and solutions, respectively. Given partial observations \((A_a a, A_u u)\), the \textit{reconstruction problem} is to predict a coefficient-solution pair \((a, u)\) that agrees with the partial observations and is physically-plausible. This perspective is closely analogous to image inpainting with multiple channels and per-channel masks, as well as many other tasks in computational imaging (e.g., X-ray computed tomography).

This generalized formulation treats the classical forward and inverse problems as special cases. Let \(\mathbf{I}\) denote the identity operator and \(\mathbf{0}\) denote the null operator. When \(A_a = \mathbf{I}\) and \(A_u = \mathbf{0}\), the reconstruction problem reduces to the classical forward problem:
\[(\mathbf{I}a, \mathbf{0}u) = (a, \emptyset) \rightarrow (a, u).\]
Conversely, when \(A_a = \mathbf{0}\) and \(A_u = \mathbf{I}\), the reconstruction problem reduces to the classical inverse problem:
\[(\mathbf{0}a, \mathbf{I}u) = (\emptyset, u) \rightarrow (a, u).\]
It also captures settings not covered by the classical dichotomy. For example, when \(A_a \notin \{\mathbf{0}, \mathbf{I}\}\) and \(A_u = \mathbf{0}\), the reconstruction problem becomes \textit{coefficient inpainting}, where the goal is to predict the coefficients given partial coefficients. More generally, both the coefficients and solution may be incomplete, i.e., \(A_a, A_u \notin \{\mathbf{0}, \mathbf{I}\}\). All of these cases fall under our proposed generalized formulation. In the remainder of this work, we use ``reconstruction problem'' to refer to this unified task of predicting coefficient-solution pairs \((a, u)\) given partial observations \((A_a a, A_u u)\), including the classical forward and inverse problems as special cases. 

\paragraph{Objective.}
Our objective is to solve the reconstruction problem by learning the joint distribution of coefficient-solution pairs \(p(a, u)\), and then, at inference, sampling from the posterior \(p(a, u \mid A_a a, A_u u)\). Prior diffusion-based methods adopt this strategy, but they require access to complete observations \((a, u)\) during training. In many scientific and engineering settings, however, only partial observations \((A_a a, A_u u)\) are available. This makes learning the joint distribution strictly more challenging: we must learn \(p(a, u)\) directly from partial observations \((A_a a, A_u u)\).

\paragraph{Flow Matching.}
In this work, we adopt flow matching as our generative modeling framework \cite{lipman2022flow}. Flow matching learns a transport from a simple base distribution (e.g., Gaussian noise \(x_0 \sim \mathcal{N}(0, I)\)) to the data distribution (e.g., \(x_1 \sim p_{\text{data}}\)) by sampling an interpolation between noise \(x_0\) and data \(x_1\) and training a neural network to predict the corresponding flow field. Flow matching also supports conditional generation (e.g., posterior sampling).

We use rectified flow \cite{liu2022flow}, which has a particularly simple training and sampling procedure and typically performs competitively with more complex score-based diffusion models. While common practice for rectified flow is to parameterize the network to predict the instantaneous velocity, we parameterize our network to predict the final state \(x_1\).

\paragraph{Naive Training.}
With only partial observations available, a natural baseline is to train a conditional flow matching model directly on the available measurements.

Let \(\epsilon_a, \epsilon_u \sim \mathcal{N}(0, I)\) denote independent Gaussian noise and let \(t \sim \mathcal{U}([0,1))\) denote the flow time.
Following rectified flow, one forms \textit{mid-flow states} (i.e., noised partial observations) by linearly interpolating between noise and the measurements:
\begin{align*}
    a_t &= t \cdot (A_a a) + (1 - t) \cdot \epsilon_a,\\
    u_t &= t \cdot (A_u u) + (1 - t) \cdot \epsilon_u.
\end{align*}
Then, one can train a conditional model \(\mathcal{M}^\theta\) to predict the coefficient-solution pair \((a, u)\) from the mid-flow states \((a_t, u_t)\) while conditioning on partial observations \((A_a a, A_u u)\), masks \((A_a, A_u)\), and time \(t\):
\[
\mathcal{M}^\theta(A_a a_t, A_u u_t, A_a a, A_u u, A_a, A_u, t) = (\hat{a}, \hat{u}).
\]
Conditioning on partial observations \((A_a a, A_u u)\) is necessary so that, at inference, generated fields depend on, and agree with, the available measurements; removing it would yield unconditional samples that are not constrained to match partial observations \((A_a a, A_u u)\). Since only partial observations \((A_a a, A_u u)\) are available for training, supervision is restricted to the available measurements:
\[
\mathcal{L}_{\text{meas}} = \lVert A_a a - A_a \hat{a} \rVert_2 + \lVert A_u u - A_u \hat{u} \rVert_2.
\]
This naive training procedure leads to catastrophic failure: the model predicts accurately only at the available measurements and arbitrarily elsewhere (\autoref{fig:plot_predictions_naive_vs_ambient}). Since the loss \(\mathcal{L}_{\text{meas}}\) penalizes errors only at observed locations while the full partial observations \((A_a a, A_u u)\) are also provided as conditioning, the loss can be minimized by simply copying the observed values into \((\hat a, \hat u)\), yielding near-perfect predictions at observed locations. Everywhere else, however, the model receives no training signal. Predictions at unobserved locations do not affect \(\mathcal{L}_{\text{meas}}\), so the model can make arbitrary predictions.

\paragraph{Ambient Physics Training.}
We propose Ambient Physics training to learn the joint distribution of coefficient-solution pairs directly from partial observations $(A_a a, A_u u)$. Naive training provides no training signal at unobserved locations, so to obtain physically plausible predictions everywhere, additional training signal is needed at unobserved locations. The key idea is to randomly mask a subset of already-observed measurements and supervise on these artificially masked measurements. As a result, the model cannot distinguish ``truly unobserved'' from ``artificially unobserved'' and must produce physically plausible predictions across the entire domain to minimize error.

Let \(B_a: \mathbb{R}^n \rightarrow \mathbb{R}^n\) and \(B_u: \mathbb{R}^n \rightarrow \mathbb{R}^n\) denote additional masking operators that further mask the partial observations. In particular, \(B_a\) and \(B_u\) are chosen so that they mask only a subset of already-observed measurements (i.e., they zero-out entries that are observed under \(A_a\) and \(A_u\), and never reveal new ones). Consequently, the inputs \((B_a A_a a,\, B_u A_u u)\) contain strictly less information than the training observations \((A_a a,\, A_u u)\).
For notational convenience, we define \(\tilde B_x := B_x A_x\).

Let \(\epsilon_a, \epsilon_u \sim \mathcal{N}(0, I)\) denote independent Gaussian noise and let \(t \sim \mathcal{U}([0,1))\) denote the flow time.
Following rectified flow, we form mid-flow states by linearly interpolating between noise and the masked partial observations:
\begin{align*}
    a_t &= t \cdot (\tilde B_a a) + (1 - t) \cdot \epsilon_a,\\
    u_t &= t \cdot (\tilde B_u u) + (1 - t) \cdot \epsilon_u.
\end{align*}
Then, we train a conditional model \(\mathcal{M}^\theta\) to predict the clean fields \((a,u)\) from \((a_t, u_t)\) while conditioning on the masked partial observations \((\tilde B_a a, \tilde B_u u)\), masks \((\tilde B_a, \tilde B_u)\), and time \(t\):
\[
\mathcal{M}^\theta(\tilde B_a a_t, \tilde B_u u_t, \tilde B_a a, \tilde B_u u, \tilde B_a, \tilde B_u, t) = (\hat{a}, \hat{u}).
\]
Critically, the model is given strictly less information than the training observations, but it is still supervised on all available measurements \((A_a a, A_u u)\):
\[
\mathcal{L}_{\text{meas}} = \lVert A_a a - A_a \hat{a} \rVert_2 + \lVert A_u u - A_u \hat{u} \rVert_2.
\]
In contrast to naive training, which can minimize \(\mathcal{L}_{\text{meas}}\) by simply copying the observed values and predicting arbitrarily elsewhere, Ambient Physics withholds a random subset of the available measurements from the conditioning using \((B_a, B_u)\). The model is still supervised on the full set of available measurements \((A_a a, A_u u)\) through \(\mathcal{L}_{\text{meas}}\), so it must predict the held-out (``artificially unobserved'') measurements from the remaining context. Since the model cannot distinguish truly unobserved points from artificially unobserved ones, the only way to minimize \(\mathcal{L}_{\text{meas}}\) is to learn the joint distribution of coefficient-solution pairs, enabling posterior sampling of fields that are physically plausible across the entire domain.

\paragraph{Ambient Physics Sampling.}
During training, the Ambient Physics model receives as input the masked mid-flow states $(\tilde B_a a_t, \tilde B_u u_t)$ and conditions on the masked partial observations $(\tilde B_a a, \tilde B_u u)$, the corresponding masks $(\tilde B_a, \tilde B_u)$, and time $t$. At inference, we are given partial observations $(A_a a, A_u u)$ and our goal is to sample from the posterior $p(a, u\mid A_a a, A_u u)$. Crucially, sampling must replicate the same masking and conditioning structure used during training.

We first sample Gaussian noise $a_0, u_0 \sim \mathcal{N}(0, I)$ and masking operators $(B_a, B_u)$ from the same distribution used during training, i.e., $B_a \sim p(B_a \mid A_a)$ and $B_u \sim p(B_u \mid A_u)$. We then numerically integrate the rectified-flow ODE from $t=0$ to $t \approx 1$.

To do so, we discretize time as $t_k = k/K$ for $k=0,\dots,K$, where $K$ is the number of sampling steps (function evaluations), and we maintain states $(a_k, u_k)$ that approximate $(a_{t_k}, u_{t_k})$.

At each step $k$, we predict the corresponding clean fields by applying the same masking/conditioning structure used during training:
\[
\mathcal{M}^\theta(\tilde B_a a_k, \tilde B_u u_k, \tilde B_a a, \tilde B_u u, \tilde B_a, \tilde B_u, t_k) = (\hat{a}_k, \hat{u}_k).
\]
We then compute the rectified-flow velocity at $t_k$ as
\begin{align*}
    v_a(t_k) &= \frac{\hat{a}_k - a_k}{(1 - t_k)},\\
    v_u(t_k) &= \frac{\hat{u}_k - u_k}{(1 - t_k)},
\end{align*}
and step forward with an ODE solver (e.g., Euler or Heun; we use Euler):
\begin{align*}
    a_{k+1} &= a_k + \frac{1}{K} \cdot v_a(t_k),\\
    u_{k+1} &= u_k + \frac{1}{K} \cdot v_u(t_k).
\end{align*}
The final state $(a_K, u_K)$ is returned as a reconstruction sample $(\hat a,\hat u)$.

Sampling the masking operators $(B_a, B_u)$ introduces an additional source of randomness beyond the Gaussian noise. This allows us to draw different samples from the conditional distributions $p(a, u \mid \tilde B_a a, \tilde B_u u)$ induced by different masking operators, and can be viewed as sampling from a mixture over masks. In particular,
$$ p(a, u \mid A_a a, A_u u) \approx \mathbb{E}_{B_a, B_u} \left[ p(a, u \mid \tilde B_a a, \tilde B_u u) \right].$$
This can improve robustness and provide a practical way to quantify uncertainty. In practice, we find that a single mask draw and a single reconstruction sample from $p(a, u \mid \tilde B_a a, \tilde B_u u)$ performs well enough.

\begin{table*}[t]
    \centering
    \resizebox{\textwidth}{!}{%
    \begin{tabular}{lcccccccccc}
        \toprule
         & \multirow{2}{*}{\makecell{Partial Obs.\\Training}} & \multirow{2}{*}{NFE} & \multicolumn{2}{c}{Darcy Flow} & \multicolumn{2}{c}{Helmholtz} & \multicolumn{2}{c}{Navier-Stokes} & \multicolumn{2}{c}{Poisson} \\
        \cline{4-5}\cline{6-7}\cline{8-9}\cline{10-11}
         & & & Coefficient & Solution & Coefficient & Solution & Coefficient & Solution & Coefficient & Solution \\
        \midrule
        DeepONet & \color{red}{\xmark} & $-$ & 41.1\% & 38.3\% & 132.8\% & 123.1\% & 97.2\% & 103.2\% & 105.8\% & 155.5\% \\
        FNO & \color{red}{\xmark} & $-$ & 49.3\% & 28.2\% & 218.2\% & 98.2\% & 96.0\% & 101.4\% & 232.7\% & 100.9\% \\
        PINN & \color{red}{\xmark} & $-$ & 59.7\% & 48.8\% & 160.0\% & 142.3\% & 146.8\% & 142.7\% & 130.0\% & 128.1\% \\
        PINO & \color{red}{\xmark} & $-$ & 49.2\% & 35.2\% & 216.9\% & 106.5\% & 96.0\% & 101.4\% & 231.9\% & 107.1\% \\
        \midrule
        DiffusionPDE & \color{red}{\xmark} & 2000 & 7.87\% & 6.07\% & 19.07\% & 12.64\% & 9.63\% & 3.78\% & 21.10\% & 4.88\% \\
        \multirow{2}{*}{FunDPS} & \multirow{2}{*}{\color{red}{\xmark}} & 200 & 6.78\% & 2.88\% & 20.07\% & 2.20\% & 9.87\% & 3.99\% & 24.04\% & 2.04\% \\
         &  & 500 & 5.18\% & 2.49\% & 17.16\% & 2.13\% & 8.48\% & 3.32\% & 20.47\% & 1.99\% \\
        \midrule
        \rowcolor{background!75} &  & 1 & 2.08\% & 0.76\% & 7.73\% & 0.47\% & 6.41\% & 0.98\% & 7.86\% & 0.48\% \\
        \rowcolor{background!75} &  & 4 & \textbf{2.07\%} & \textbf{0.74\%} & \textbf{7.67\% }& \textbf{0.44\%} & \textbf{6.38\%} & \textbf{0.96\%} & \textbf{7.81\%} & \textbf{0.45\%} \\
        \rowcolor{background!75} \multirow{-3}{*}{\textbf{Ambient Flow}} & \multirow{-3}{*}{\color{blue}{\cmark}} & 16 & \textbf{2.07\%} & \textbf{0.74\%} & \textbf{7.67\%} & \textbf{0.44\%} & \textbf{6.38\%} & \textbf{0.96\%} & \textbf{7.81\%} & \textbf{0.45\%} \\
        \bottomrule
    \end{tabular}
    }
    \caption{\textbf{Reconstruction Performance Under Partial Observations.} Relative $L_2$ error (\%) between ground truth and reconstructed coefficients ($\hat{a}$) and solutions ($\hat{u}$) from partial observations $(A_a a, A_u u)$ with 3\% uniformly random measurements. \textit{Note: In prior work these are reported as ``Inverse" for coefficients and ``Forward" for solutions.} ``Partial Obs. Training" indicates whether training uses partial observations ({\color{blue}{\cmark}}) or complete observations ({\color{red}{\xmark}}); ``NFE" is the number of function evaluations. Across all PDEs and both fields, \setlength{\fboxsep}{0pt}\colorbox{background!75}{Ambient Flow} achieves the lowest relative $L_2$ error despite training only on partial observations and sampling with just 1--16 NFE.}
    \label{tab:1}
\end{table*}

\section{Results}
\label{results}

\paragraph{Reconstruction Under Partial Observations.} 
On the reconstruction problem, we compare traditional neural PDE solvers and prior state-of-the-art diffusion-based methods against Ambient Flow---our instantiation of Ambient Physics. The fundamental question is \textit{whether Ambient Physics can learn the joint distribution of coefficient-solution pairs directly from partial observations.}

In this experimental setup, we are given partial observations \((A_a a, A_u u)\), where \(A_a\) and \(A_u\) denote independently sampled binary masks that retain only 3\% uniformly random measurements. The goal is to predict a coefficient-solution pair $(a, u)$ that agrees with the partial observations and is physically plausible. 

We consider four diverse PDEs: Darcy flow, Helmholtz, Navier-Stokes, and Poisson. Following prior work, reconstruction performance is measured using the relative \(L_2\) error between ground truth and prediction (reported as a percentage), and computed over the entire field for each coefficient/solution and each test example, and then averaged across all examples.

We compare against traditional neural PDE solvers (DeepONet \cite{lu2019deeponet}, FNO \cite{li2020fourier}, PINN \cite{raissi2019physics}, PINO \cite{li2024physics}), which learn directional mappings and therefore require two separate models per PDE. We also include diffusion-based methods (DiffusionPDE \cite{huang2024diffusionpde}, FunDPS \cite{yao2025guided}), which learn the joint distribution of coefficient-solution pairs and then perform posterior sampling using the available partial observations. Importantly, all baselines require access to complete observations to learn either the directional mappings or the joint distribution. In contrast, Ambient Flow does not require a single complete observation and can learn the joint distribution directly from partial observations.

In Ambient Flow, we follow the Ambient Physics procedure from \autoref{method}. Starting from partial observations \((A_a a, A_u u)\), we sample additional masks \((B_a, B_u)\) that withhold one-third of the observed points. Thus, the model is conditioned on \((B_a A_a a,\, B_u A_u u)\), which retain only 2\% uniformly random measurements overall, while we supervise on all available measurements \((A_a a, A_u u)\). We use the UNet of \citet{ho2020denoising} and sample with an Euler scheme. Implementation details are provided in \autoref{implementation_details}.

\autoref{tab:1} summarizes reconstruction performance across PDEs and fields. Traditional neural PDE solvers perform poorly, with relative \(L_2\) errors ranging from 28.2\% to 232.7\%---as expected, since these solvers are not designed for reconstruction, particularly under extreme sparsity (3\% uniformly random measurements). Diffusion-based methods perform substantially better, with errors on reconstructed coefficients ranging from 5.18\% to 24.04\% and on reconstructed solutions ranging from 1.99\% to 12.64\%.

Across all PDEs and both fields, Ambient Flow achieves state-of-the-art reconstruction performance, with errors on coefficients ranging from 2.07\% to 7.86\% and errors on solutions ranging from 0.44\% to 0.96\%. Compared to the prior state-of-the-art method (FunDPS at 500 NFE), Ambient Flow achieves a 62.51\% reduction in average overall error while using 125\(\times\) fewer function evaluations, despite being trained directly on partial observations. Specifically, it reduces error by 65.16\% on Darcy flow, 67.32\% on Helmholtz, 47.92\% on Navier--Stokes, and 69.62\% on Poisson, with reductions of 50.49\% on coefficients and 74.52\% on solutions.

These results address the fundamental question: \textit{Ambient Physics can learn the joint distribution of coefficient-solution pairs directly from partial observations}, and it achieves state-of-the-art reconstruction performance.

\begin{table*}[t]
    \centering
    \resizebox{\textwidth}{!}{%
    \begin{tabular}{lcccccccccc}
        \toprule
         & \multirow{2}{*}{\makecell{Partial Obs.\\Training}} &  & \multicolumn{2}{c}{Darcy Flow} & \multicolumn{2}{c}{Helmholtz} & \multicolumn{2}{c}{Navier-Stokes} & \multicolumn{2}{c}{Poisson} \\
        \cline{4-5}\cline{6-7}\cline{8-9}\cline{10-11}
         & & & Coefficient & Solution & Coefficient & Solution & Coefficient & Solution & Coefficient & Solution \\
        \midrule
        DeepONet & \color{red}{\xmark} & $-$ & 41.1\% & 38.3\% & 132.8\% & 123.1\% & 97.2\% & 103.2\% & 105.8\% & 155.5\% \\
        FNO & \color{red}{\xmark} & $-$ & 49.3\% & 28.2\% & 218.2\% & 98.2\% & 96.0\% & 101.4\% & 232.7\% & 100.9\% \\
        PINN & \color{red}{\xmark} & $-$ & 59.7\% & 48.8\% & 160.0\% & 142.3\% & 146.8\% & 142.7\% & 130.0\% & 128.1\% \\
        PINO & \color{red}{\xmark} & $-$ & 49.2\% & 35.2\% & 216.9\% & 106.5\% & 96.0\% & 101.4\% & 231.9\% & 107.1\% \\
        \midrule
        \rowcolor{background!75} &  & $-$ & 6.79\% & \color{red}{24.90\%} & 22.86\% & 5.85\% & 25.78\% & \color{red}{45.92\%} & 22.16\% & 5.76\% \\
        \rowcolor{background!75} & & \multicolumn{1}{l}{$+$\texttt{Mask Conditioning}} & 6.11\% & \color{red}{27.20\%} & 20.69\% & 5.35\% & 26.99\% & \color{red}{41.93\%} & 21.62\% & 7.15\% \\
        \rowcolor{background!75} \multirow{-3}{*}{\textbf{Ambient FNO}} & \multirow{-3}{*}{\color{blue}{\cmark}} & \multicolumn{1}{l}{$+$\texttt{Joint Dist. Modeling}} & \textbf{5.38\%} & \textbf{2.56\%} & \textbf{16.57\%} & \textbf{2.86\%} & \textbf{8.84\%} & \textbf{2.12\%} & \textbf{17.20\%} & \textbf{2.99\%} \\
        \midrule
        \rowcolor{background!75} &  & $-$ & 2.55\% & 2.59\% & 9.16\% & 2.00\% & 8.25\% & 3.57\% & 9.65\% & 1.94\% \\
        \rowcolor{background!75} & & \multicolumn{1}{l}{$+$\texttt{Mask Conditioning}} & \textbf{2.21\%} & 2.57\% & \textbf{7.97\%} & 1.99\% & \textbf{7.35\%} & 3.36\% & \textbf{8.32\%} & 1.87\% \\
        \rowcolor{background!75} \multirow{-3}{*}{\textbf{Ambient UNet}} & \multirow{-3}{*}{\color{blue}{\cmark}} & \multicolumn{1}{l}{$+$\texttt{Joint Dist. Modeling}} & 2.46\% & \textbf{0.64\%} & 10.81\% & \textbf{0.49\%} & 8.33\% & \textbf{1.25\%} & 10.53\% & \textbf{0.48\%} \\
        \bottomrule
    \end{tabular}
    }
    \caption{\textbf{Reconstruction Performance Under Partial Observations.} Relative $L_2$ error (\%) between ground truth and reconstructed coefficients ($\hat{a}$) and solutions ($\hat{u}$) from partial observations $(A_a a, A_u u)$ with 3\% uniformly random measurements. \textit{Note: In prior work these are reported as ``Inverse" for coefficients and ``Forward" for solutions.} ``Partial Obs. Training" indicates whether training uses partial observations ({\color{blue}{\cmark}}) or complete observations ({\color{red}{\xmark}}). {\color{red}{Red}} entries indicate settings that exhibit visible artifacts (\autoref{fig:plot_fno_artifacts}). \setlength{\fboxsep}{0pt}\colorbox{background!75}{Ambient FNO} and \setlength{\fboxsep}{0pt}\colorbox{background!75}{Ambient UNet} perform competitively to diffusion-based methods despite training only on partial observations.}
    \label{tab:2}
\end{table*}

\paragraph{Beyond Diffusion and Flow Matching.}
Prior work observed that DeepONet and FNO perform poorly with sparse inputs \cite{huang2024diffusionpde, yao2025guided}, and therefore use complete observations during training and partial observations only at inference. We ask \textit{whether Ambient Physics can adapt traditional neural PDE solvers to learn directly from partial observations.}

We consider the same experimental setup as before and start from an FNO backbone, which we adapt for reconstruction under partial observations. The goal is to learn directional reconstruction maps, $A_a a \rightarrow u$ and $A_u u \rightarrow a$, while supervising only on the available measurements. This objective can be viewed as an instance of Ambient Physics training: the model is conditioned on masked partial observations $(B_a A_a a, B_u A_u u)$ and supervised on the available measurements $(A_a a, A_u u)$. For $A_a a \rightarrow u$, we effectively set $B_a = I$ and $B_u = 0$, and for $A_u u \rightarrow a$, we set $B_a = 0$ and $B_u = I$. We refer to this model as Ambient FNO.

\begin{figure}[!t]
    \centering
    \includegraphics[width=1.0\linewidth]{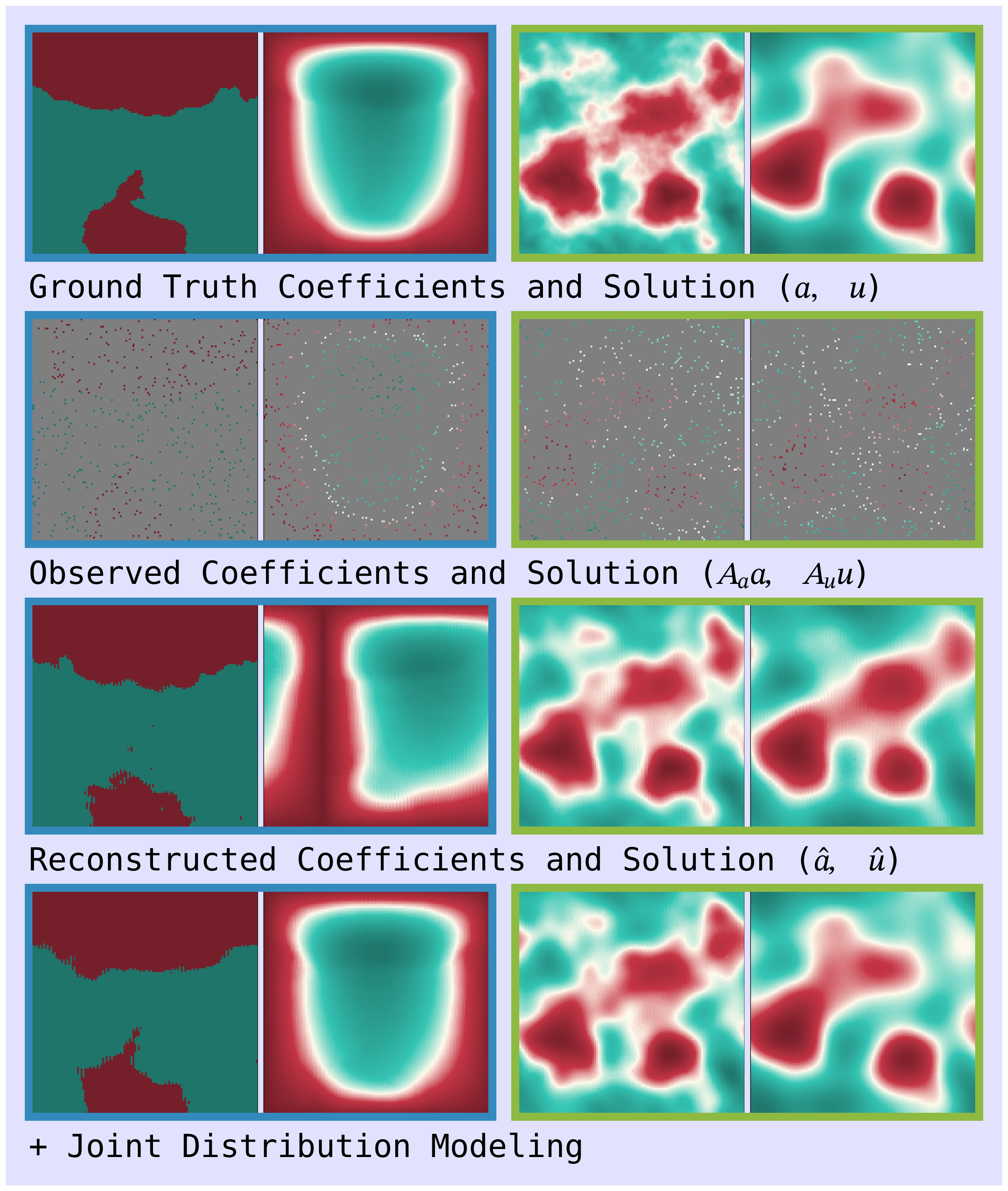}
    \caption{\textbf{Examples of \setlength{\fboxsep}{0pt}\colorbox{background!75}{Ambient FNO.} Reconstructions From Partial Observations.} \textbf{Top to Bottom:} Ground truth coefficients and solution $(a, u)$, partial observations ($A_a a, A_u u$) with 3\% uniformly random measurements, \setlength{\fboxsep}{0pt}\colorbox{background!75}{Ambient FNO} reconstructed coefficients and solutions $(\hat{a}, \hat{u})$, \setlength{\fboxsep}{0pt}\colorbox{background!75}{Ambient FNO} $+$\texttt{Joint Dist. Modeling} reconstructed coefficients and solutions $(\hat{a}, \hat{u})$. \textbf{Left:} \setlength{\fboxsep}{0pt}\colorbox{darcyflow!50}{Darcy flow}. \textbf{Right:} \setlength{\fboxsep}{0pt}\colorbox{navierstokes!50}{Navier-Stokes}. Within each panel, the left field represents the coefficients (or initial condition) and the right field represents the solution (or final state).}
    \label{fig:plot_fno_artifacts}
\end{figure}

\autoref{tab:2} summarizes reconstruction performance across PDEs and fields. Compared to a vanilla FNO, Ambient FNO achieves a 74.27\% reduction in average overall error despite training only on partial observations. However, the reconstructed solutions $\hat{u}$ occasionally exhibit visible artifacts: Darcy flow solutions can appear spatially shifted, and Navier-Stokes final states can resemble blurred initial conditions (see \autoref{fig:plot_fno_artifacts}). We test two modifications: \(+\)\texttt{Mask Conditioning}, which yields minimal improvement, and \(+\)\texttt{Joint Dist. Modeling}, where $B_a$ and $B_u$ mask only a subset of already-observed values ($B_a, B_u \notin \{I, 0\}$), which yields substantial improvements and eliminates visible artifacts. Compared to DiffusionPDE, Ambient FNO \(+\)\texttt{Joint Dist. Modeling} achieves a 36.16\% reduction in average overall error and is competitive with FunDPS at 200 NFE. We repeat these ablations using the UNet of \citet{ho2020denoising}: compared to the prior state-of-the-art method (FunDPS at 500 NFE), Ambient UNet achieves a 53.76\% reduction in average overall error.

These results address our question: \textit{Ambient Physics can adapt traditional neural PDE solvers to learn directly from partial observations,} and it is not tied to diffusion or flow matching.

\begin{figure*}[t]
    \centering
    \includegraphics[width=1.0\linewidth]{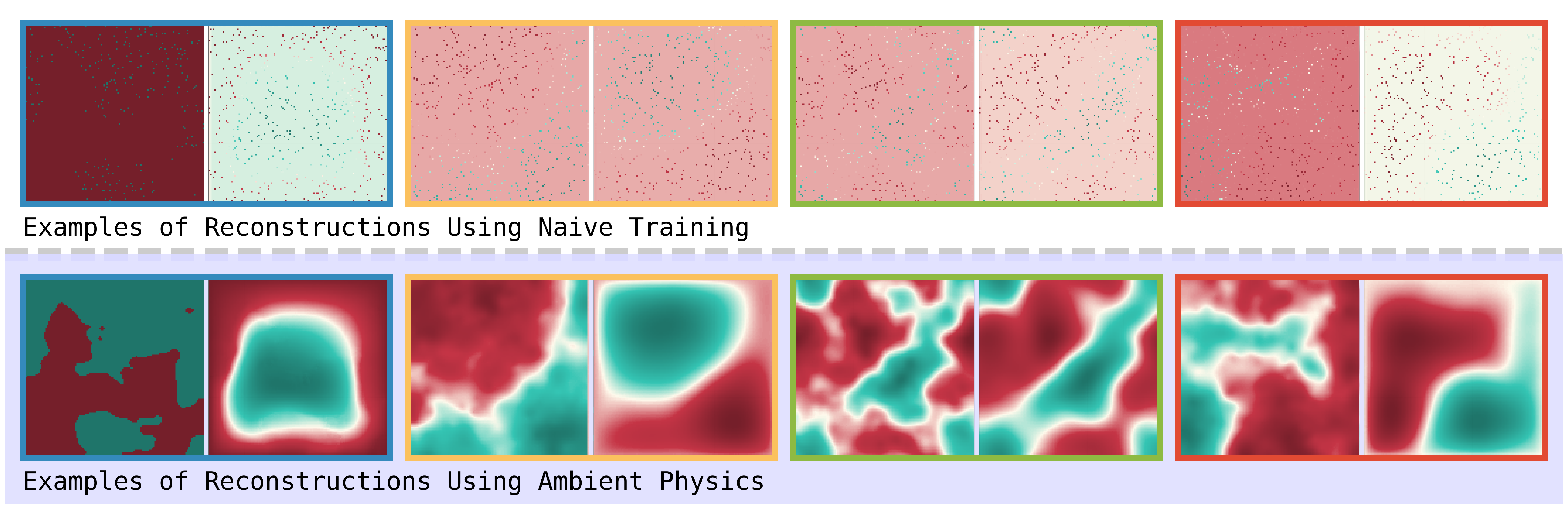}
    \caption{\textbf{Naive Training vs Ambient Physics Predictions from Partial Observations.} \textbf{Top:} Naively training on partial observations yields a model that predicts accurately only at observed locations and arbitrarily elsewhere. \textbf{Bottom:} \setlength{\fboxsep}{0pt}\colorbox{background!75}{Ambient Physics}, \textit{trained on the same partial observations,} yields a model that predicts physically-plausible fields everywhere. \textbf{Left to Right:} \setlength{\fboxsep}{0pt}\colorbox{darcyflow!50}{Darcy flow} with point observations, \setlength{\fboxsep}{0pt}\colorbox{helmholtz!50}{Helmholtz} with patch observations, \setlength{\fboxsep}{0pt}\colorbox{navierstokes!50}{Navier-Stokes} with slice observations, and \setlength{\fboxsep}{0pt}\colorbox{poisson!50}{Poisson} with window observations. Within each panel, the left field represents the coefficients (or initial condition) and the right field represents the solution (or final state).}
    \label{fig:plot_predictions_naive_vs_ambient}
\end{figure*}

\begin{figure}[!t]
    \centering
    \includegraphics[width=1.0\linewidth]{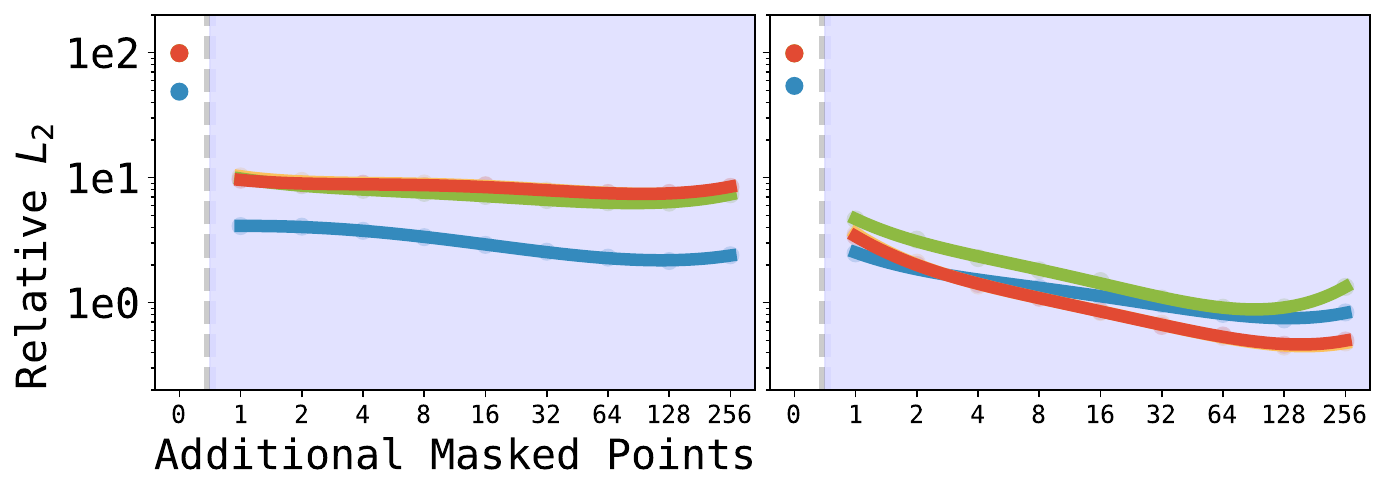}
    \caption{\textbf{One-Point Transition to \setlength{\fboxsep}{0pt}\colorbox{background!75}{Ambient Physics}.} Relative $L_2$ error (log scale) versus the number of additional masked already-observed points used during training, across \setlength{\fboxsep}{0pt}\colorbox{darcyflow!50}{Darcy flow}, \setlength{\fboxsep}{0pt}\colorbox{helmholtz!50}{Helmholtz}, \setlength{\fboxsep}{0pt}\colorbox{navierstokes!50}{Navier-Stokes}, and \setlength{\fboxsep}{0pt}\colorbox{poisson!50}{Poisson}. \textbf{Left:} Coefficient error. \textbf{Right:} Solution error. The point at 0 additional masked points corresponds to naive training ($B=I$); masking even a single already-observed point leads to a $\sim 10$--$100\times$ reduction in error.}
    \label{fig:plot_mask_results}
\end{figure}

\paragraph{One-Point Transition to Ambient Physics.}
In Ambient Physics training (\autoref{method}), we randomly mask a subset of already-observed measurements but still supervise on all available measurements. We ask how large this subset must be. We consider the same experimental setup as before. Starting from naive training ($B = I$), we sample masking operators that withhold $\{0,1,2,4,8,16,32,64,128,256\}$ observed points. \autoref{fig:plot_mask_results} shows a sharp threshold: masking a single observed point yields an $\sim 10\times$ reduction in coefficient error and an $\sim 100\times$ reduction in solution error; performance improves up to $128$ extra masked points. This ``one-point transition'' from naive to Ambient Physics training, under an otherwise identical setup, provides evidence that the improvements are primarily attributable to the Ambient Physics framework, rather than incidental choices such as architecture or optimization (see \autoref{one-point_transition}).

\paragraph{Robustness Across Patterns and Super-Resolution.}
Thus far, we have studied Ambient Physics under $3\%$ uniformly random measurements. In many scientific settings, measurements are structured (e.g., localized around probes, collected along slices, or restricted to a region). Here, we study Darcy flow under structured observation patterns---\textit{patch}, \textit{column}, and \textit{window} measurements---across multiple sparsity levels. Across all patterns, we show that Ambient Physics accurately reconstructs both coefficients and solutions, and that its reconstruction performance is not limited to uniformly random sampling (see \autoref{robustness_across_patterns}).

In addition, we can treat low-resolution measurements as partial observations of an underlying high-resolution field. We study this setting, and show that Ambient Physics can super-resolution a $32\times32$ measurement into a $128\times128$ reconstruction, learning directly from low-resolution measurements, without a single $128\times128$ measurement.

\section{Conclusions}
\label{conclusions}

We introduced Ambient Physics, a framework for learning the joint distribution of coefficient--solution pairs $p(a,u)$ directly from partial observations, without requiring a single fully observed training example. Across four diverse PDE, our instantiation Ambient Flow achieves state-of-the-art reconstruction performance under extreme sparsity while reducing the number of function evaluations by orders of magnitude. We further showed that Ambient Physics is not tied to diffusion or flow matching, and identified a ``one-point transition" in which masking a single already-observed point is sufficient to move from catastrophic failure to effective learning.

Future work can apply Ambient Physics to real-world scientific and engineering settings where complete observations are unavailable, and explore training a single model that generalizes across diverse measurement patterns and resolutions.

\section*{Impact Statement}
This paper presents work whose goal is to advance the field of machine learning. There are many potential societal consequences of our work, none of which we feel must be specifically highlighted here.

\bibliography{example_paper}
\bibliographystyle{icml2026}

\newpage
\appendix
\onecolumn

\section{Implementation Details}
\label{implementation_details}

\paragraph{Hardware Details.} All models were trained on NVIDIA H100 GPUs. In total, we trained roughly 200 models for the main results and roughly 100 models for early-stage experimentation (e.g., preliminary exploration and hyperparameter tuning), for a total of $\sim$300 models. Across experiments, the average training time was approximately 250--300 minutes per model (including data loading and evaluation under our standard experimental setup).

\paragraph{Data Details.}
We use the same PDE datasets as DiffusionPDE~\cite{huang2024diffusionpde} and FunDPS~\cite{yao2025guided}: Darcy flow, Helmholtz, unbounded Navier--Stokes, and Poisson. For each PDE, we use 50,000 training samples and 1,000 test samples at a spatial resolution of $128\times128$.

DiffusionPDE and FunDPS also include a bounded Navier--Stokes dataset; however, we found that approximately 10\% of its samples contain non-physical artifacts, and we therefore deemed it unsuitable for our evaluation.

We emphasize that Ambient Physics models never receive fully observed fields as inputs during training or inference. Instead, partial observations $(A_a a, A_u u)$ are generated in the data loader by applying the corresponding measurement operators. In our implementation, the observation pattern is fixed per sample throughout training (i.e., from one epoch to the next), so unobserved locations remain unobserved for that sample.

\paragraph{Architecture Details.}
For Ambient Flow and Ambient UNet, we use the UNet backbone from DDPM \cite{ho2020denoising}. For Ambient FNO, we use a modern Fourier Neural Operator (FNO) backbone based on \citet{kossaifi2023multi}. Architecture hyperparameters are listed in \autoref{tab:hparams}.

\paragraph{Optimization Details.}
For all models, we use the AdamW optimizer \cite{loshchilov2017decoupled}. Optimization hyperparameters are listed in \autoref{tab:hparams}.

\begin{table}[h]
    \centering
    \small
    \setlength{\tabcolsep}{8pt}
    \begin{tabular}{llll}
        \toprule
        \textbf{Hyperparameter} & \textbf{Ambient Flow} & \textbf{Ambient UNet} & \textbf{Ambient FNO} \\
        \midrule
        \multicolumn{4}{l}{\textbf{Architecture}} \\
        \midrule
        \texttt{fourier\_channels} & 64 & -- & -- \\
        \texttt{hidden\_channels} & 192 & 64 & 64 \\
        \texttt{ch\_mult} & [1, 2, 3, 4] & [1, 2, 3, 4] & -- \\
        \texttt{is\_attn} & [\texttt{False}, \texttt{False}, \texttt{True}, \texttt{True}] & [\texttt{False}, \texttt{False}, \texttt{False}, \texttt{True}] & -- \\
        \texttt{md\_attn} & \texttt{True} & \texttt{True} & -- \\
        \texttt{dropout} & 0.1 & 0.0 & -- \\
        \texttt{attn\_heads} & 4 & 4 & -- \\
        \texttt{attn\_head\_channels} & 32 & 32 & -- \\
        \texttt{depth} & -- & -- & 8 \\
        \texttt{fft\_modes} & -- & -- & 16 \\
        \texttt{activation} & -- & -- & \texttt{GELU} \cite{hendrycks2016gaussian} \\
        \midrule
        \multicolumn{4}{l}{\textbf{Optimization}} \\
        \midrule
        \texttt{epochs} & 25 & 100 & 100 \\
        \texttt{clip\_grad\_max\_norm} & 1.0 & 1.0 & 1.0 \\
        \texttt{mixed\_precision} & \texttt{True} & \texttt{False} & \texttt{False} \\
        \texttt{lr} & $1.0\times 10^{-4}$ & $1.0\times 10^{-4}$ & $1.0\times 10^{-4}$ \\
        \texttt{weight\_decay} & $1.0\times 10^{-5}$ & $1.0\times 10^{-5}$ & $1.0\times 10^{-5}$ \\
        \texttt{betas} & [0.9, 0.95] & [0.9, 0.999] & [0.9, 0.999] \\
        \texttt{eps} & $1.0\times 10^{-8}$ & $1.0\times 10^{-8}$ & $1.0\times 10^{-8}$ \\
        \bottomrule
    \end{tabular}
    \caption{Hyperparameters used by Ambient Flow, Ambient UNet, and Ambient FNO.}
    \label{tab:hparams}
\end{table}

\newpage

\section{Partial Differential Equations}
\label{partial_differential_equations}

\paragraph{Darcy Flow Equation.} The Darcy flow equation is a second-order elliptic partial differential equation used to model the flow of viscous incompressible fluid through a porous medium. Its two-dimensional variant can be expressed as:
\[
- \nabla \cdot \left( a \left( \mathbf{x} \right) \nabla u \left( \mathbf{x} \right) \right) = f \left( \mathbf{x} \right), \quad \mathbf{x} \in \left( 0,1 \right)^2,
\]
with zero boundary conditions \(u|_{\partial \Omega} = 0\). Here, the coefficient \(a \left( \mathbf{x} \right)\) represents the spatially varying permeability of the medium and typically takes binary values to model heterogeneous materials, while \(f \left( \mathbf{x} \right)\) (often set to a constant) represents the forcing term. The binary nature of \(a \left( \mathbf{x} \right)\) introduces sharp discontinuities in permeability, and learning-based methods must represent these discontinuities while keeping the solution field physically consistent.

\paragraph{Inhomogeneous Helmholtz Equation.} The inhomogeneous Helmholtz equation is a second-order elliptic partial differential equation used to model time-harmonic wave propagation in heterogeneous media. Its two-dimensional variant can be expressed as:
\[
\nabla^2 u \left( \mathbf{x} \right) + k^2 u \left( \mathbf{x} \right) = a \left( \mathbf{x} \right), \quad \mathbf{x} \in \left( 0,1 \right)^2,
\]
with zero boundary conditions \(u|_{\partial \Omega} = 0\). Here, \(k\) denotes the wave number (typically set to \(k=1\)), and \(a \left( \mathbf{x} \right) \) represents the spatially varying source term, often modeled as a piecewise-constant function. The Laplacian term \(\nabla^2 u\) governs diffusion and the \(k^2 u\) term introduces oscillations, which can yield interference patterns and local resonances. When \(k = 0\), the equation reduces to the Poisson equation.

\paragraph{Unbounded Incompressible Navier-Stokes Equation.} The incompressible Navier-Stokes equations are a system of nonlinear partial differential equations that govern the dynamics of viscous fluid flow. For the unbounded velocity formulation, the two-dimensional variant can be expressed as:
\begin{align*}
\partial_t w \left( \mathbf{x}, \tau \right) + \mathbf{v} \left( \mathbf{x}, \tau \right) \cdot \nabla w \left( \mathbf{x}, \tau \right) &= \nu \Delta w \left( \mathbf{x}, \tau \right) + q \left( \mathbf{x} \right), \quad \mathbf{x} \in \left( 0,1 \right)^2, \; \tau \in \left( 0,T \right], \\
\nabla \cdot \mathbf{v} \left( \mathbf{x}, \tau \right) &= 0,
\end{align*}
where \( w = \nabla \times \mathbf{v} \) is the vorticity, \( \mathbf{v} \left( \mathbf{x}, \tau \right) \) is the velocity field, \( q \left( \mathbf{x} \right) \) is a fixed force field, and \( \nu \) is the kinematic viscosity. With \( \nu = 10^{-3} \), the corresponding Reynolds number \( Re = 1 / \nu = 1000 \) places the flow in the turbulent regime, where nonlinear advection and viscous diffusion generate vortices, shear layers, and energy cascades. The divergence-free constraint \( \nabla \cdot \mathbf{v} = 0 \) enforces incompressibility, coupling the velocity and vorticity fields through the stream function. Learning-based methods must model the time evolution while respecting these constraints.

\paragraph{Poisson Equation.} The Poisson equation is a second-order elliptic partial differential equation that describes steady-state diffusion processes. Its two-dimensional variant can be expressed as:
\[
\nabla^2 u \left( \mathbf{x} \right) = a \left( \mathbf{x} \right), \quad \mathbf{x} \in \left( 0,1 \right)^2,
\]
with homogeneous Dirichlet boundary conditions \( u|_{\partial \Omega} = 0 \). Here, \( a \left( \mathbf{x} \right) \) represents the source term, which often exhibits piecewise-constant structure to model localized sources or sinks in heterogeneous domains. Although the solution field is typically smooth, discontinuities in \(a\left(\mathbf{x}\right)\) influence \(u\left(\mathbf{x}\right)\) through the elliptic operator, and models must respect the boundary conditions.

\newpage

\section{One-Point Transition to Ambient Physics}
\label{one-point_transition}

\begin{figure}[H]
    \centering
    \includegraphics[width=0.45\textwidth]{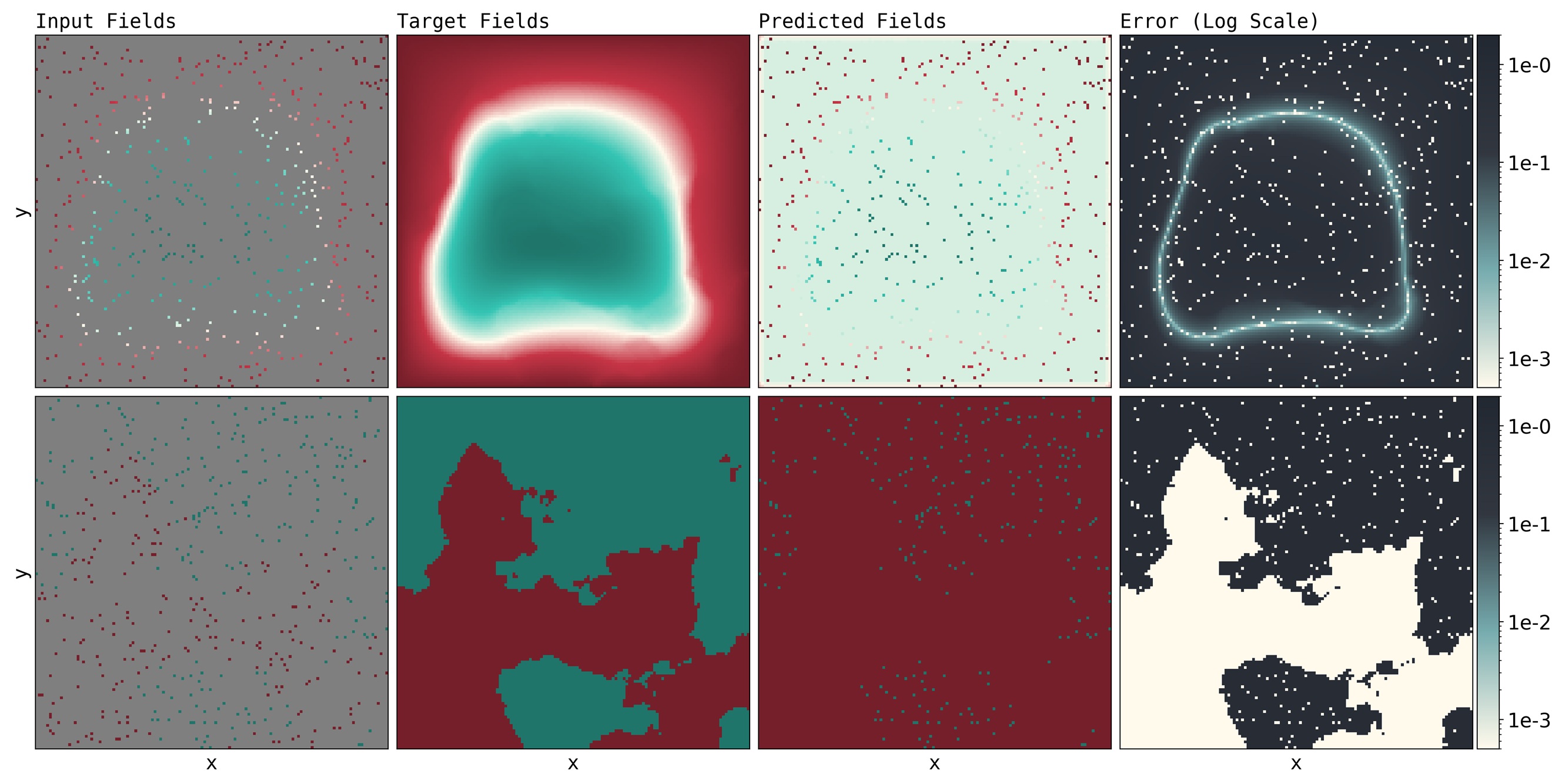}
    \includegraphics[width=0.45\textwidth]{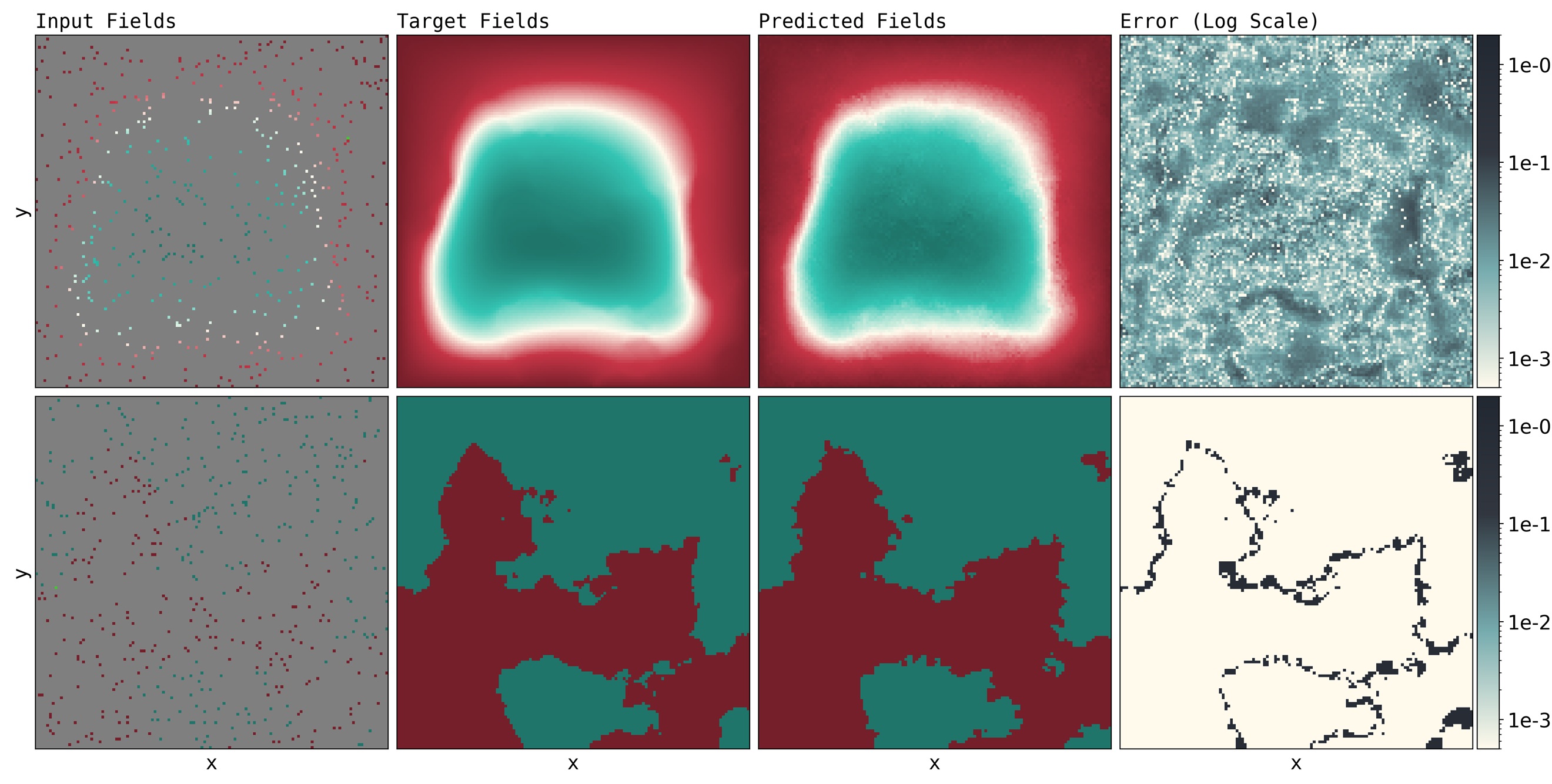}\\[1em]
    \includegraphics[width=0.45\textwidth]{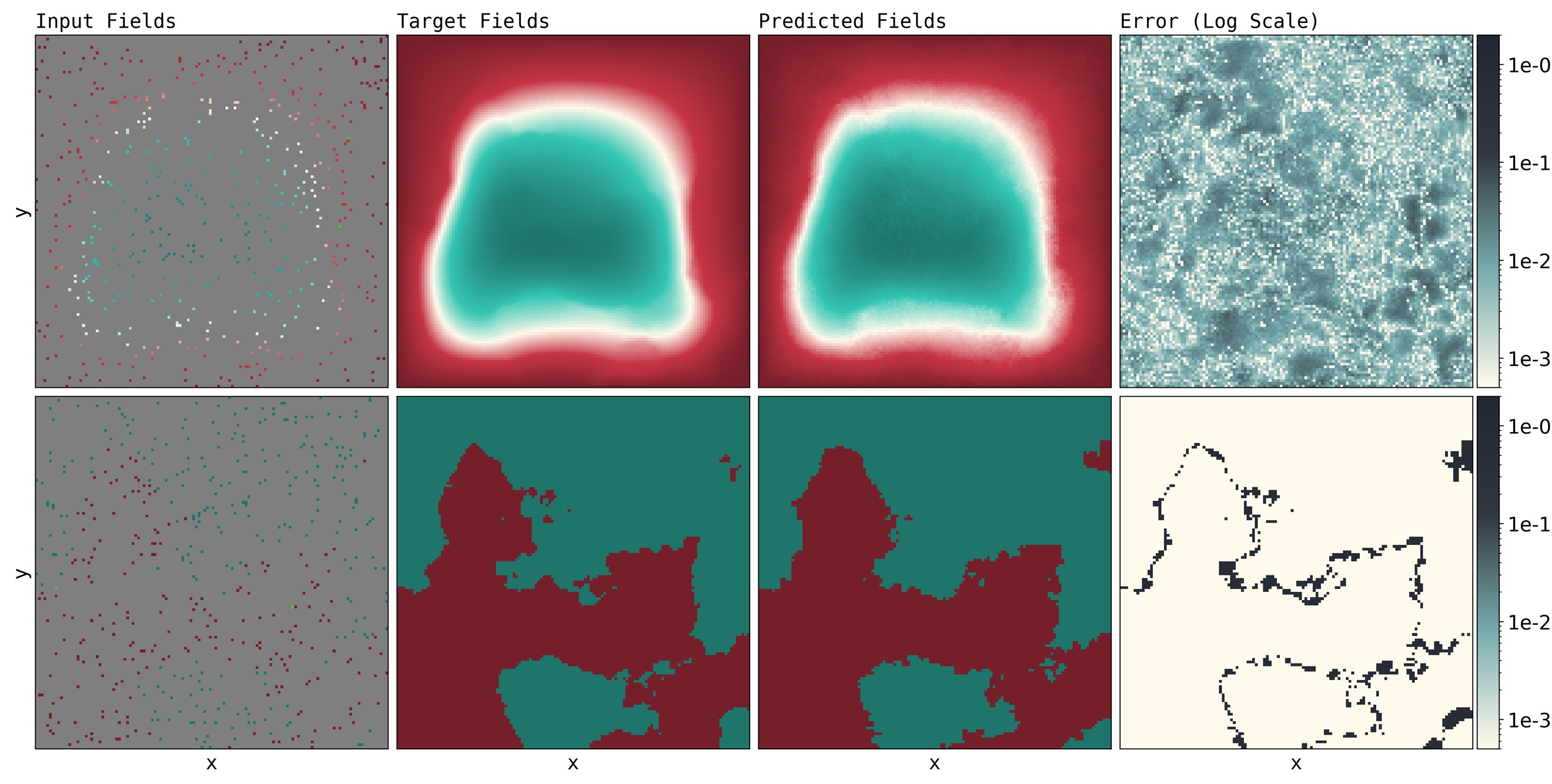}
    \includegraphics[width=0.45\textwidth]{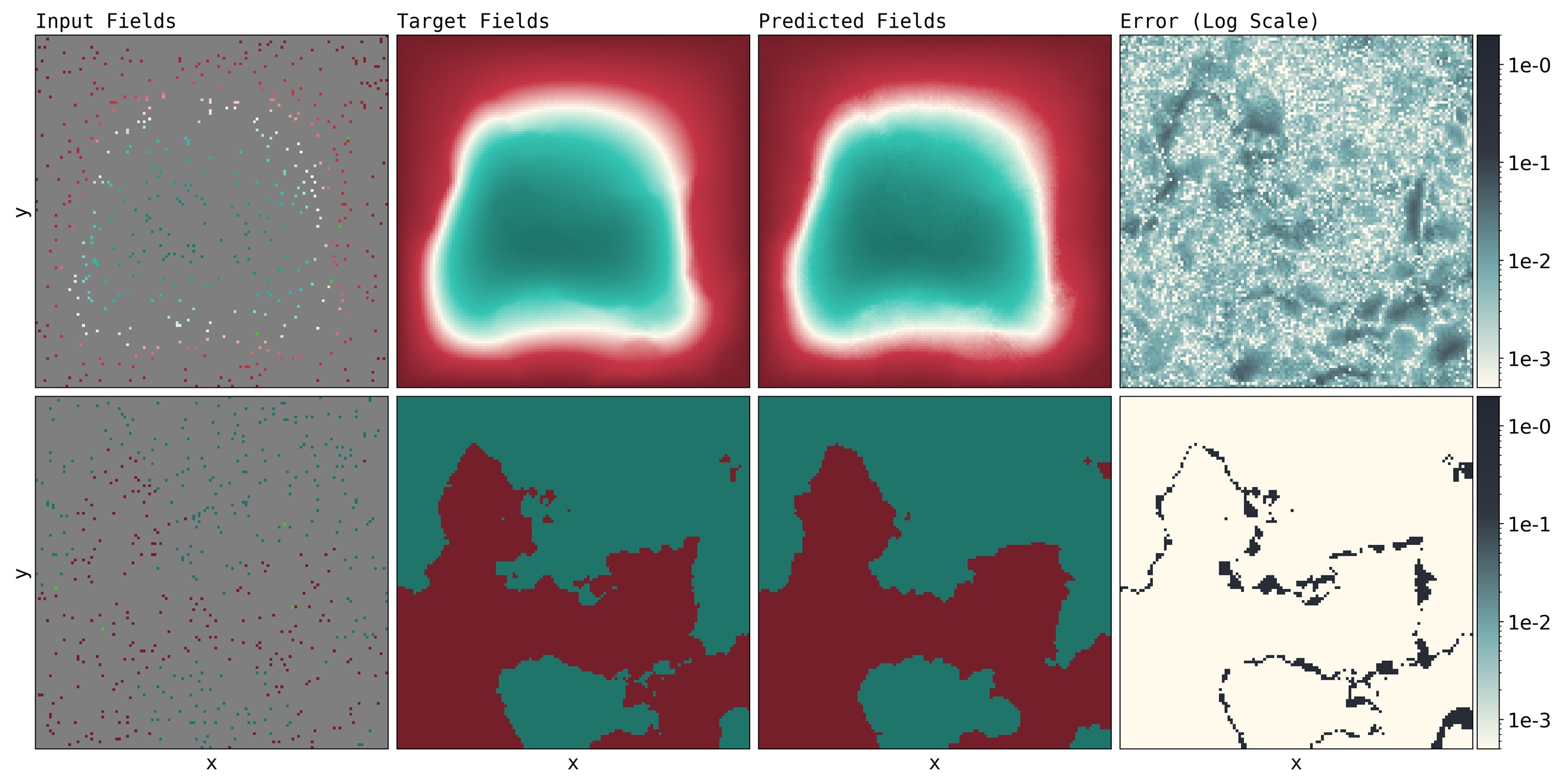}\\[1em]
    \includegraphics[width=0.45\textwidth]{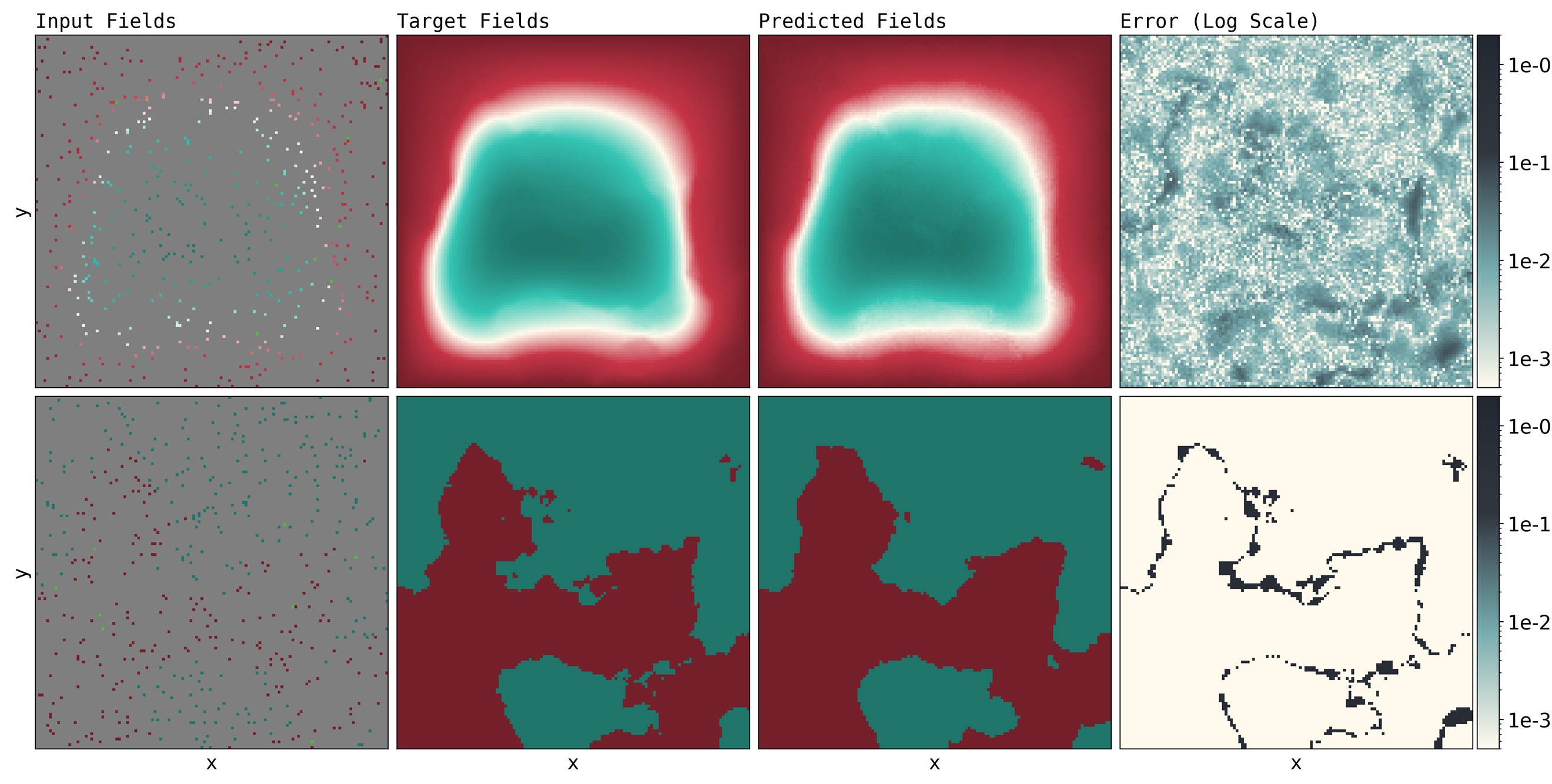}
    \includegraphics[width=0.45\textwidth]{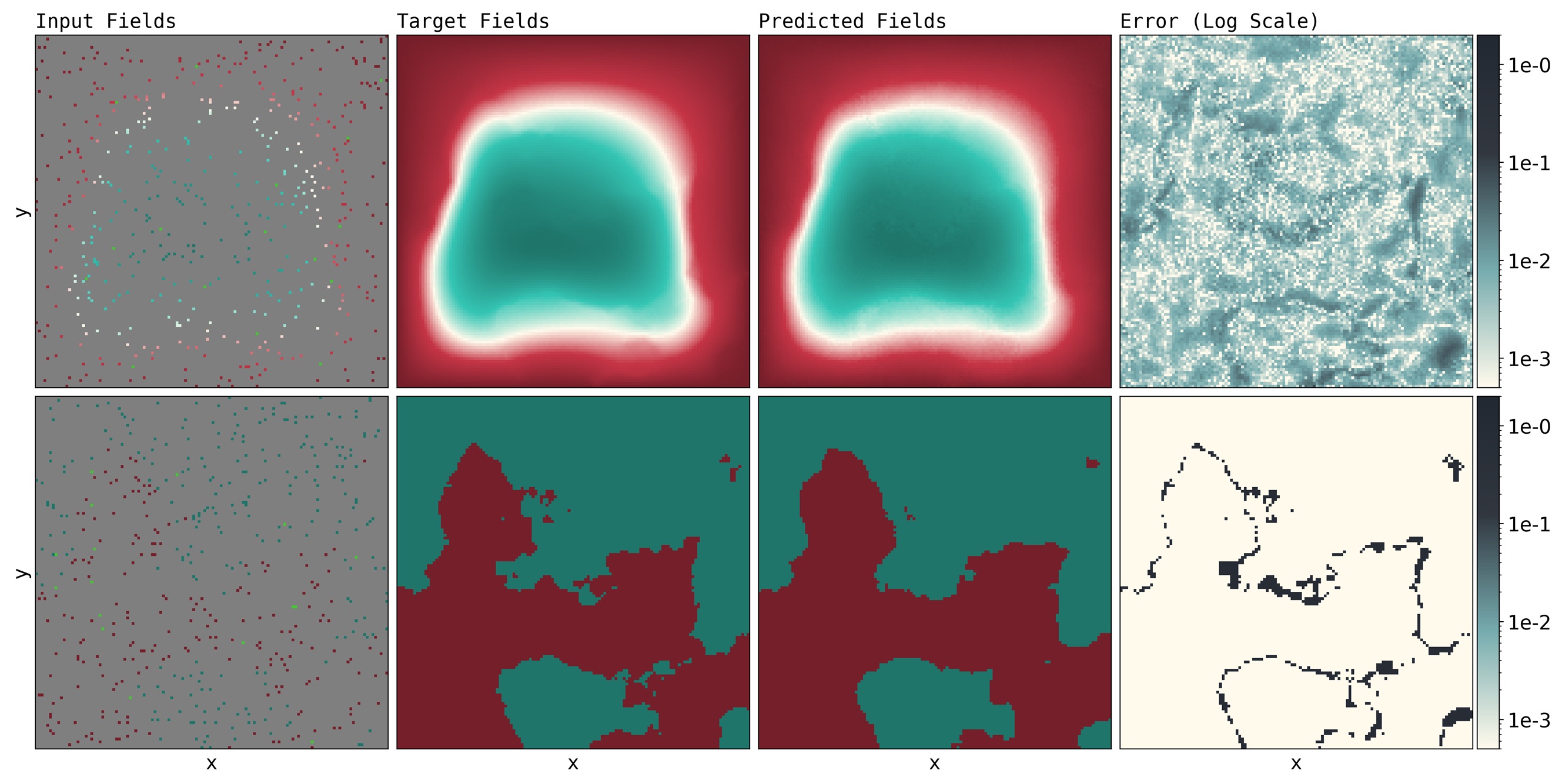}\\[1em]
    \includegraphics[width=0.45\textwidth]{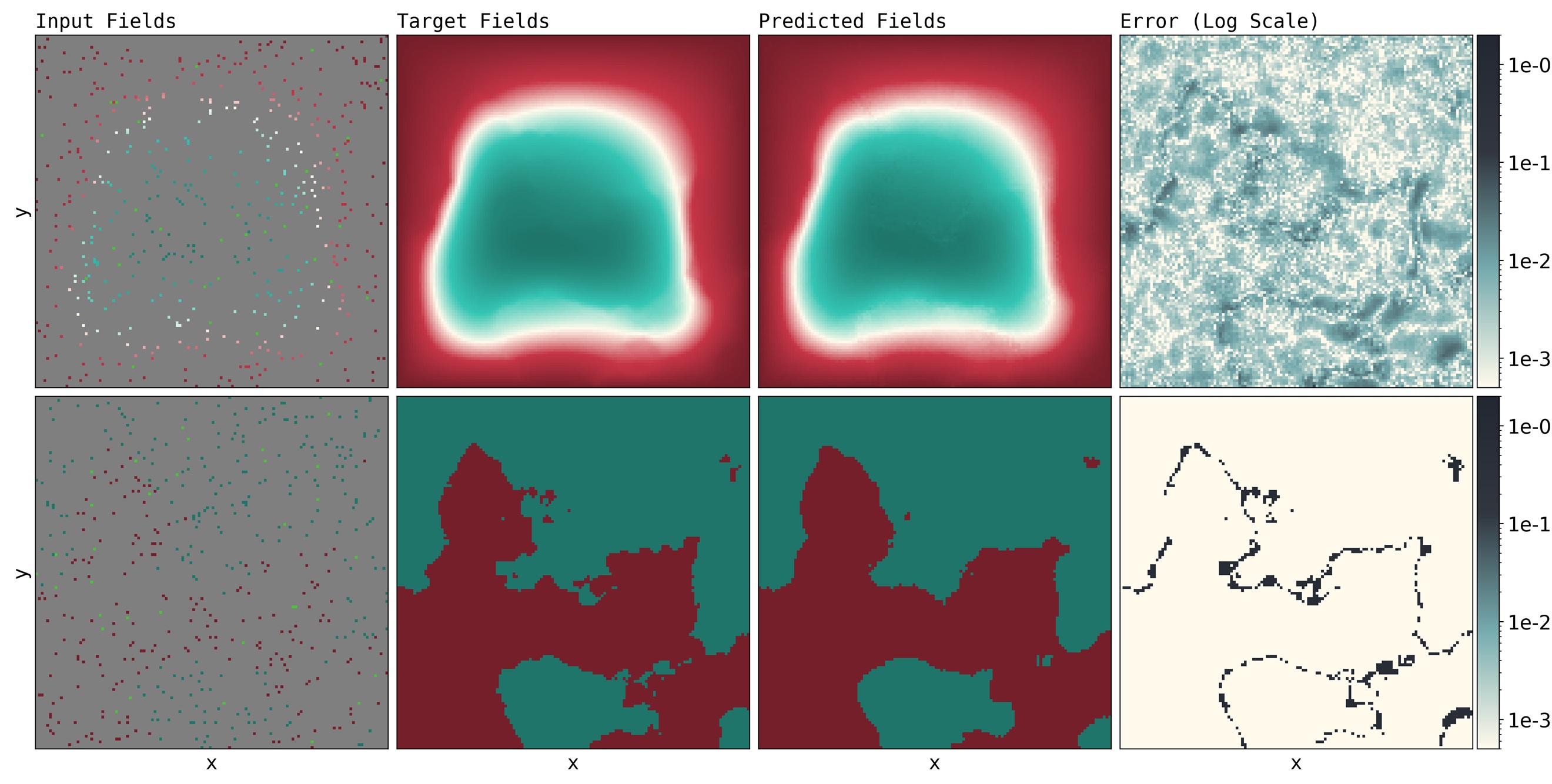}
    \includegraphics[width=0.45\textwidth]{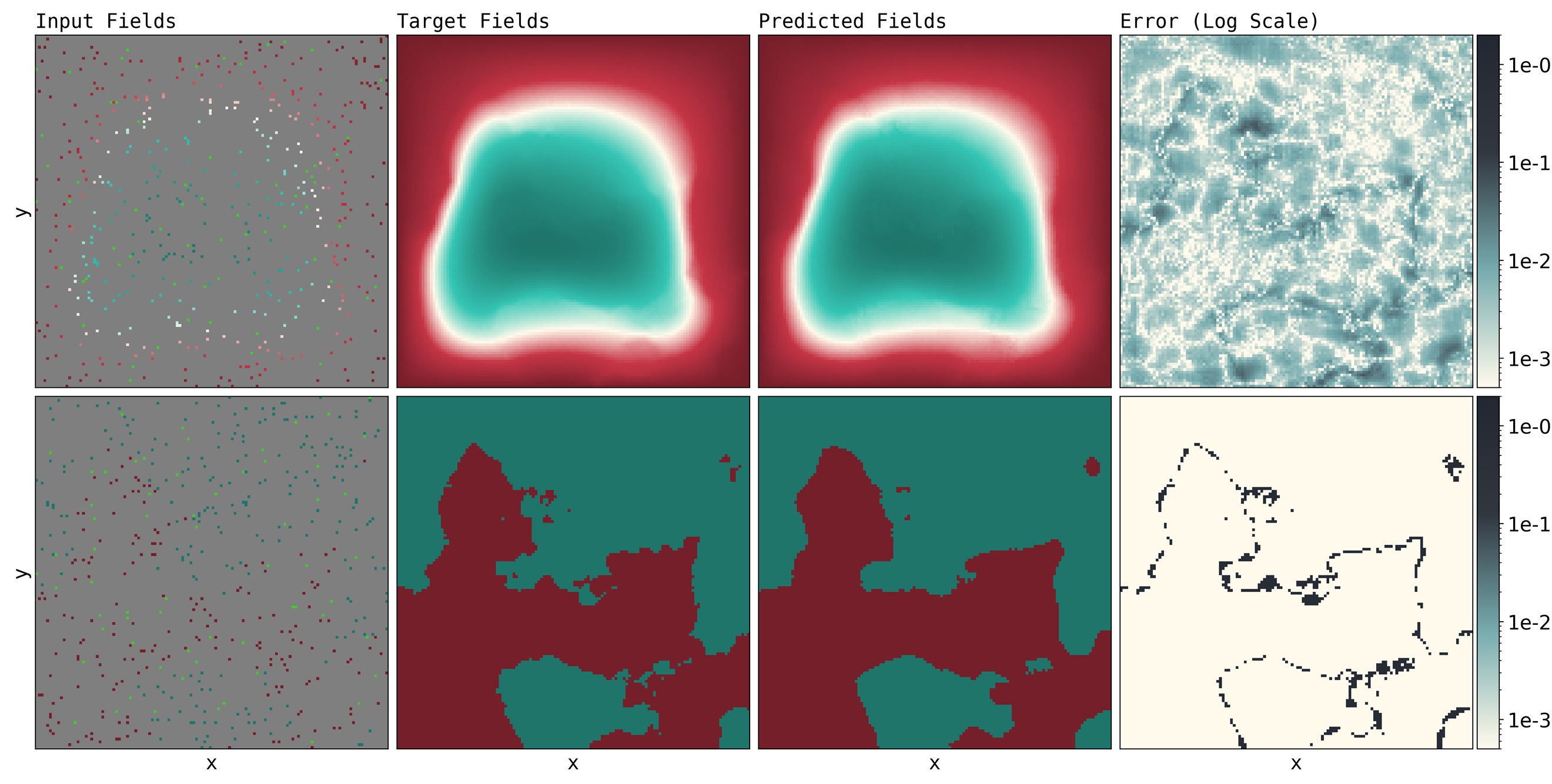}
    \caption{\textbf{Example of ``One-Point Transition" (\setlength{\fboxsep}{0pt}\colorbox{darcyflow!50}{Darcy Flow}).} Each subplot shows an Ambient Flow reconstruction from partial observations (3\% uniformly random points), varying the number of additional masked (already-observed) points used during training (increasing across the grid; top-left: 0, which corresponds to naive training). \textbf{Within Each Subplot:} \textbf{Top:} Solution field. \textbf{Bottom:} Coefficient Field. \textbf{Left to Right:} Input Fields, Target Fields, Reconstructed Fields, Error (Log Scale). Within each input panel, gray pixels represent unobserved locations and green pixels represent masked measurements withheld from the model.}
\end{figure}

\newpage

\begin{figure}[H]
    \centering
    \includegraphics[width=0.45\textwidth]{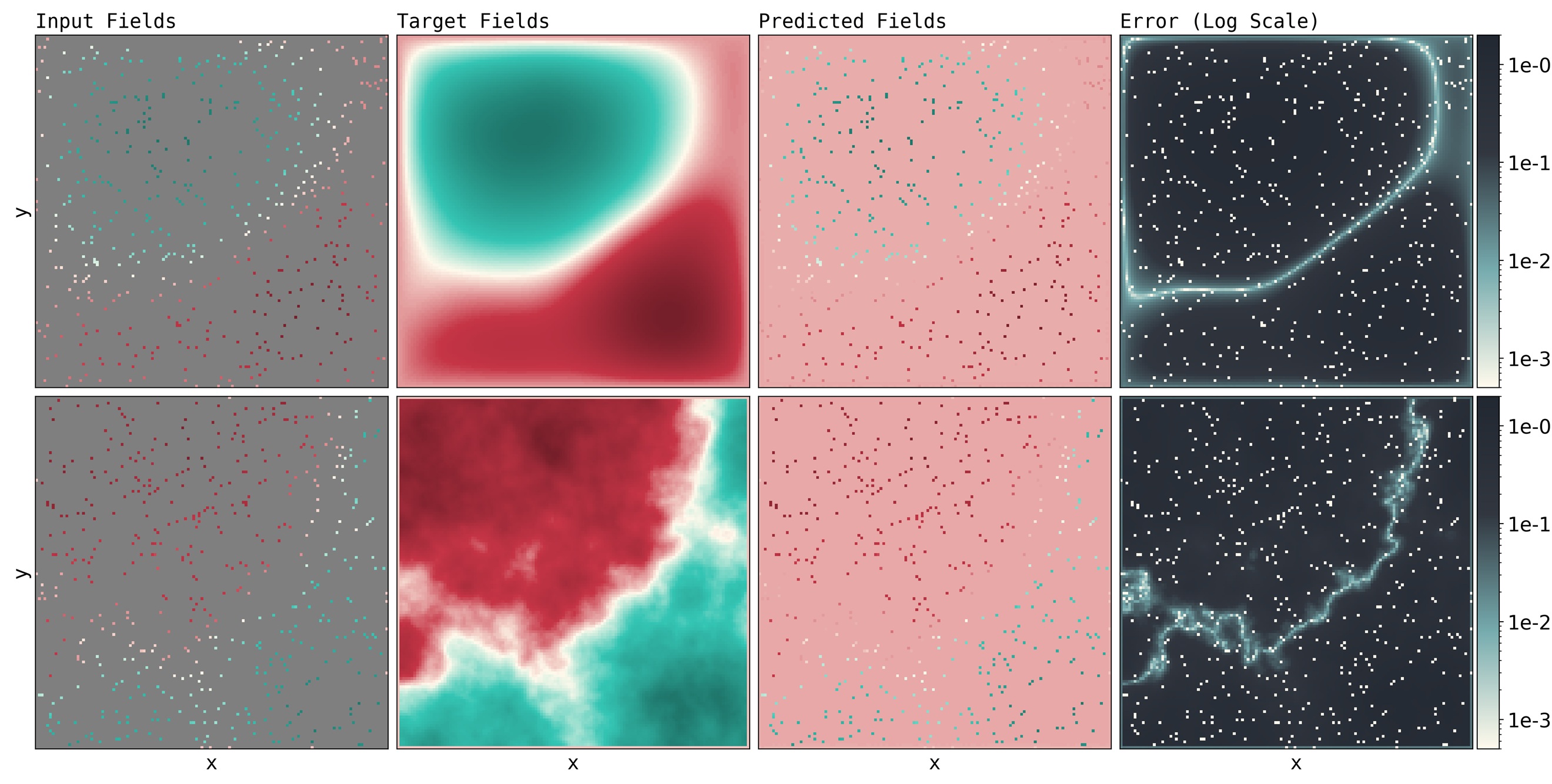}
    \includegraphics[width=0.45\textwidth]{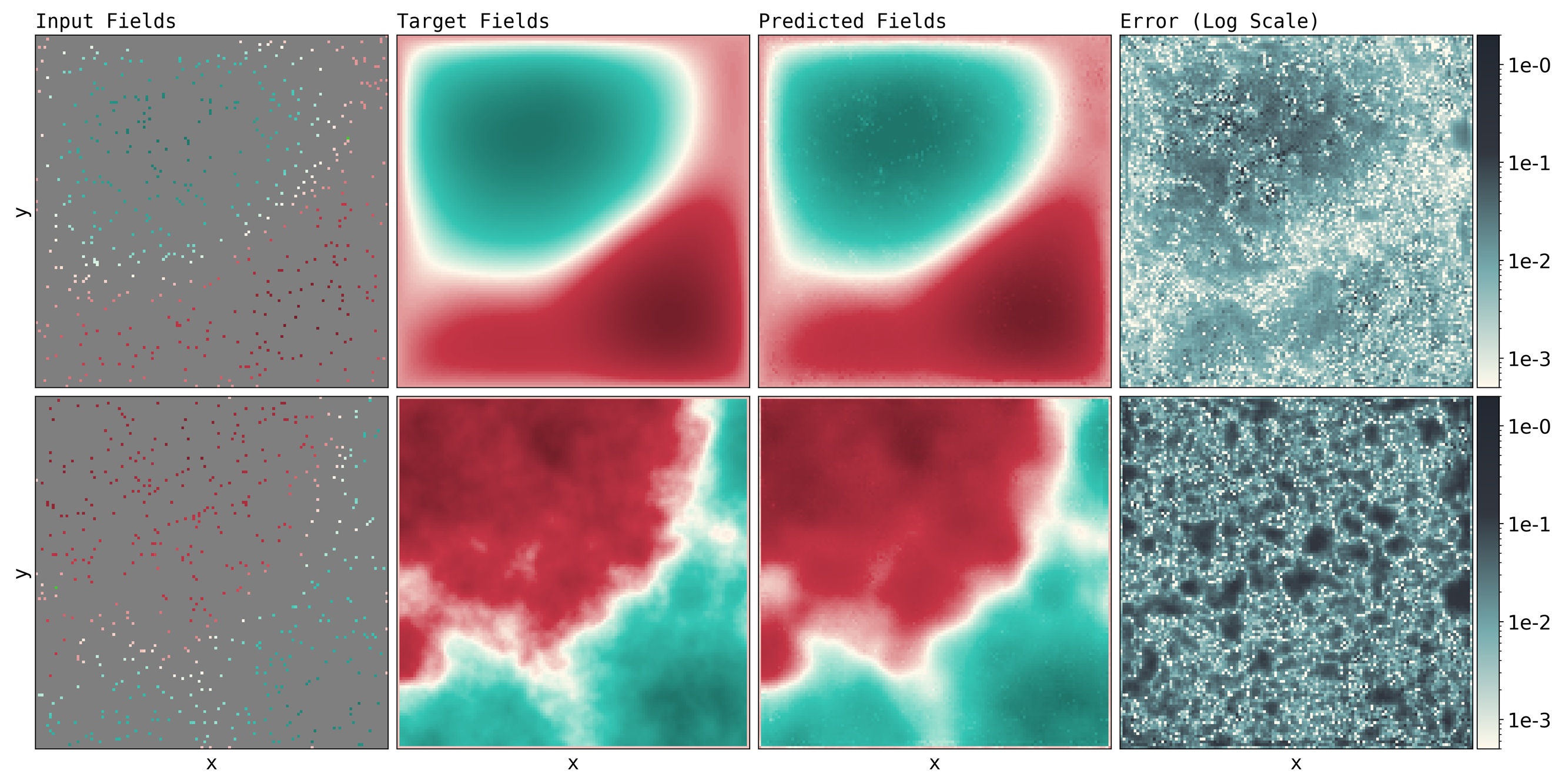}\\[1em]
    \includegraphics[width=0.45\textwidth]{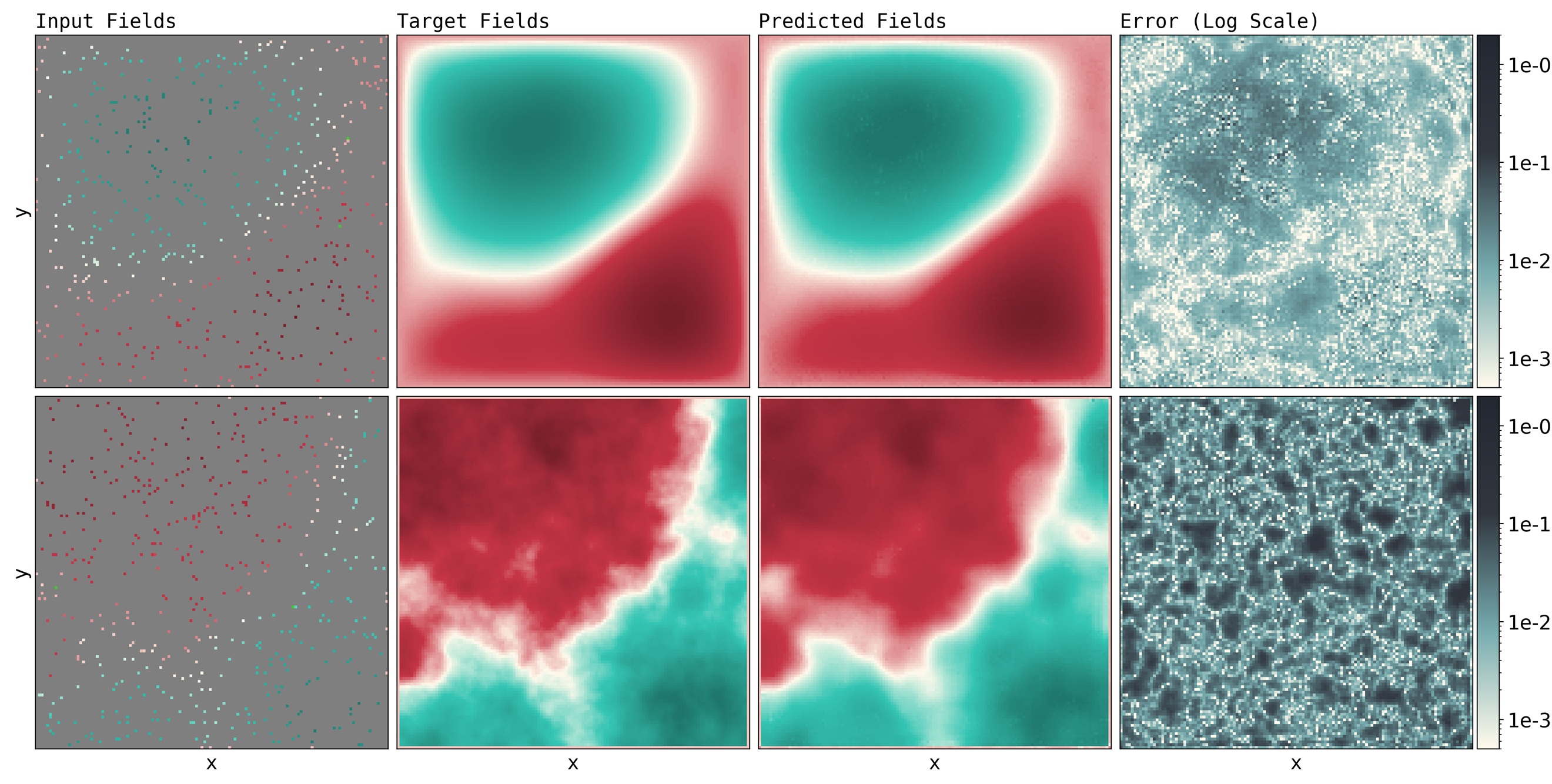}
    \includegraphics[width=0.45\textwidth]{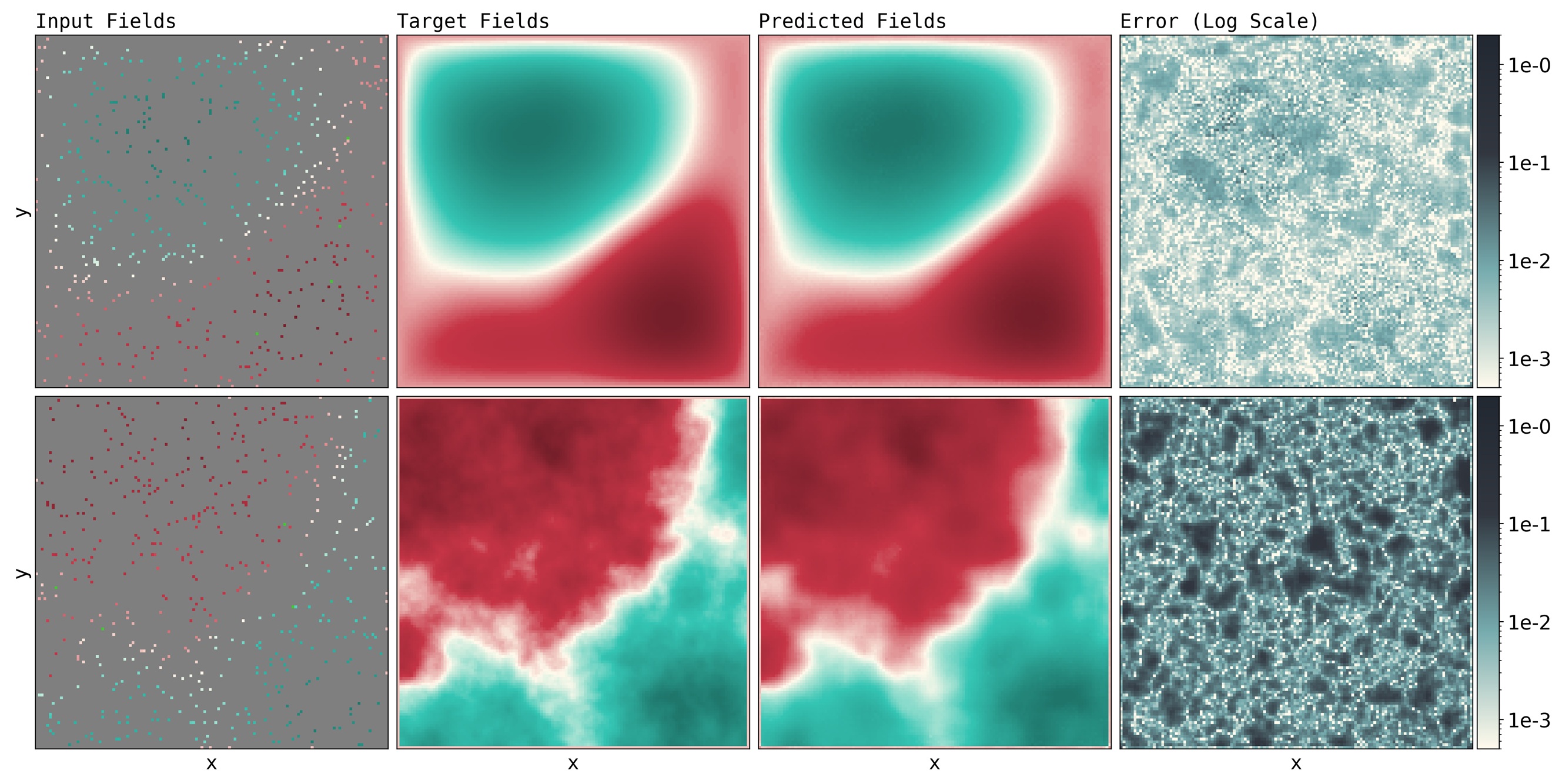}\\[1em]
    \includegraphics[width=0.45\textwidth]{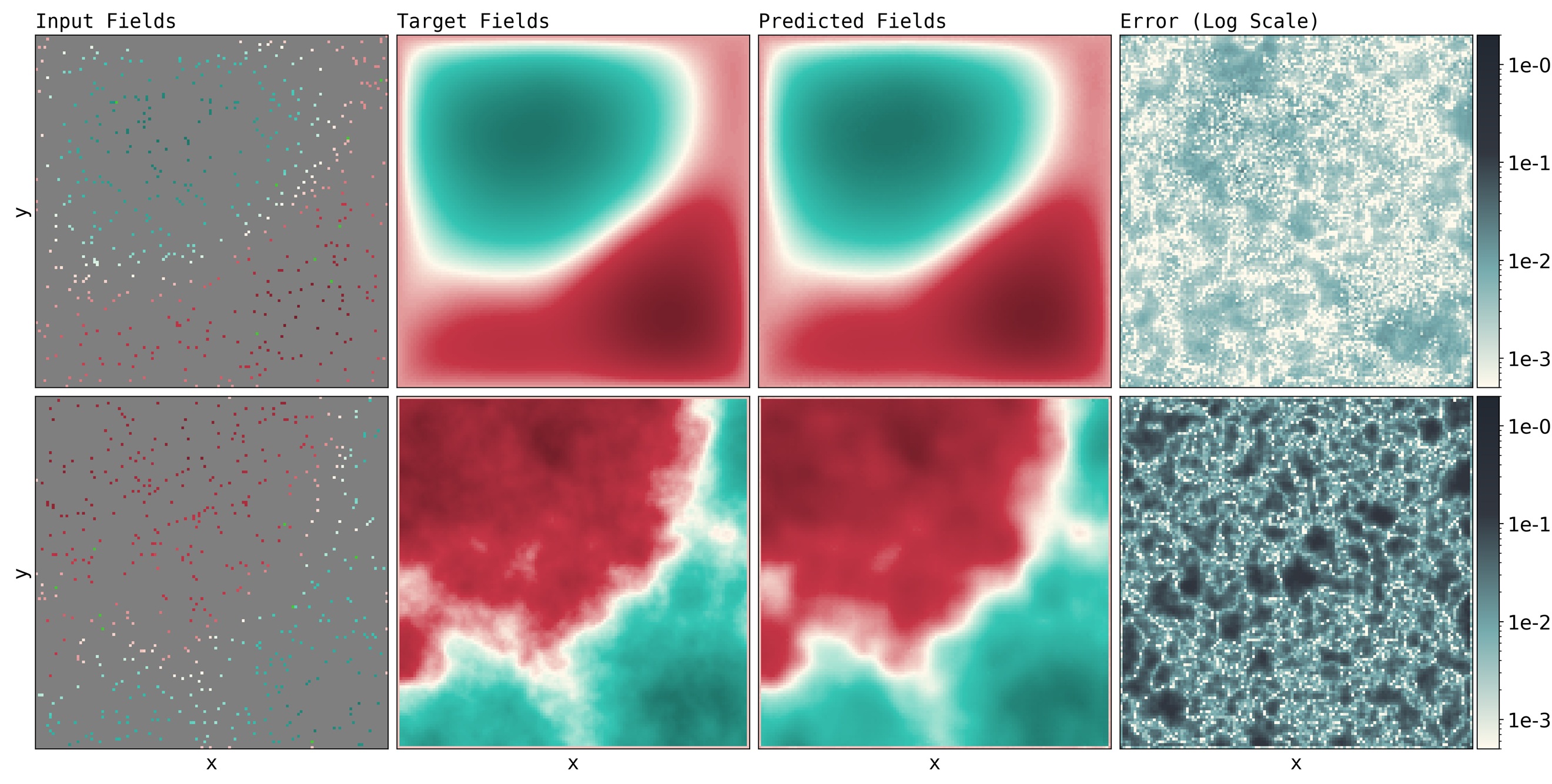}
    \includegraphics[width=0.45\textwidth]{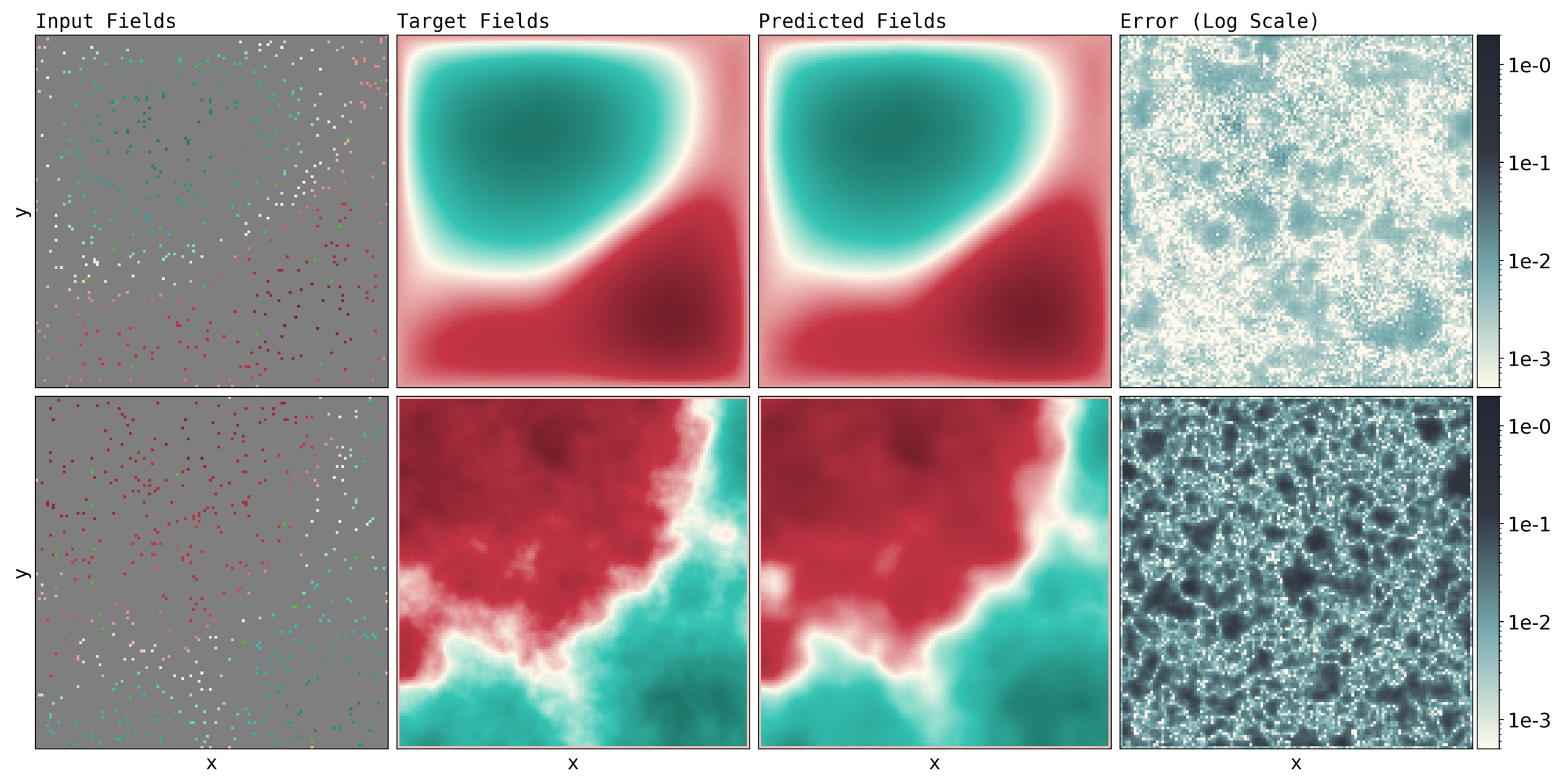}\\[1em]
    \includegraphics[width=0.45\textwidth]{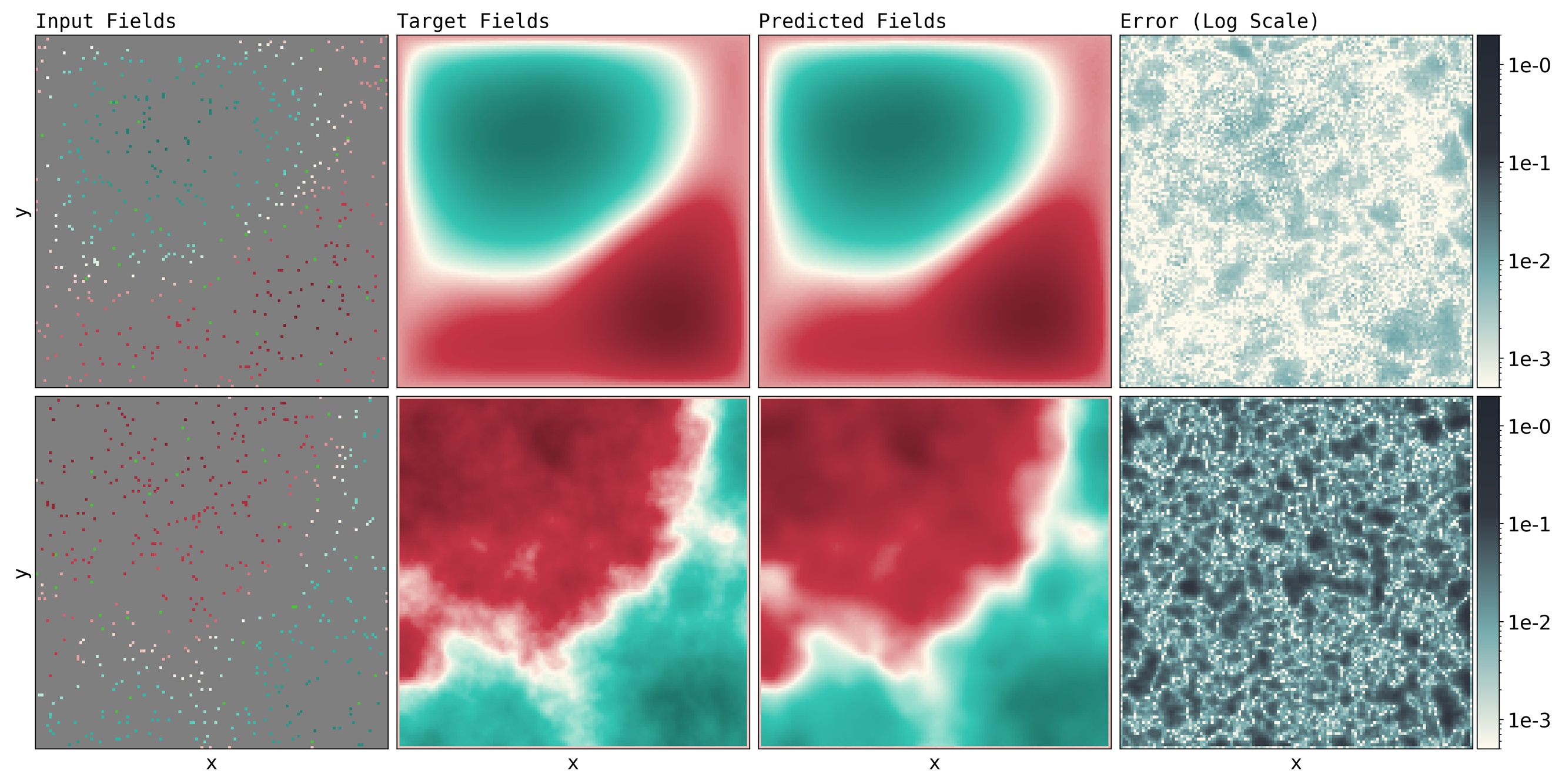}
    \includegraphics[width=0.45\textwidth]{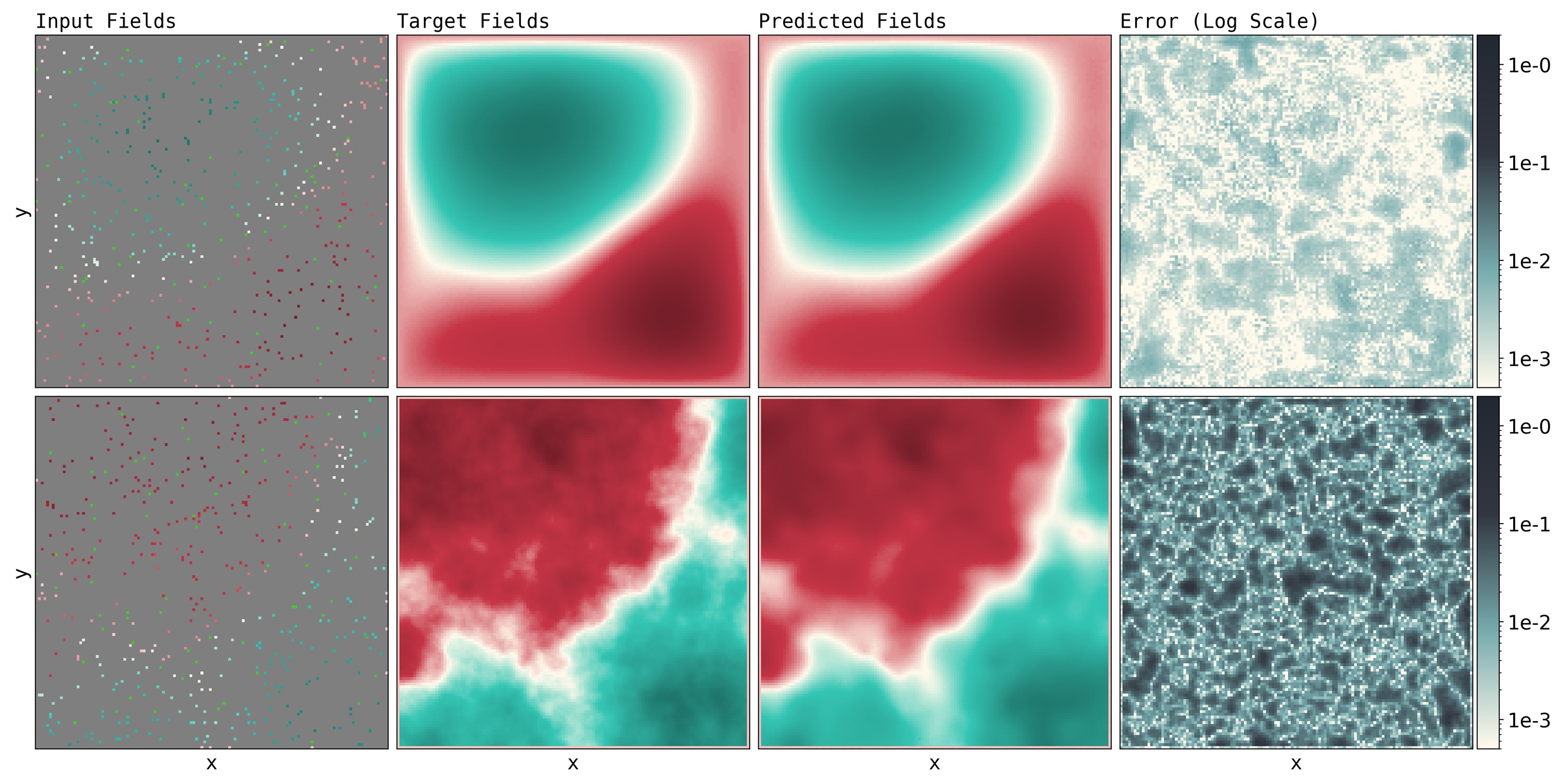}
    \caption{\textbf{Example of ``One-Point Transition" (\setlength{\fboxsep}{0pt}\colorbox{helmholtz!50}{Helmholtz}).} Each subplot shows an Ambient Flow reconstruction from partial observations (3\% uniformly random points), varying the number of additional masked (already-observed) points used during training (increasing across the grid; top-left: 0, which corresponds to naive training). \textbf{Within Each Subplot:} \textbf{Top:} Solution field. \textbf{Bottom:} Coefficient Field. \textbf{Left to Right:} Input Fields, Target Fields, Reconstructed Fields, Error (Log Scale). Within each input panel, gray pixels represent unobserved locations and green pixels represent masked measurements withheld from the model.}
\end{figure}

\begin{figure}[H]
    \centering
    \includegraphics[width=0.45\textwidth]{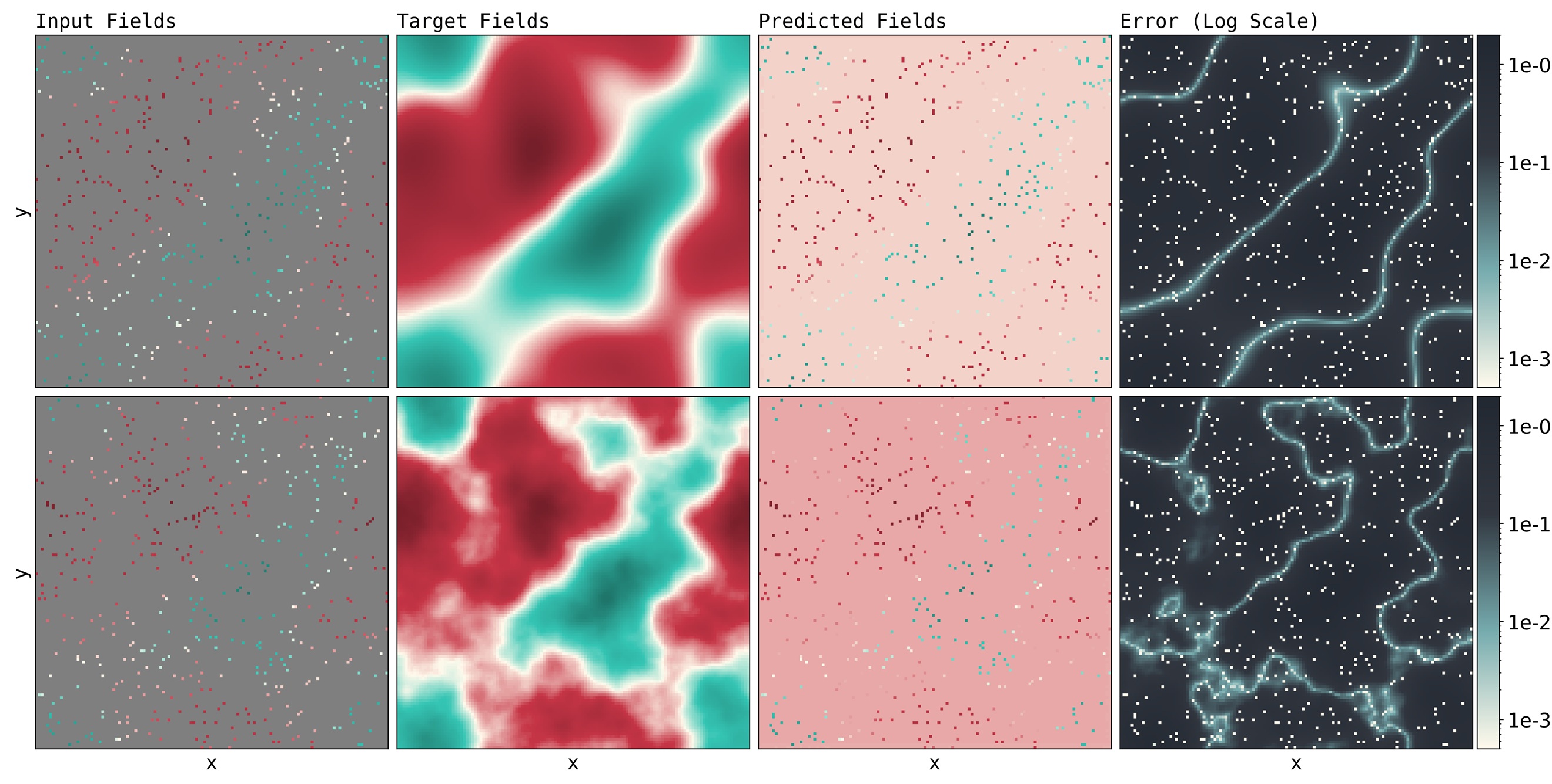}
    \includegraphics[width=0.45\textwidth]{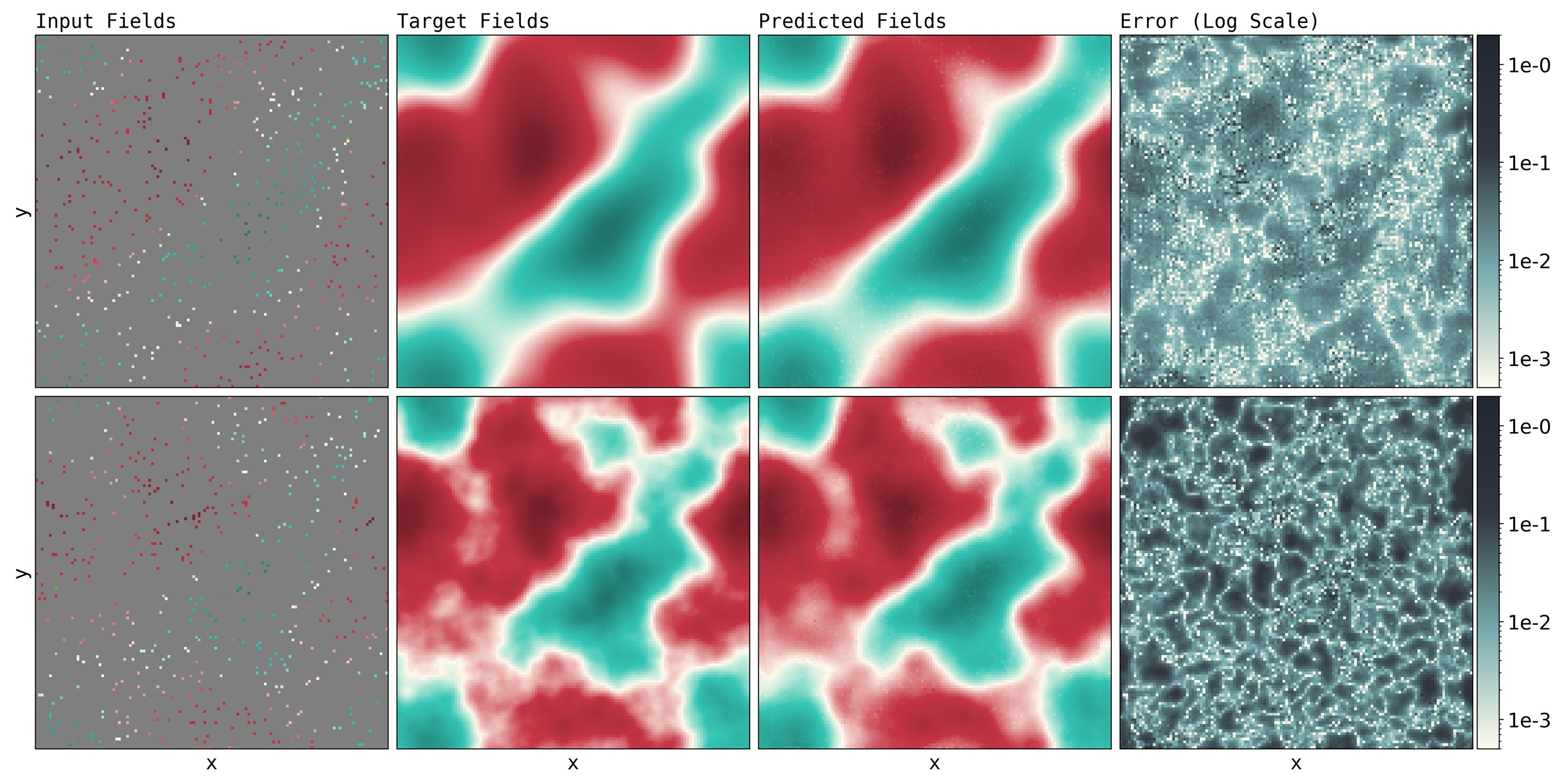}\\[1em]
    \includegraphics[width=0.45\textwidth]{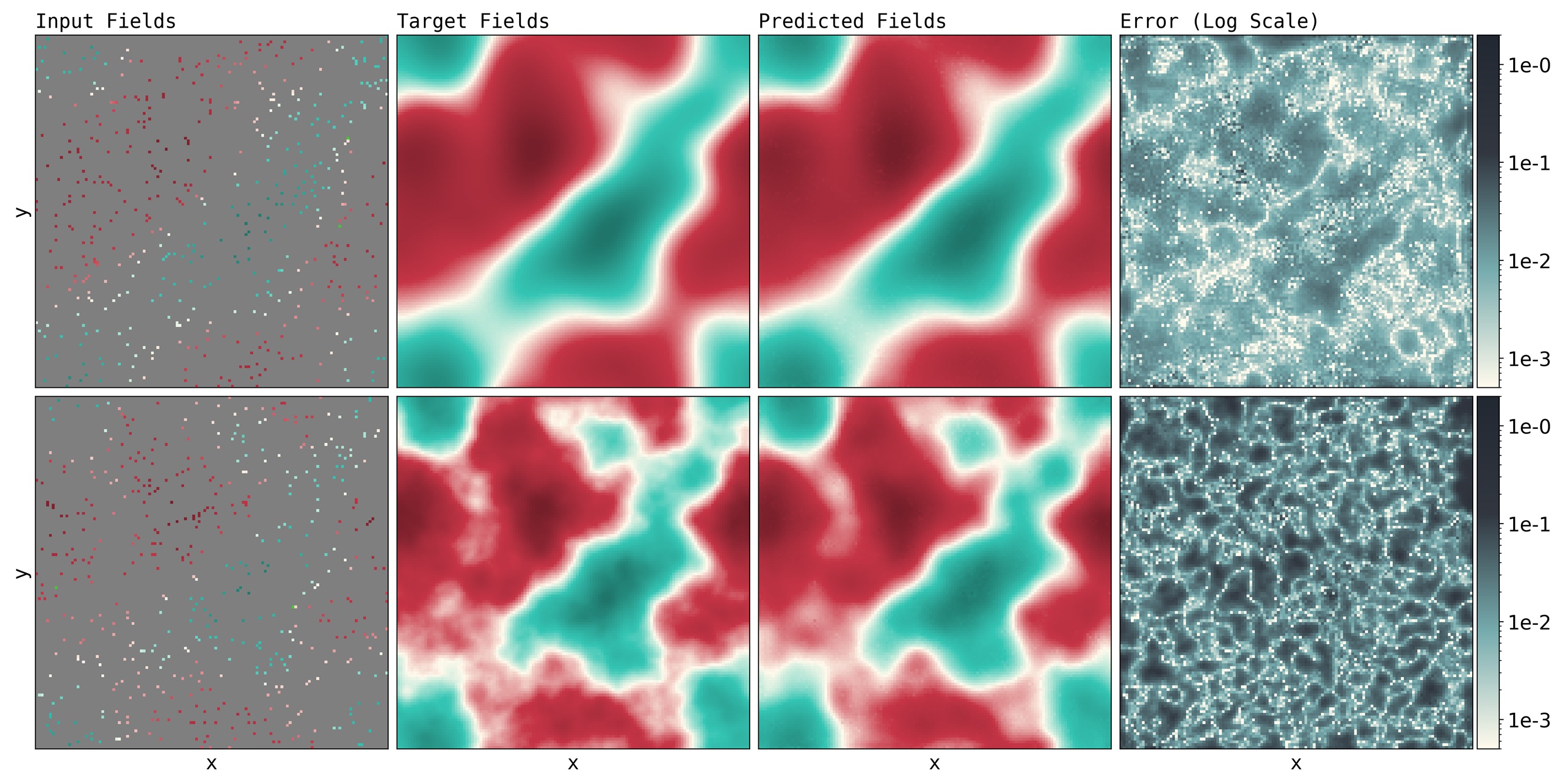}
    \includegraphics[width=0.45\textwidth]{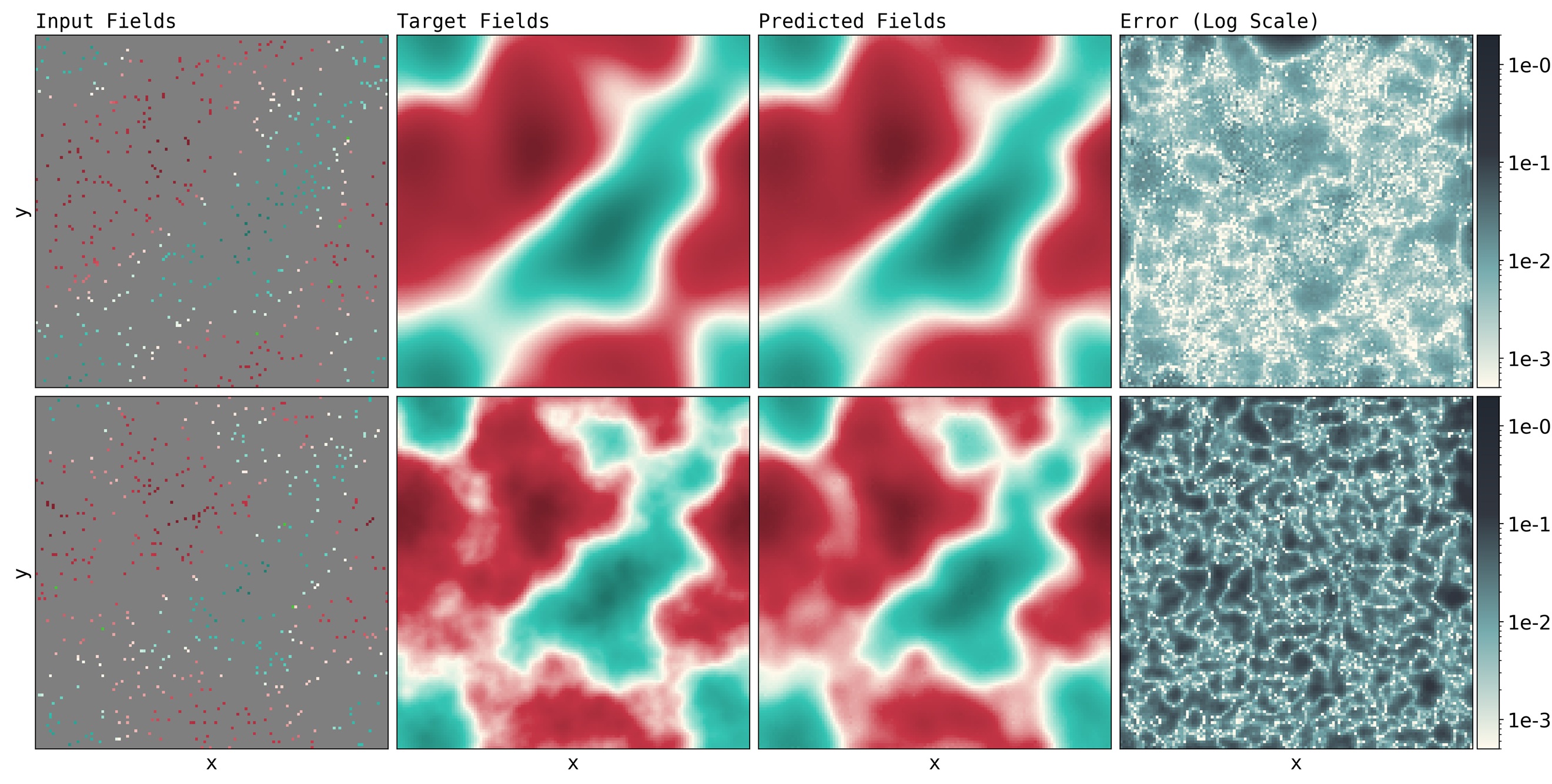}\\[1em]
    \includegraphics[width=0.45\textwidth]{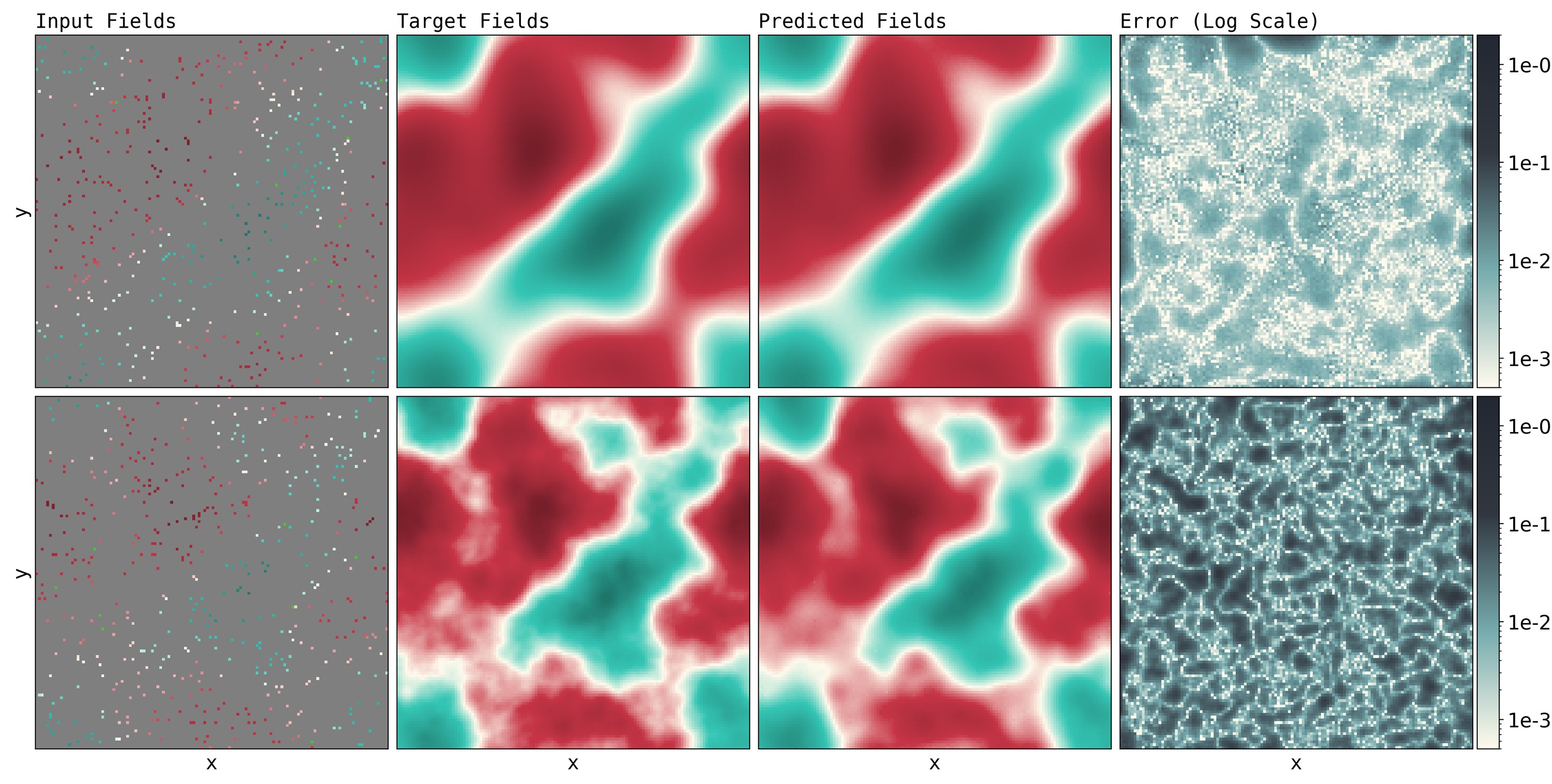}
    \includegraphics[width=0.45\textwidth]{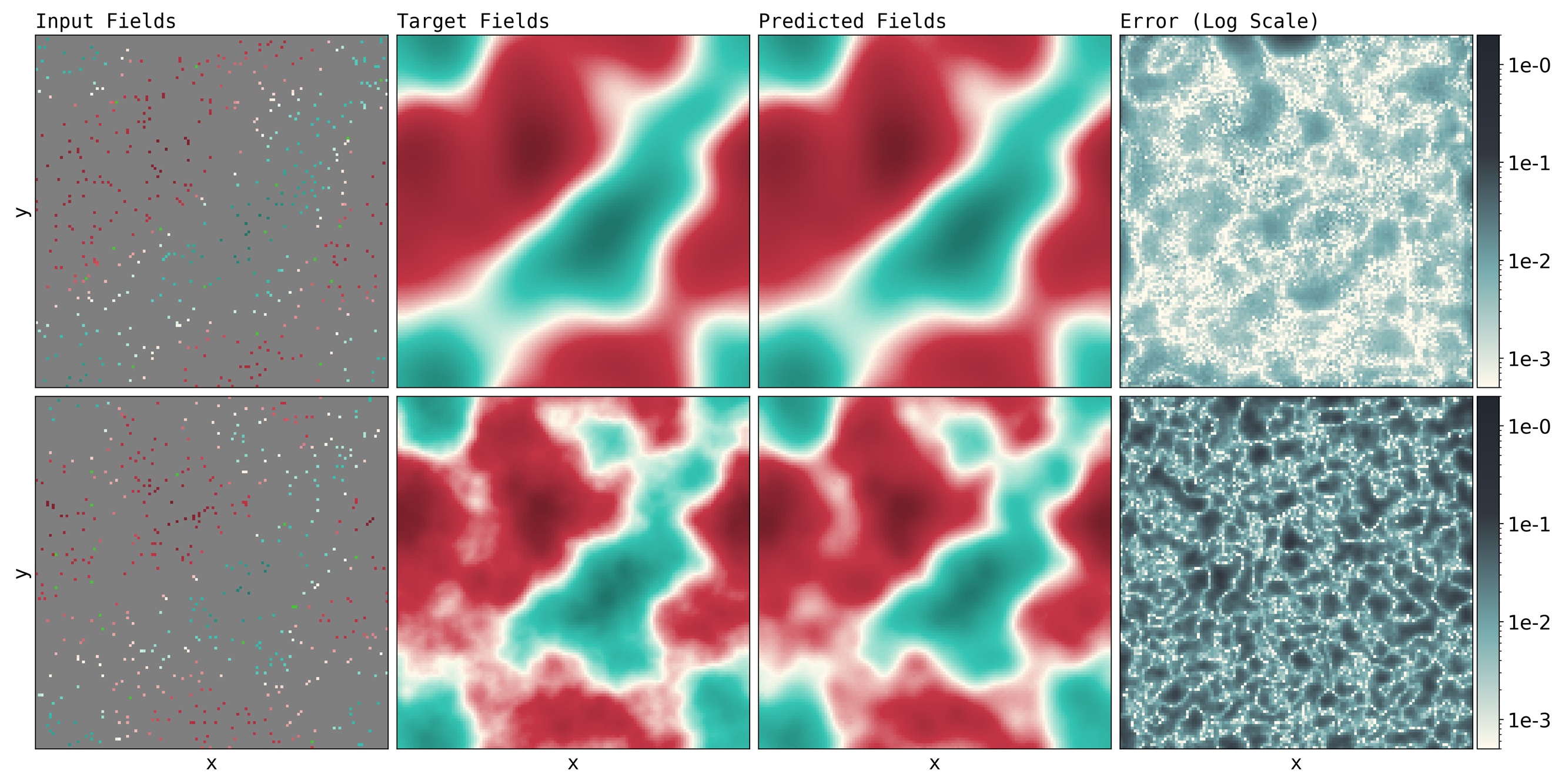}\\[1em]
    \includegraphics[width=0.45\textwidth]{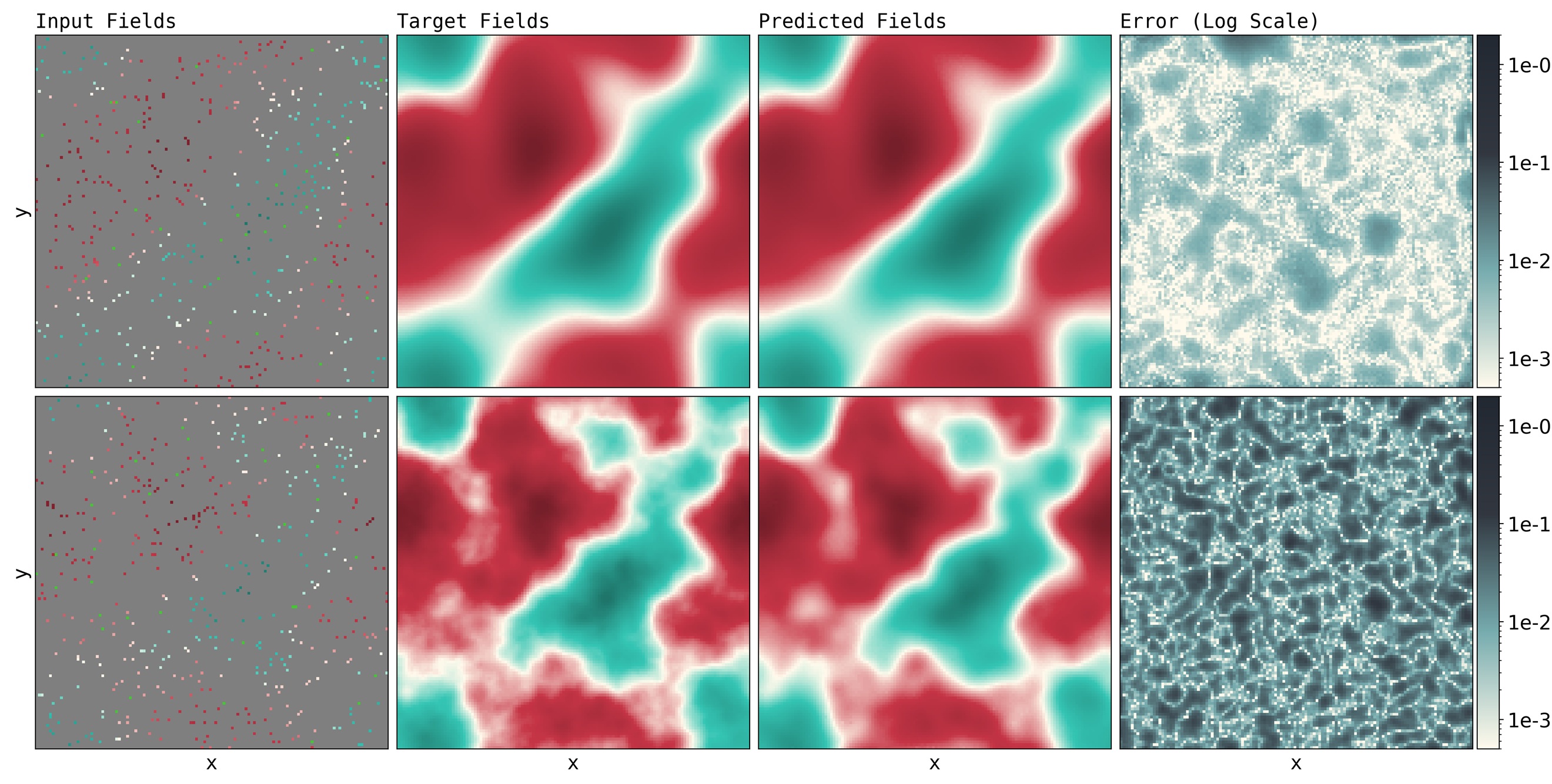}
    \includegraphics[width=0.45\textwidth]{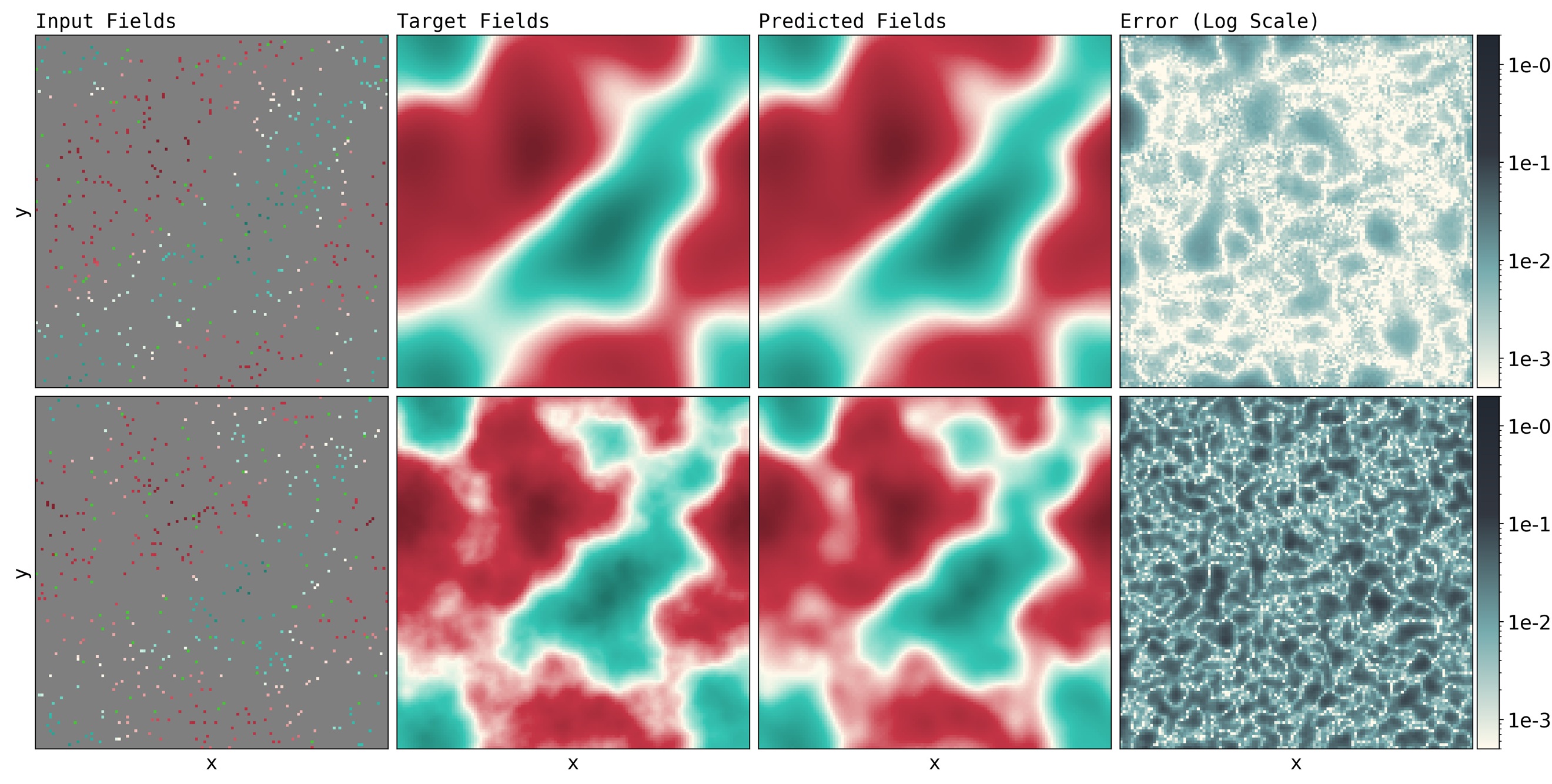}
    \caption{\textbf{Example of ``One-Point Transition" (\setlength{\fboxsep}{0pt}\colorbox{navierstokes!50}{Navier-Stokes}).} Each subplot shows an Ambient Flow reconstruction from partial observations (3\% uniformly random points), varying the number of additional masked (already-observed) points used during training (increasing across the grid; top-left: 0, which corresponds to naive training). \textbf{Within Each Subplot:} \textbf{Top:} Solution field. \textbf{Bottom:} Coefficient Field. \textbf{Left to Right:} Input Fields, Target Fields, Reconstructed Fields, Error (Log Scale). Within each input panel, gray pixels represent unobserved locations and green pixels represent masked measurements withheld from the model.}
\end{figure}

\begin{figure}[H]
    \centering
    \includegraphics[width=0.45\textwidth]{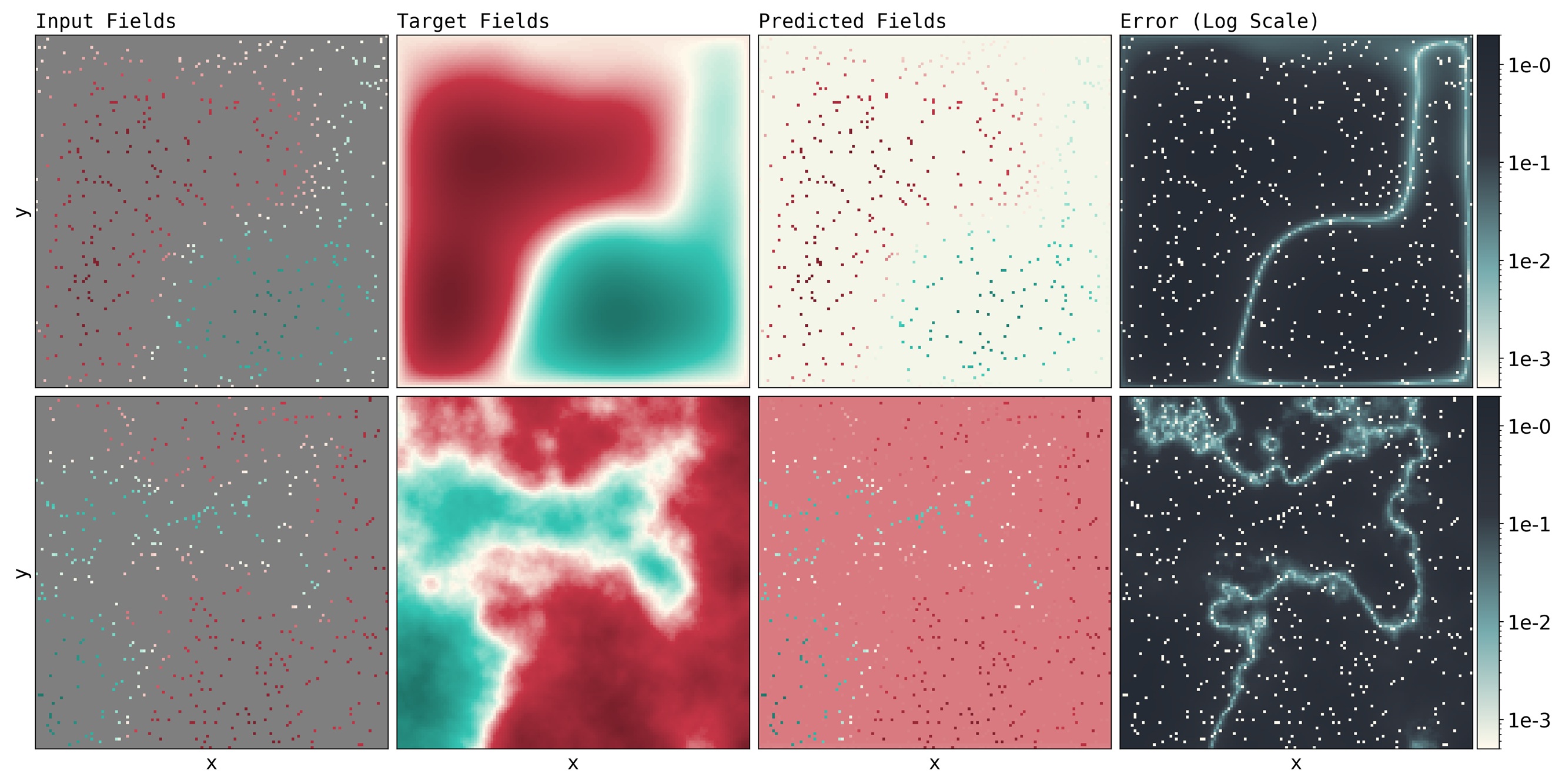}
    \includegraphics[width=0.45\textwidth]{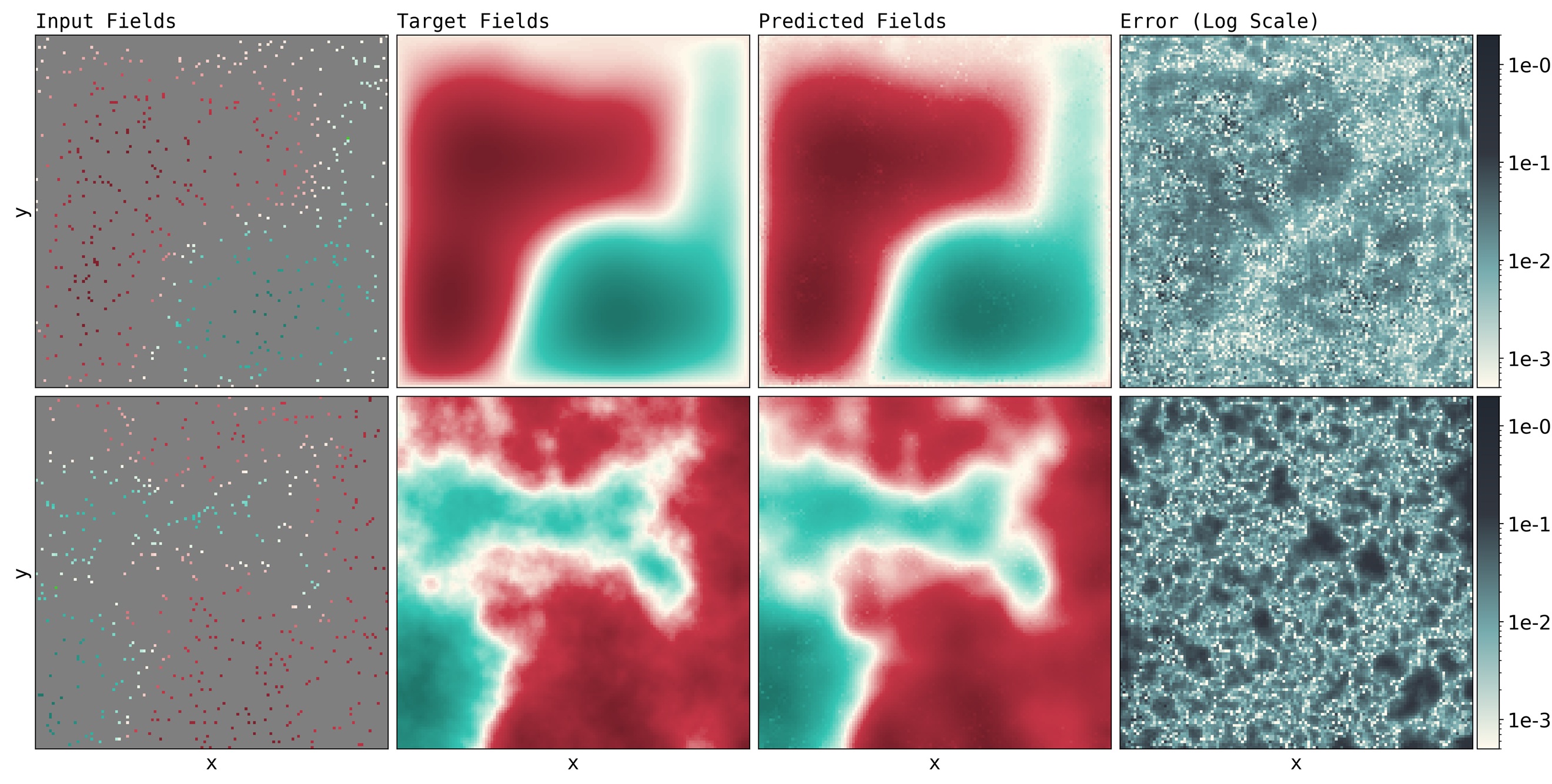}\\[1em]
    \includegraphics[width=0.45\textwidth]{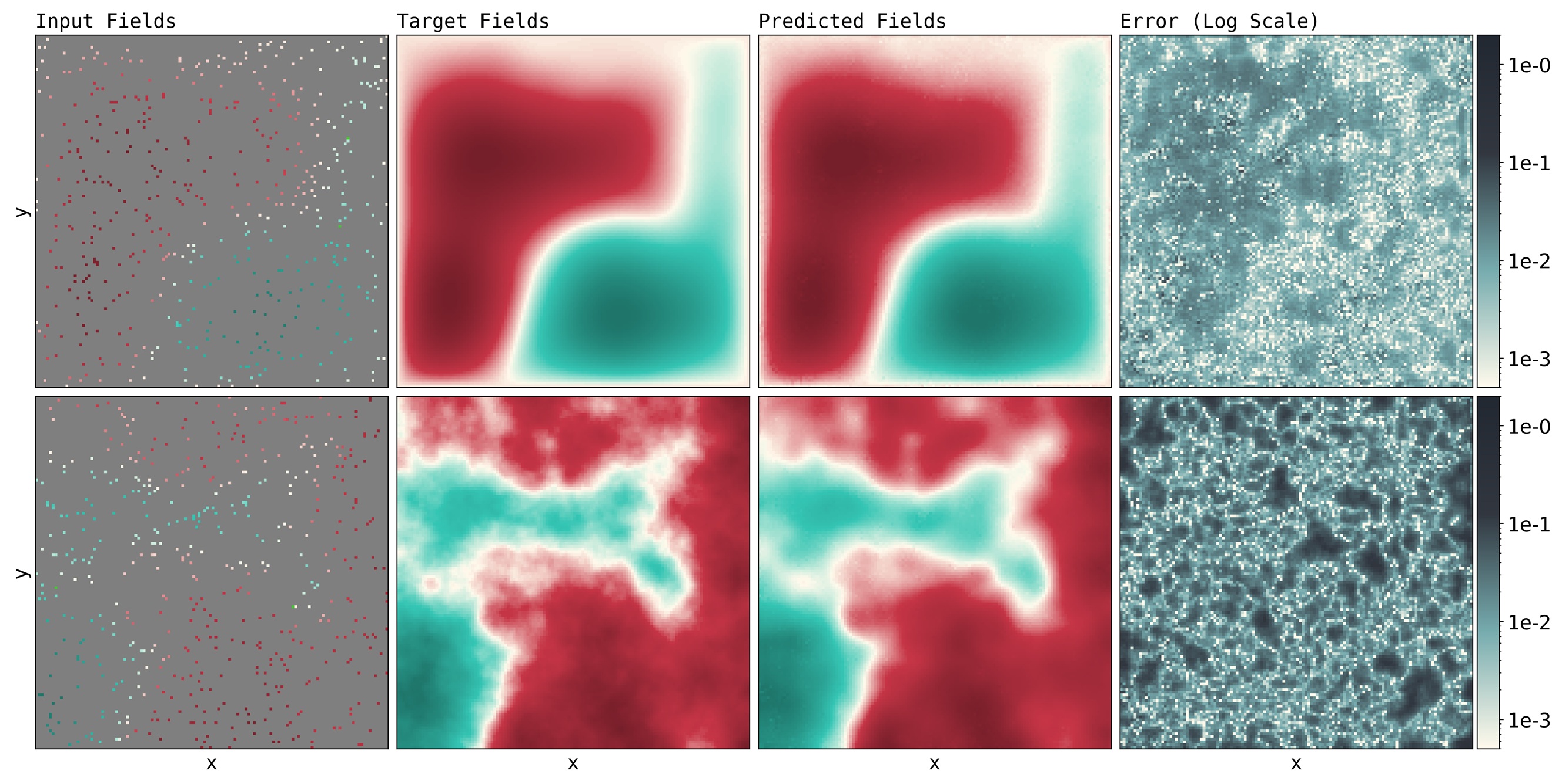}
    \includegraphics[width=0.45\textwidth]{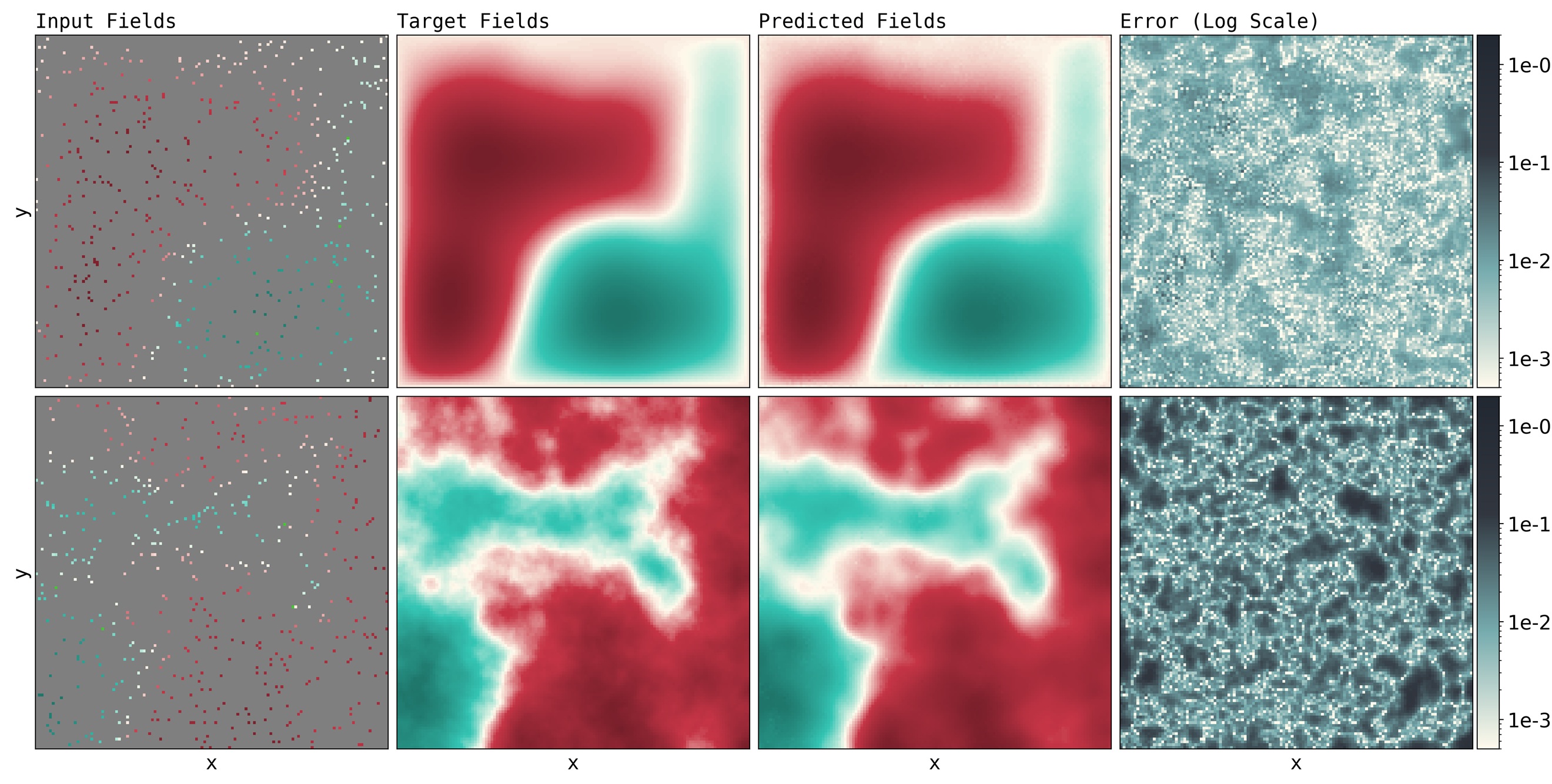}\\[1em]
    \includegraphics[width=0.45\textwidth]{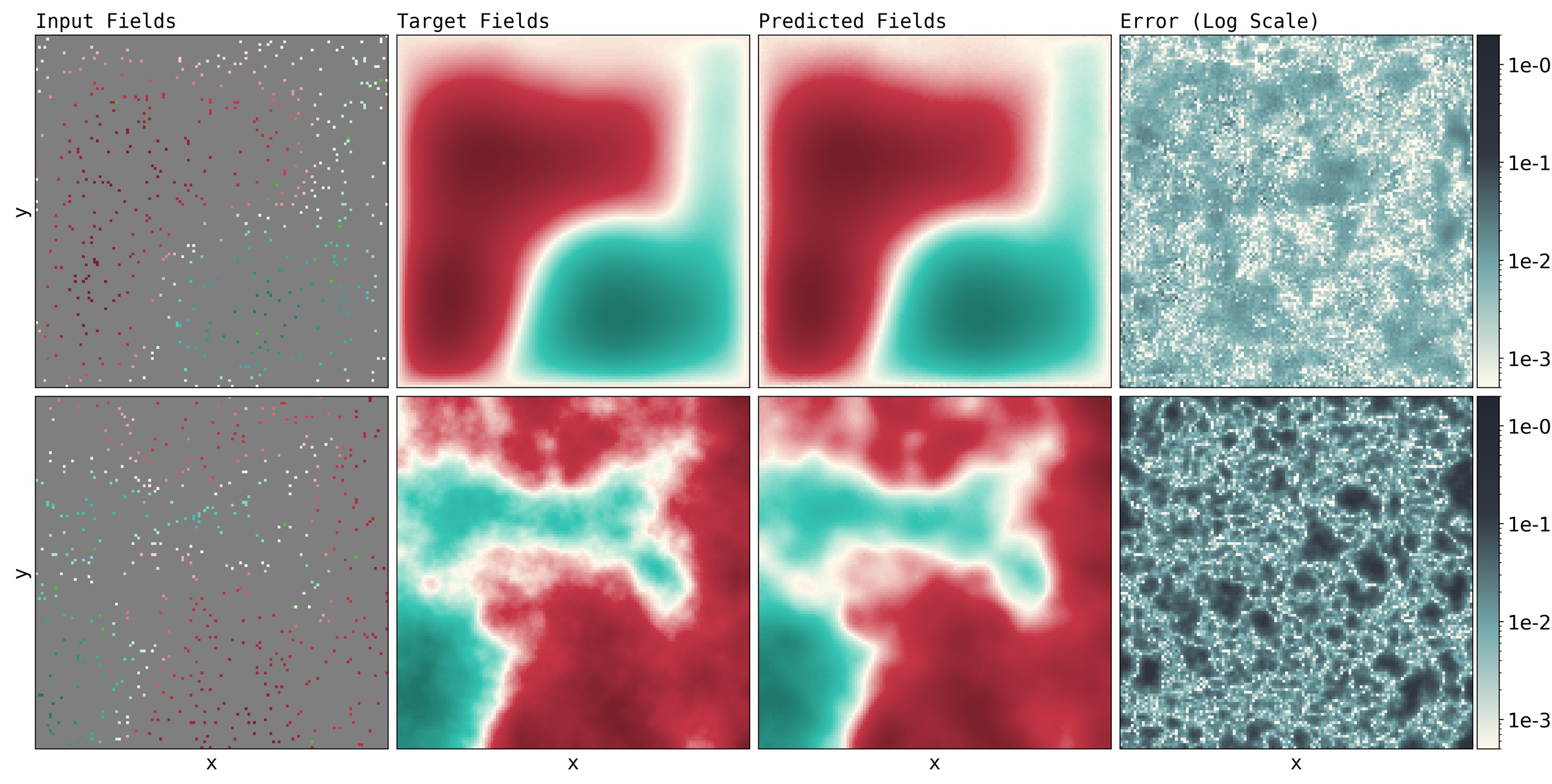}
    \includegraphics[width=0.45\textwidth]{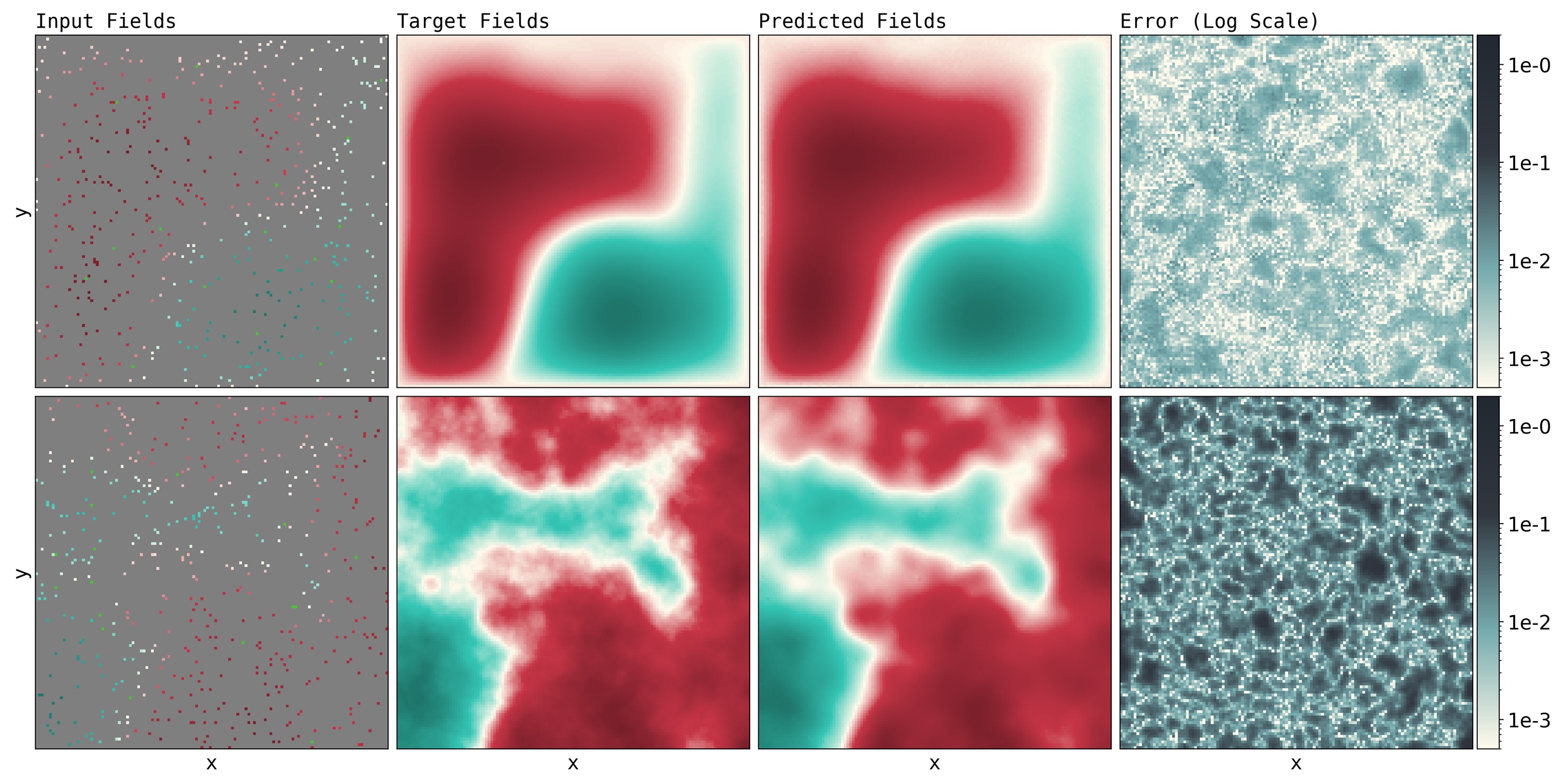}\\[1em]
    \includegraphics[width=0.45\textwidth]{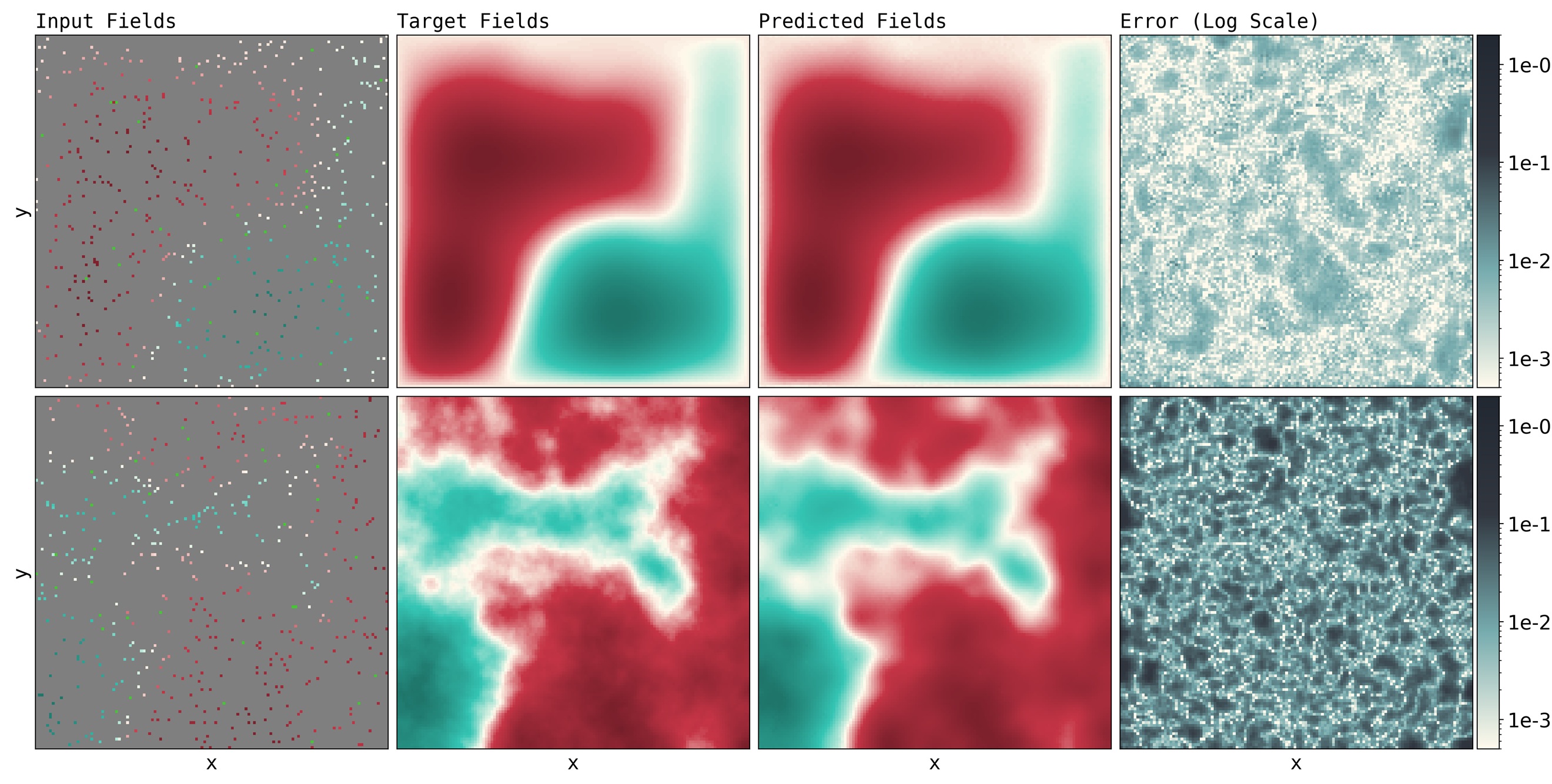}
    \includegraphics[width=0.45\textwidth]{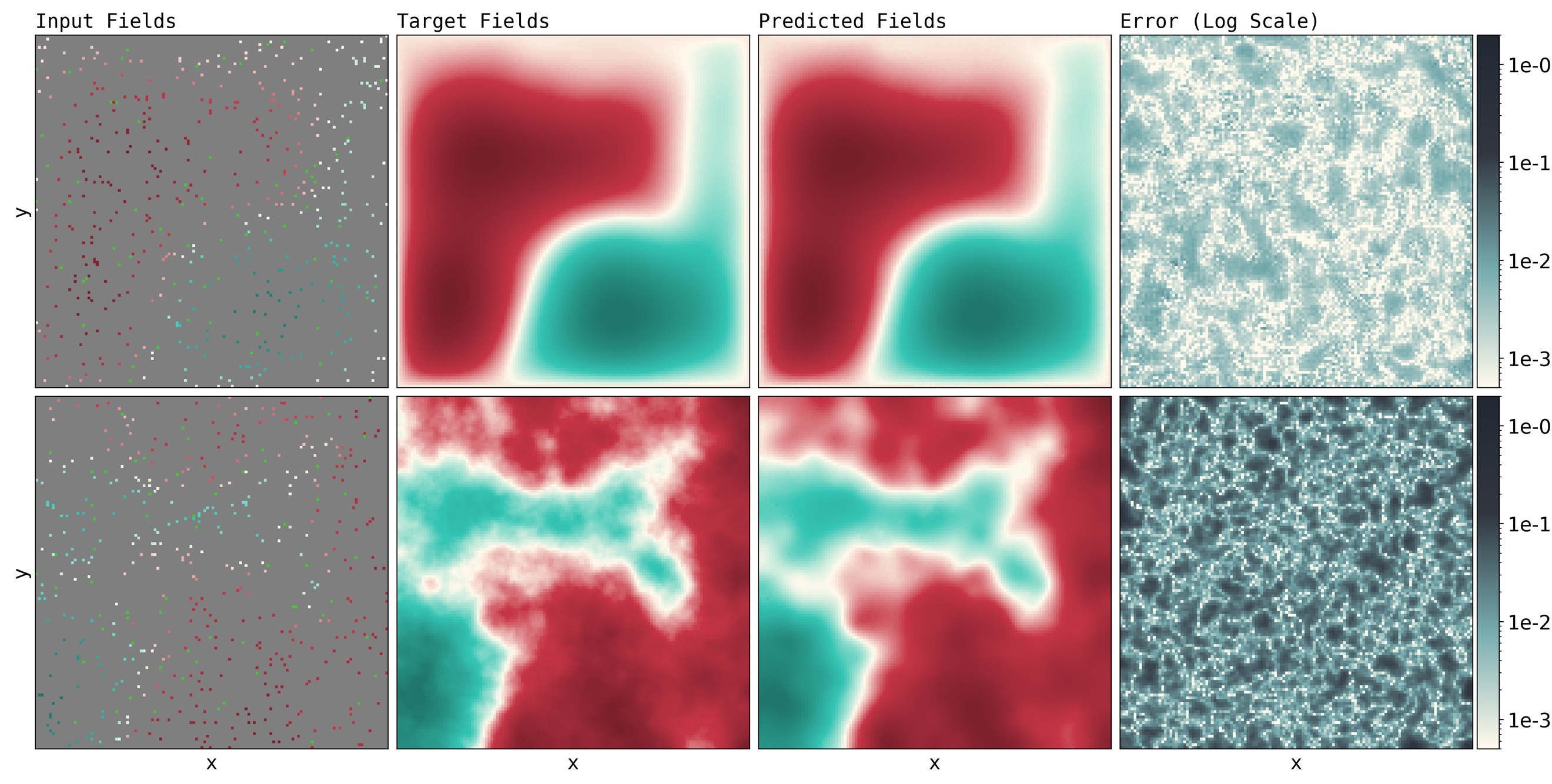}
    \caption{\textbf{Example of ``One-Point Transition" (\setlength{\fboxsep}{0pt}\colorbox{poisson!50}{Poisson}).} Each subplot shows an Ambient Flow reconstruction from partial observations (3\% uniformly random points), varying the number of additional masked (already-observed) points used during training (increasing across the grid; top-left: 0, which corresponds to naive training). \textbf{Within Each Subplot:} \textbf{Top:} Solution field. \textbf{Bottom:} Coefficient Field. \textbf{Left to Right:} Input Fields, Target Fields, Reconstructed Fields, Error (Log Scale). Within each input panel, gray pixels represent unobserved locations and green pixels represent masked measurements withheld from the model.}
\end{figure}

\newpage

\section{Robustness Across Patterns and Super-Resolution}
\label{robustness_across_patterns}

\begin{figure}[H]
    \centering
    \includegraphics[width=0.75\linewidth]{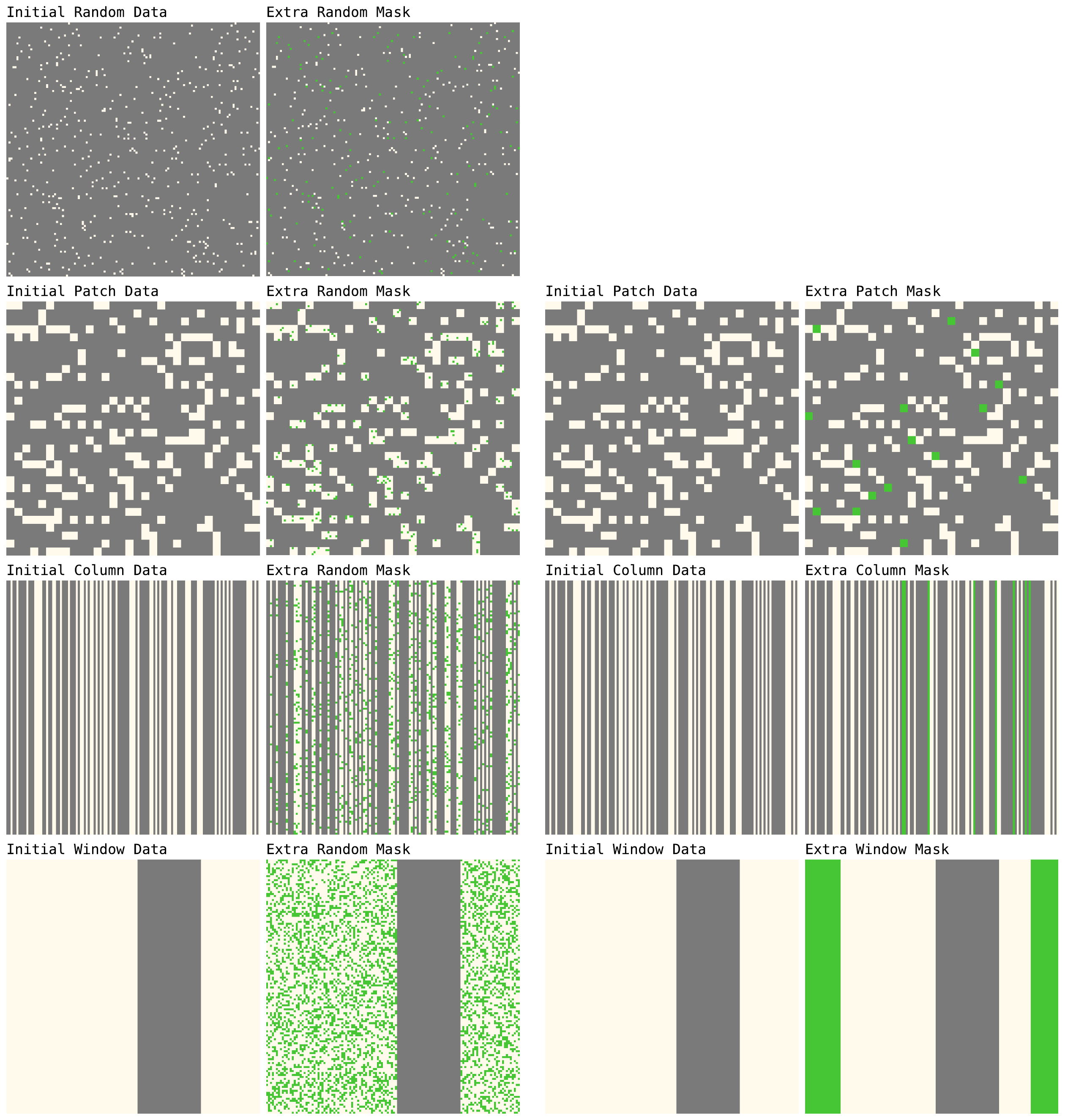}
    \caption{\textbf{Measurement patterns and extra masking operators.} Visualization of measurement patterns $A$ and additional masking operators $B$ used during Ambient Physics training. Additional masking $B$ may match the measurement pattern (pattern-matched) or differ from it (mismatched); when mismatched, stronger additional masking is typically required to achieve similar training effects. \textbf{Top to bottom:} Random, Patch, Column, Window. Gray pixels represent unobserved locations and green pixels represent masked measurements withheld from the model.}
    \label{fig:placeholder}
\end{figure}

\newpage

\begin{figure}[H]
    \centering
    \includegraphics[width=0.45\textwidth]{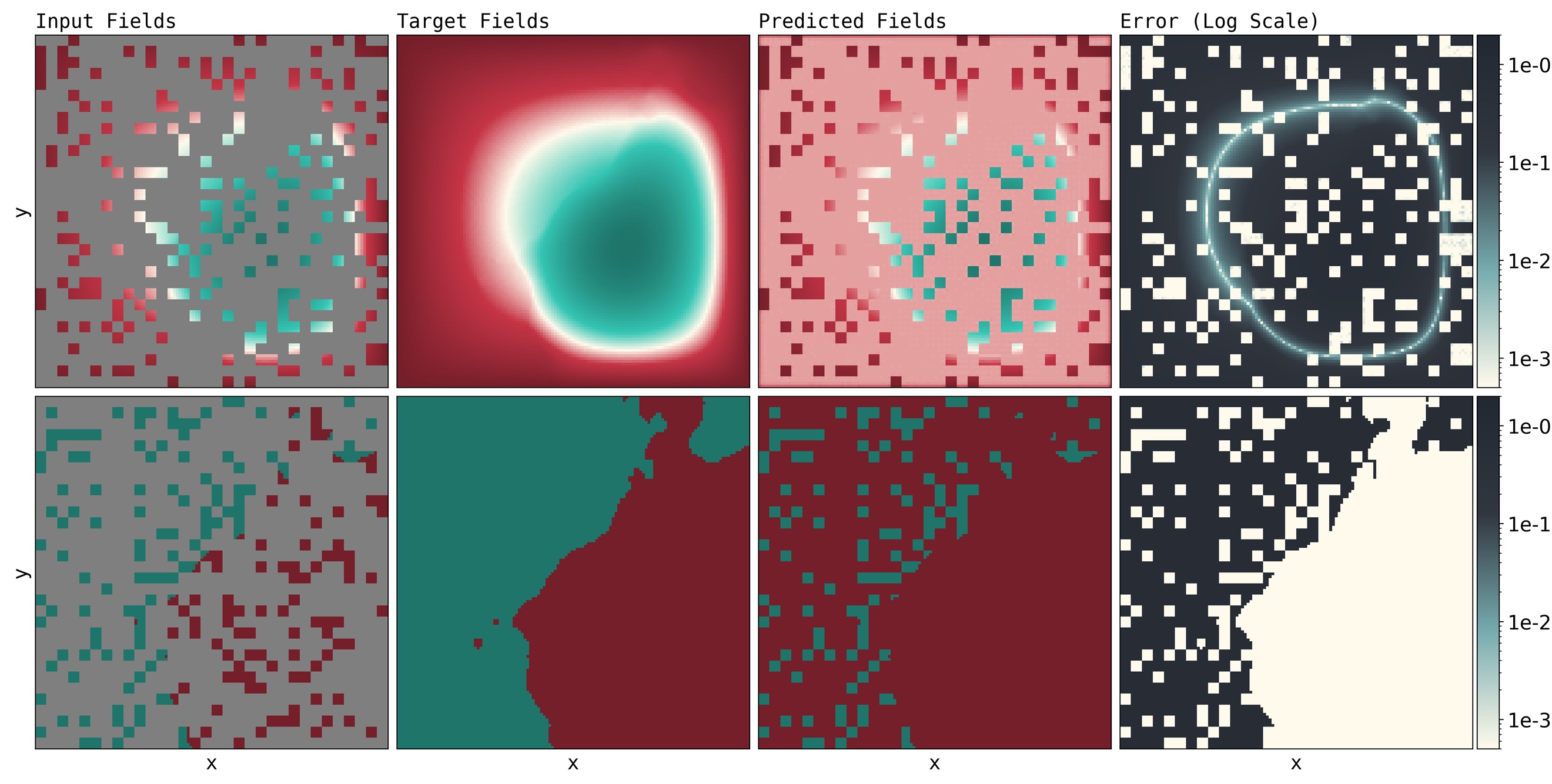}
    \includegraphics[width=0.45\textwidth]{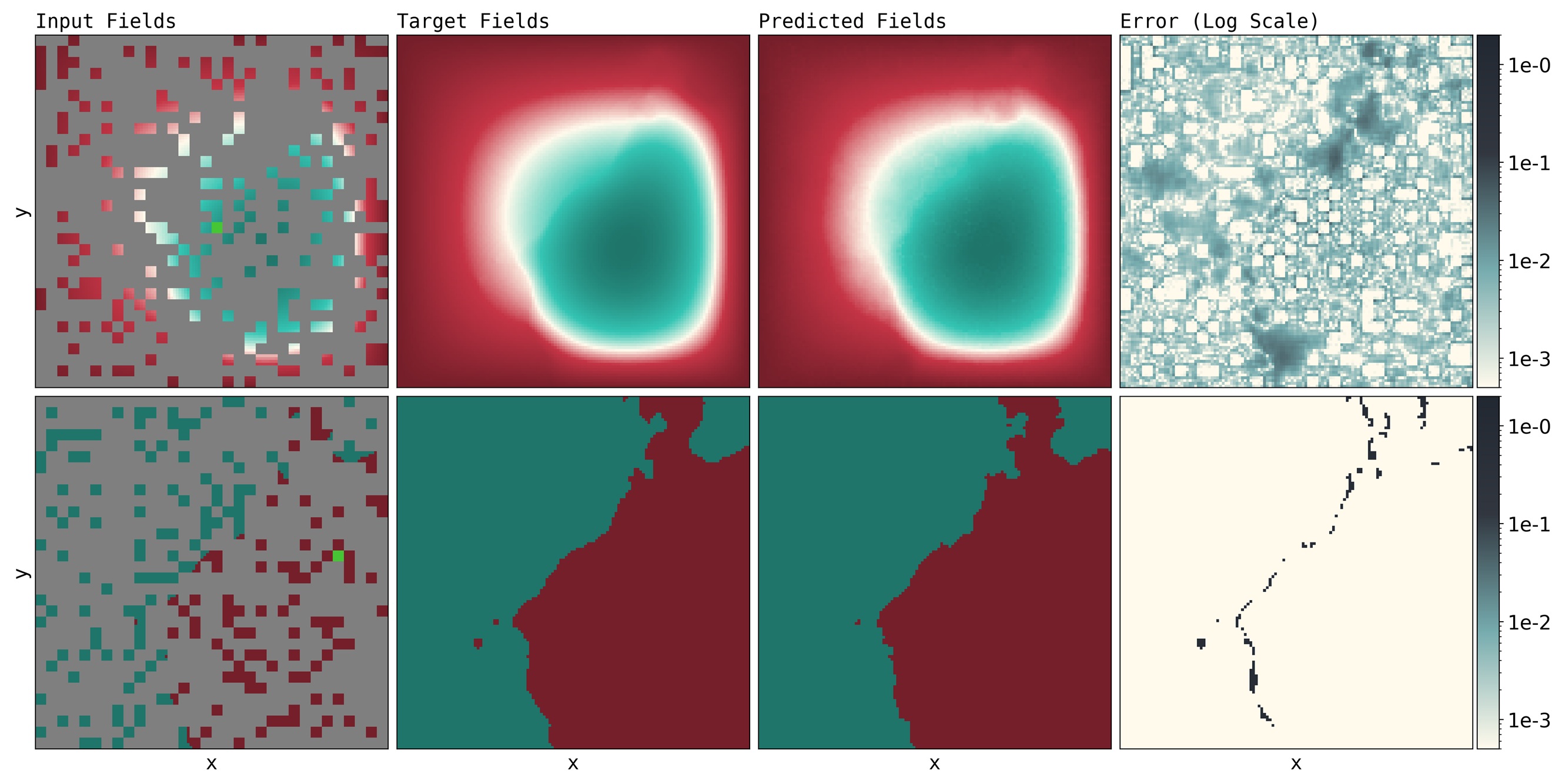}\\[1em]
    \includegraphics[width=0.45\textwidth]{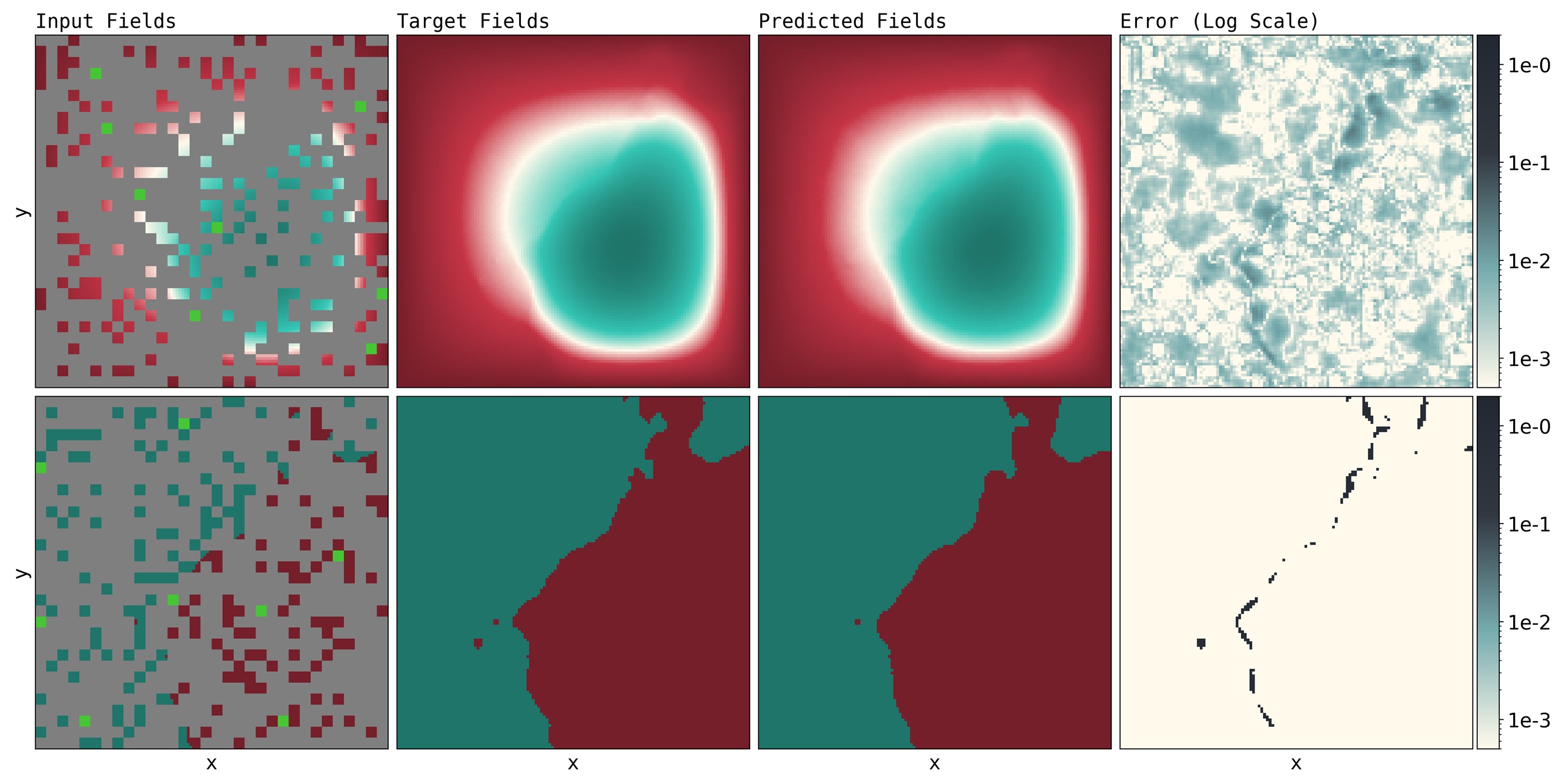}
    \includegraphics[width=0.45\textwidth]{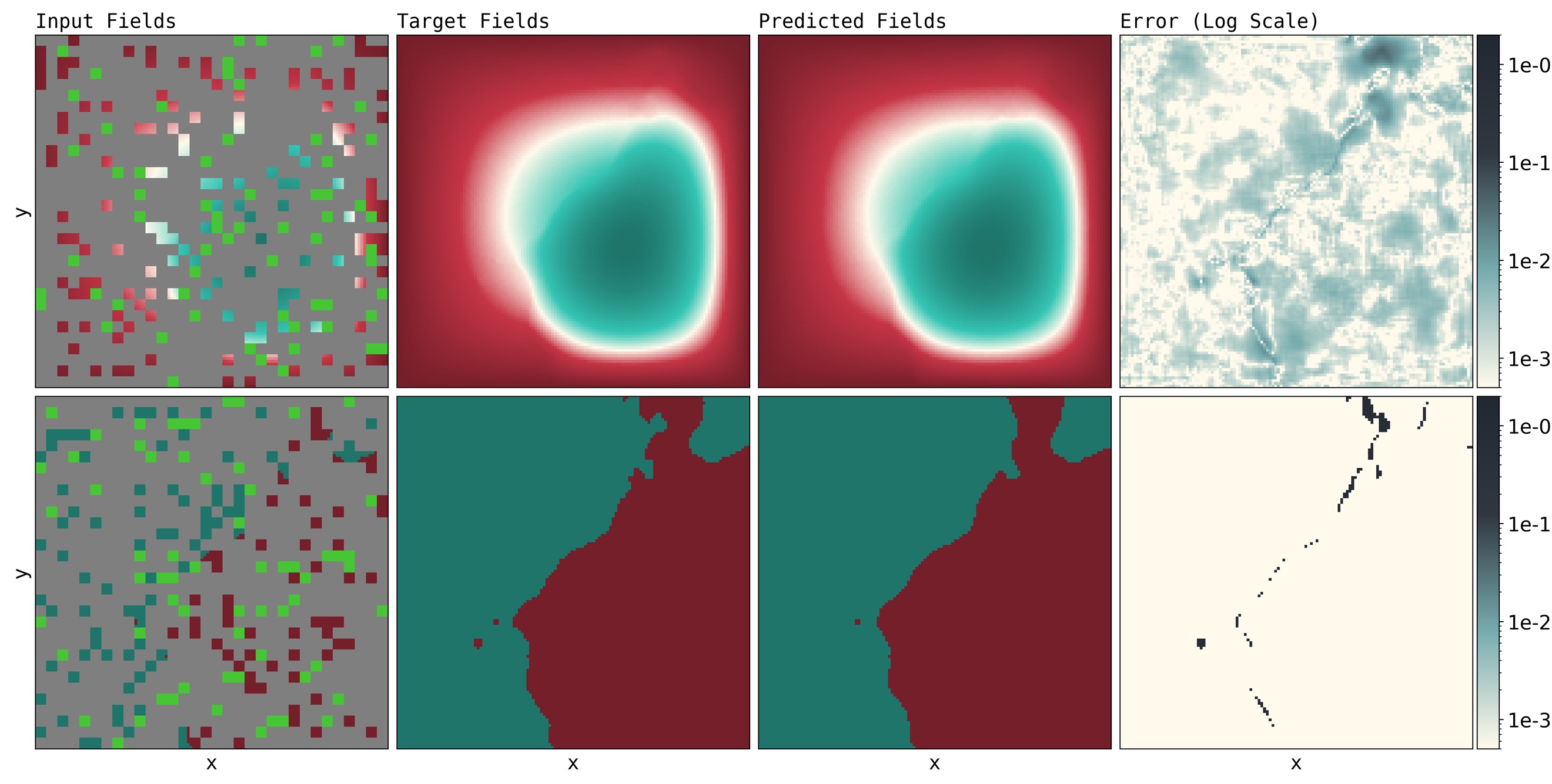}
    \caption{\textbf{Example of Patch Measurements (\setlength{\fboxsep}{0pt}\colorbox{darcyflow!50}{Darcy Flow}).} Each subplot shows an Ambient Flow reconstruction from partial observations (20\% uniformly random patches), varying the number of additional masked (already-observed) patches used during training (increasing across the grid; top-left: 0, which corresponds to naive training). \textbf{Within Each Subplot:} \textbf{Top:} Solution field. \textbf{Bottom:} Coefficient Field. \textbf{Left to Right:} Input Fields, Target Fields, Reconstructed Fields, Error (Log Scale). Within each input panel, gray pixels represent unobserved locations and green pixels represent masked measurements withheld from the model.}
\end{figure}

\begin{figure}[H]
    \centering
    \includegraphics[width=0.45\textwidth]{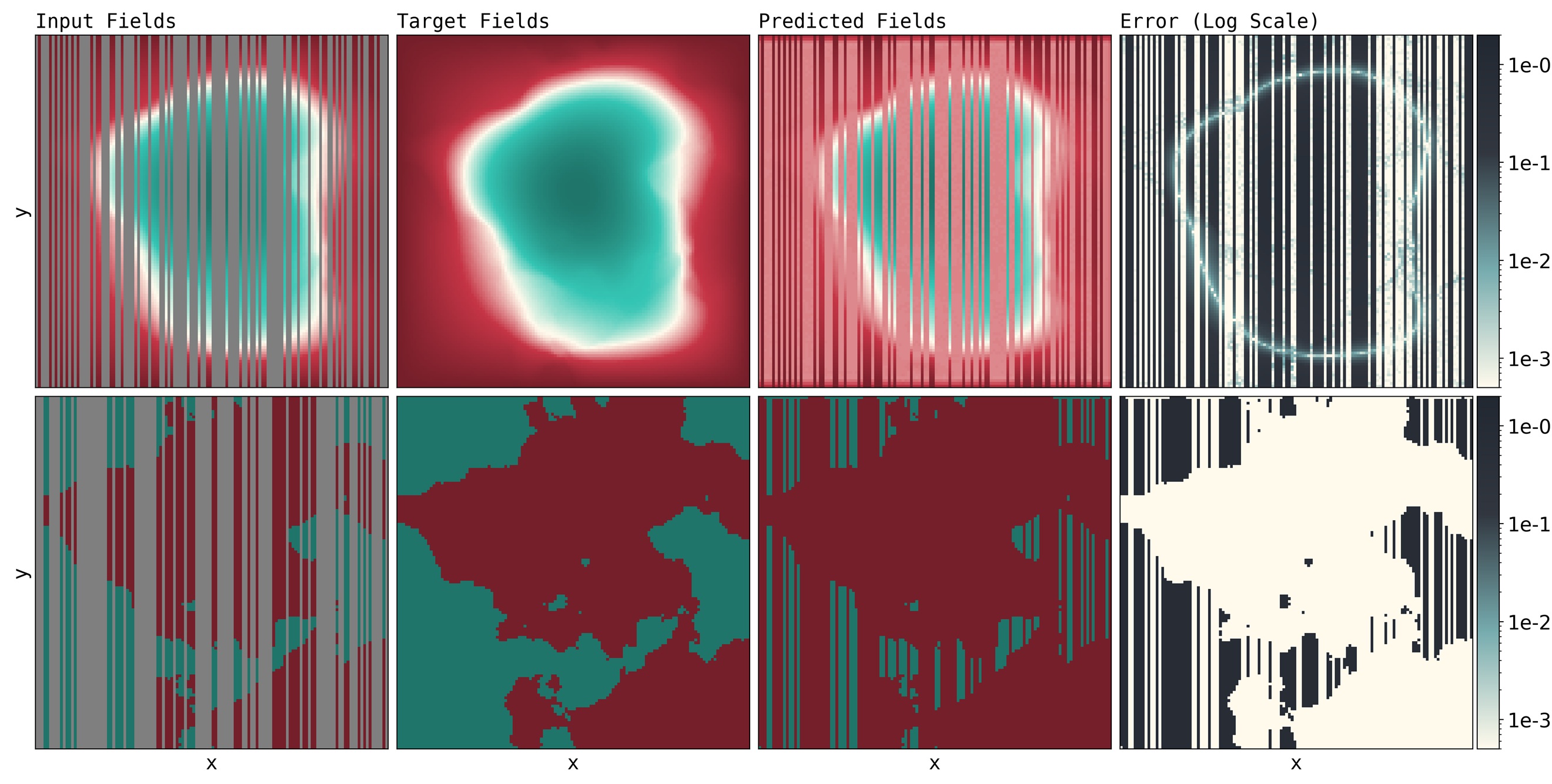}
    \includegraphics[width=0.45\textwidth]{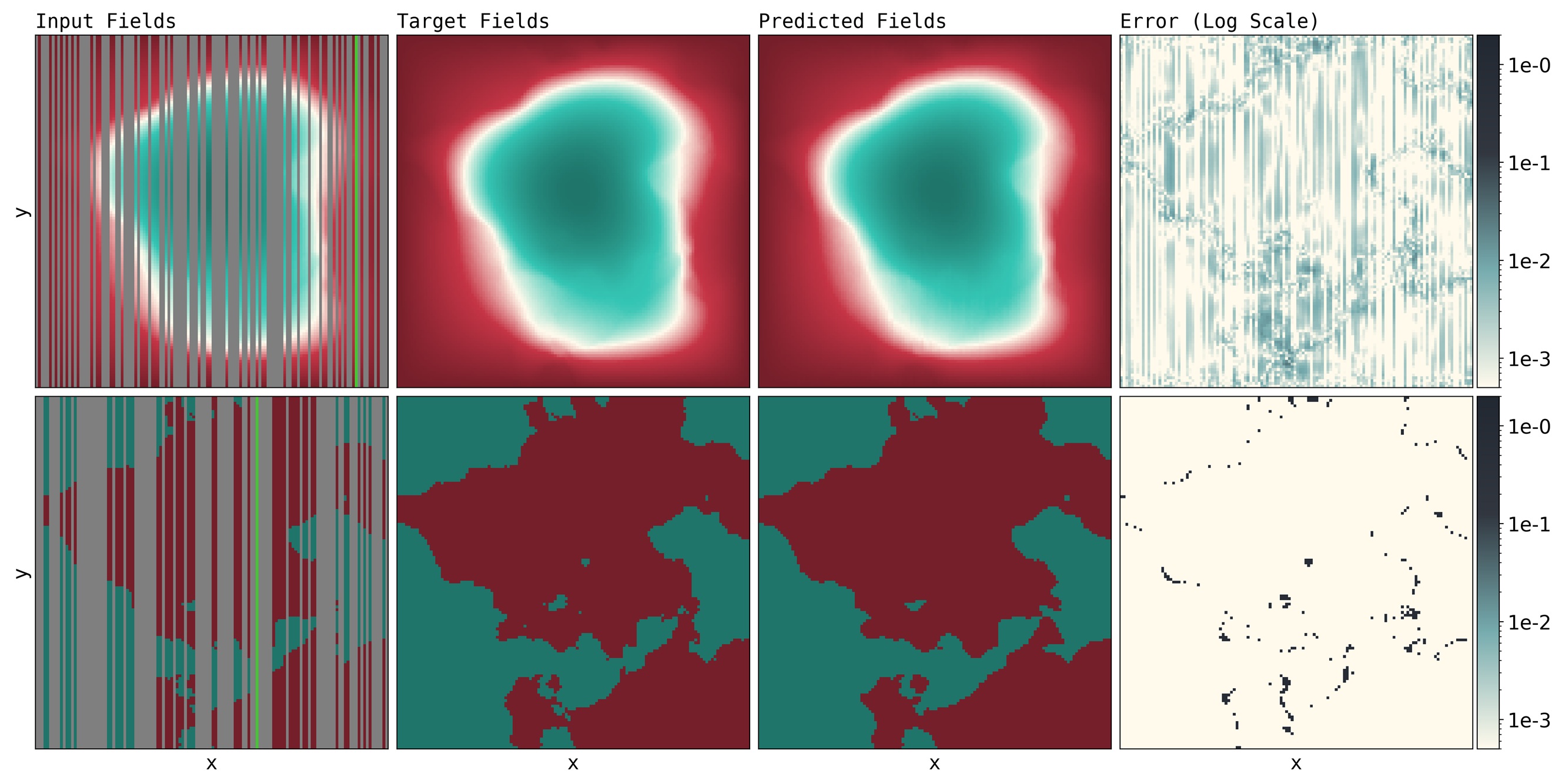}\\[1em]
    \includegraphics[width=0.45\textwidth]{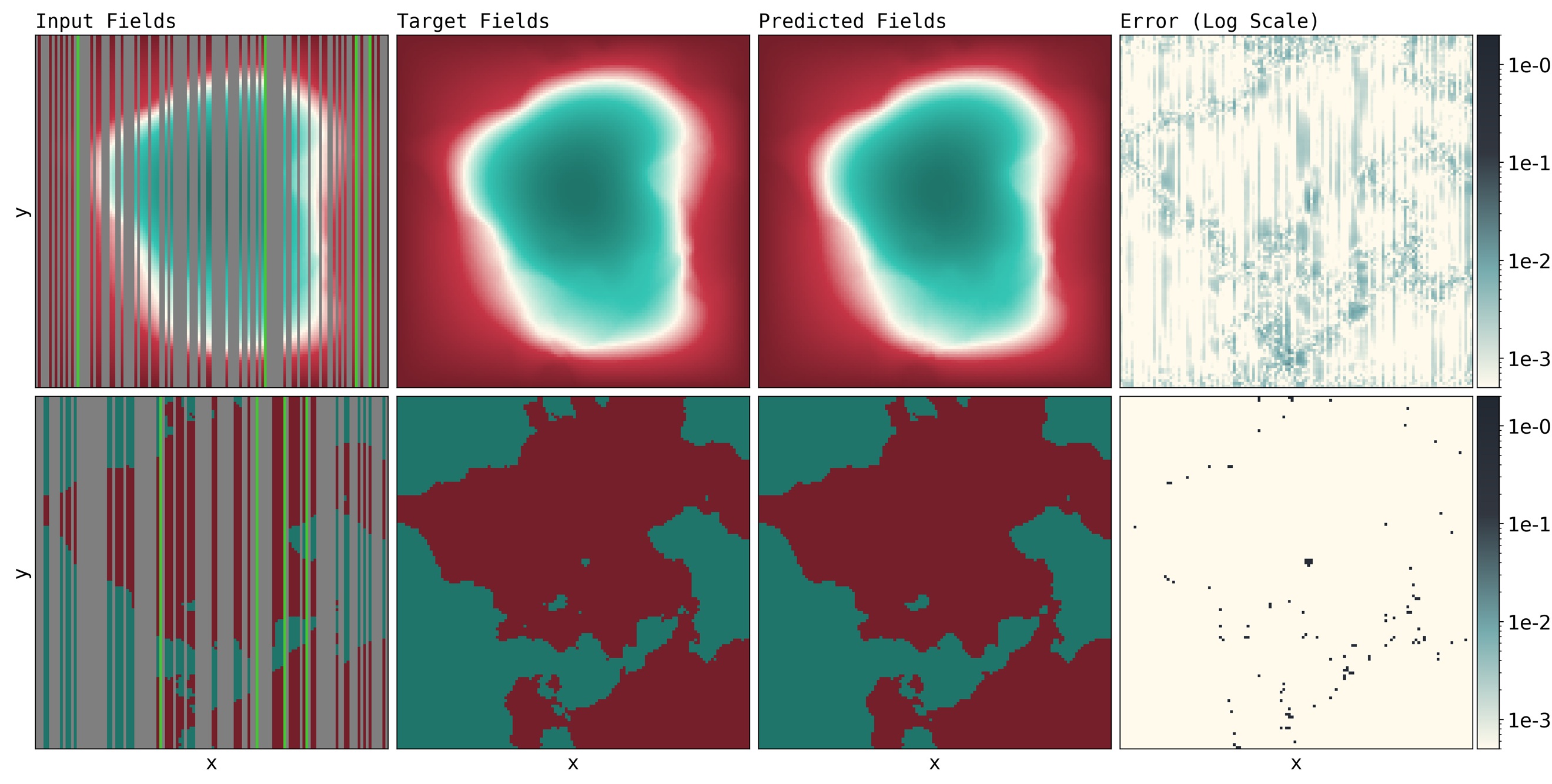}
    \includegraphics[width=0.45\textwidth]{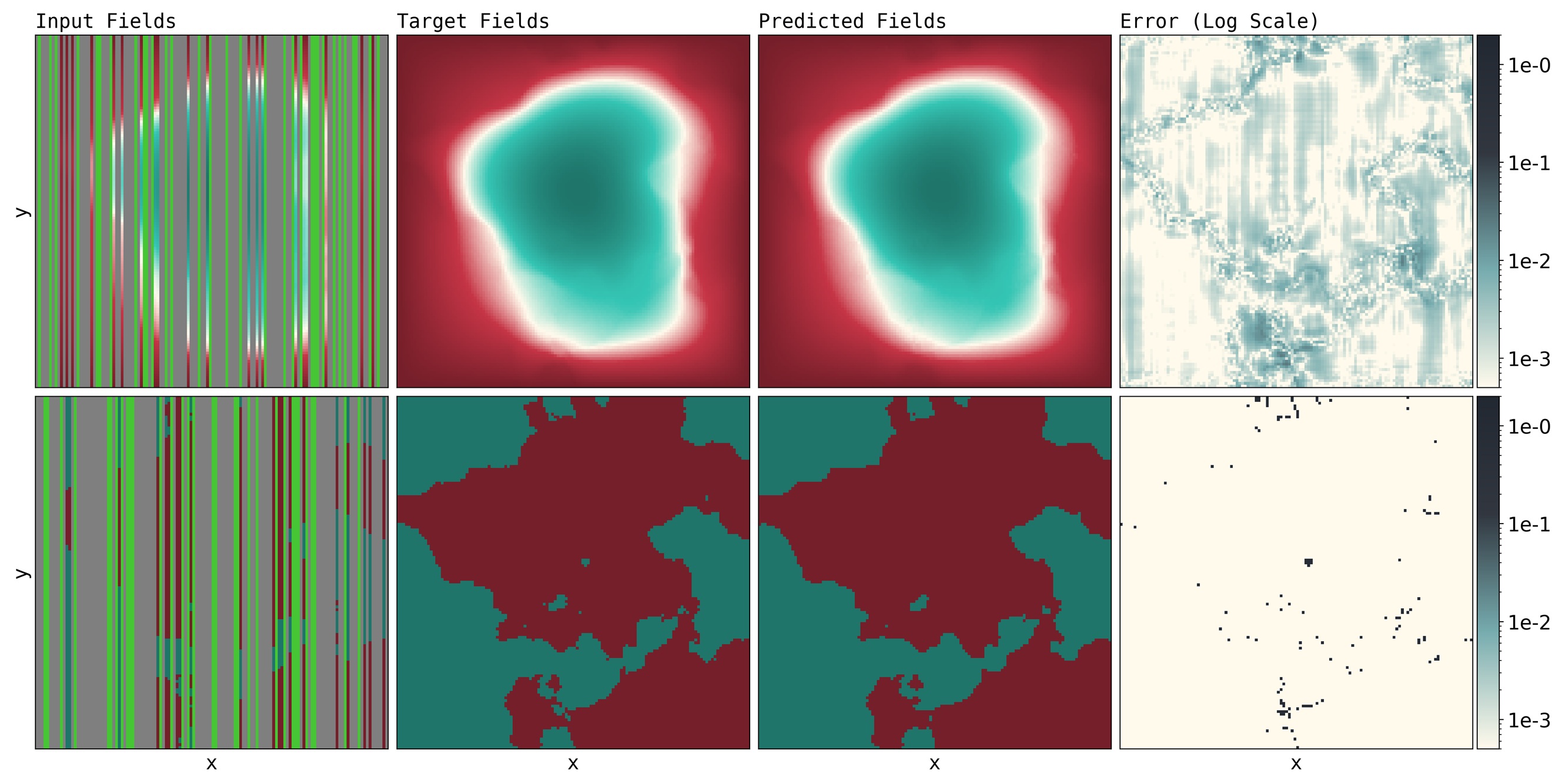}
    \caption{\textbf{Example of Column Measurements (\setlength{\fboxsep}{0pt}\colorbox{darcyflow!50}{Darcy Flow}).} Each subplot shows an Ambient Flow reconstruction from partial observations (40\% uniformly random columns), varying the number of additional masked (already-observed) columns used during training (increasing across the grid; top-left: 0, which corresponds to naive training). \textbf{Within Each Subplot:} \textbf{Top:} Solution field. \textbf{Bottom:} Coefficient Field. \textbf{Left to Right:} Input Fields, Target Fields, Reconstructed Fields, Error (Log Scale). Within each input panel, gray pixels represent unobserved locations and green pixels represent masked measurements withheld from the model.}
\end{figure}

\begin{figure}[H]
    \centering
    \includegraphics[width=0.45\textwidth]{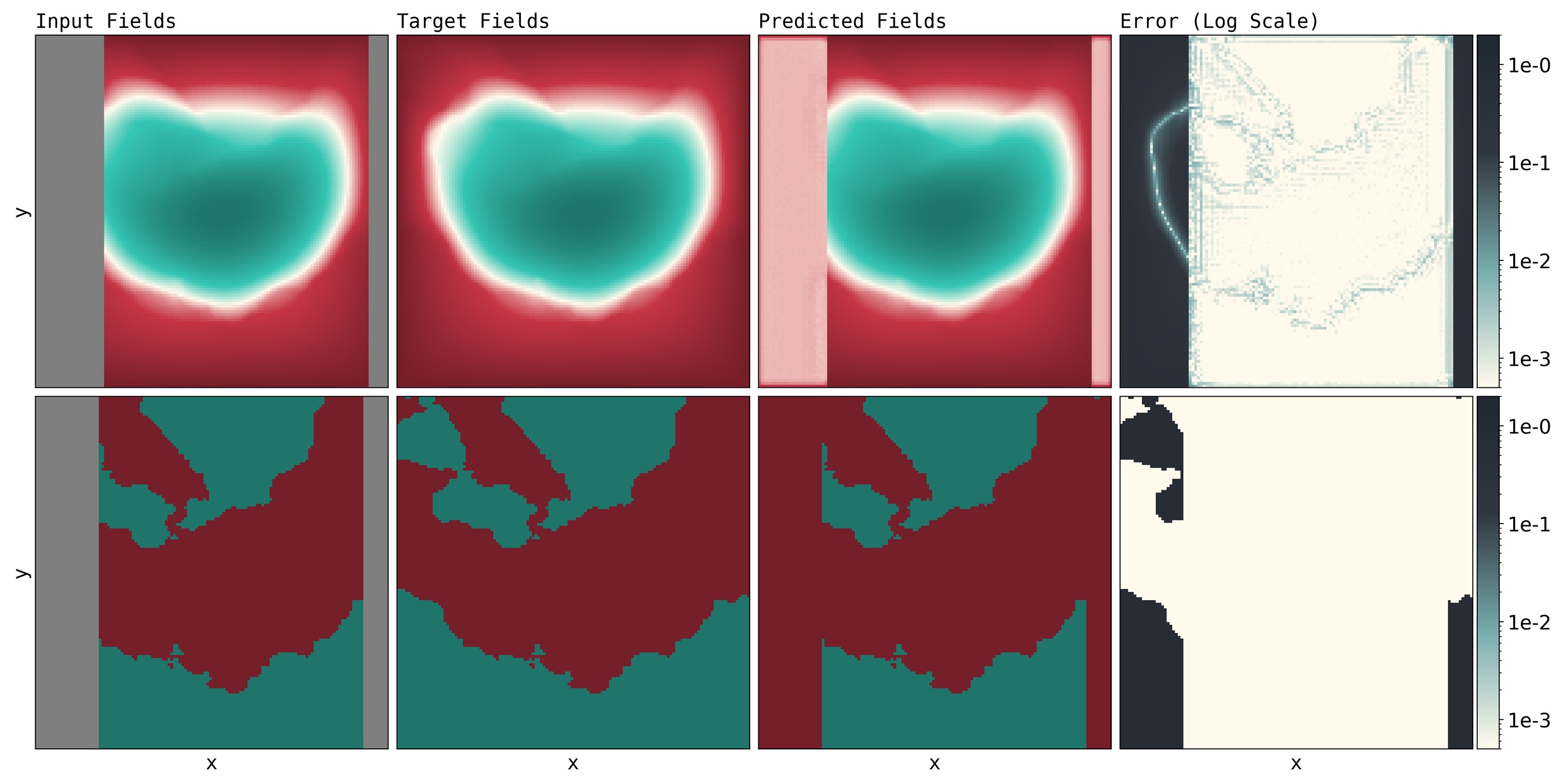}
    \includegraphics[width=0.45\textwidth]{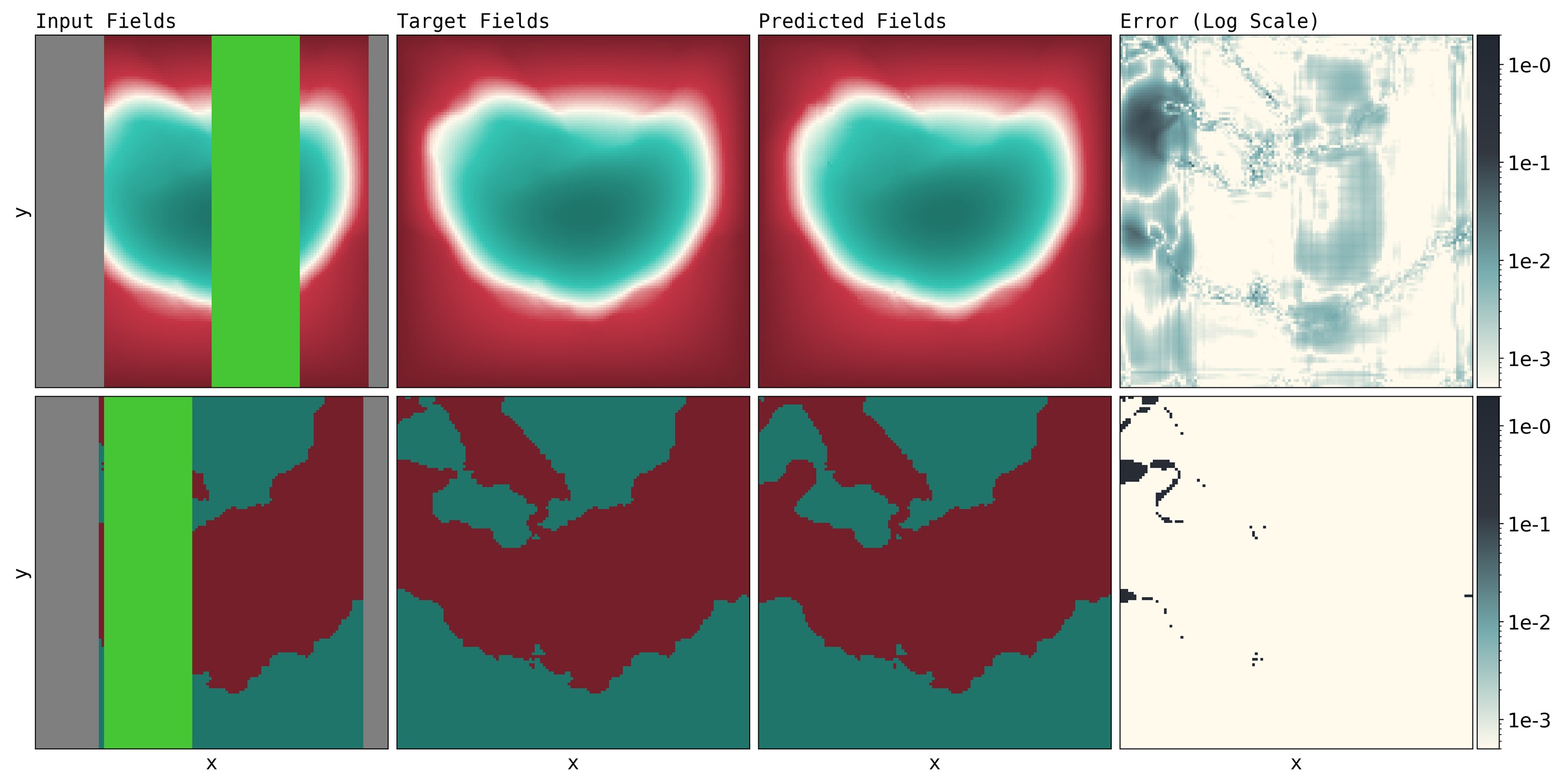}
    \caption{\textbf{Example of Window Measurements (\setlength{\fboxsep}{0pt}\colorbox{darcyflow!50}{Darcy Flow}).} Each subplot shows an Ambient Flow reconstruction from partial observations (75\% contiguous window). \textbf{Left:} No additional mask (naive training). \textbf{Right:} One additional mask.\textbf{Within Each Subplot:} \textbf{Top:} Solution field. \textbf{Bottom:} Coefficient Field. \textbf{Left to Right:} Input Fields, Target Fields, Reconstructed Fields, Error (Log Scale). Within each input panel, gray pixels represent unobserved locations and green pixels represent masked measurements withheld from the model.}
\end{figure}

\begin{figure}[H]
    \centering
    \includegraphics[width=0.56\linewidth]{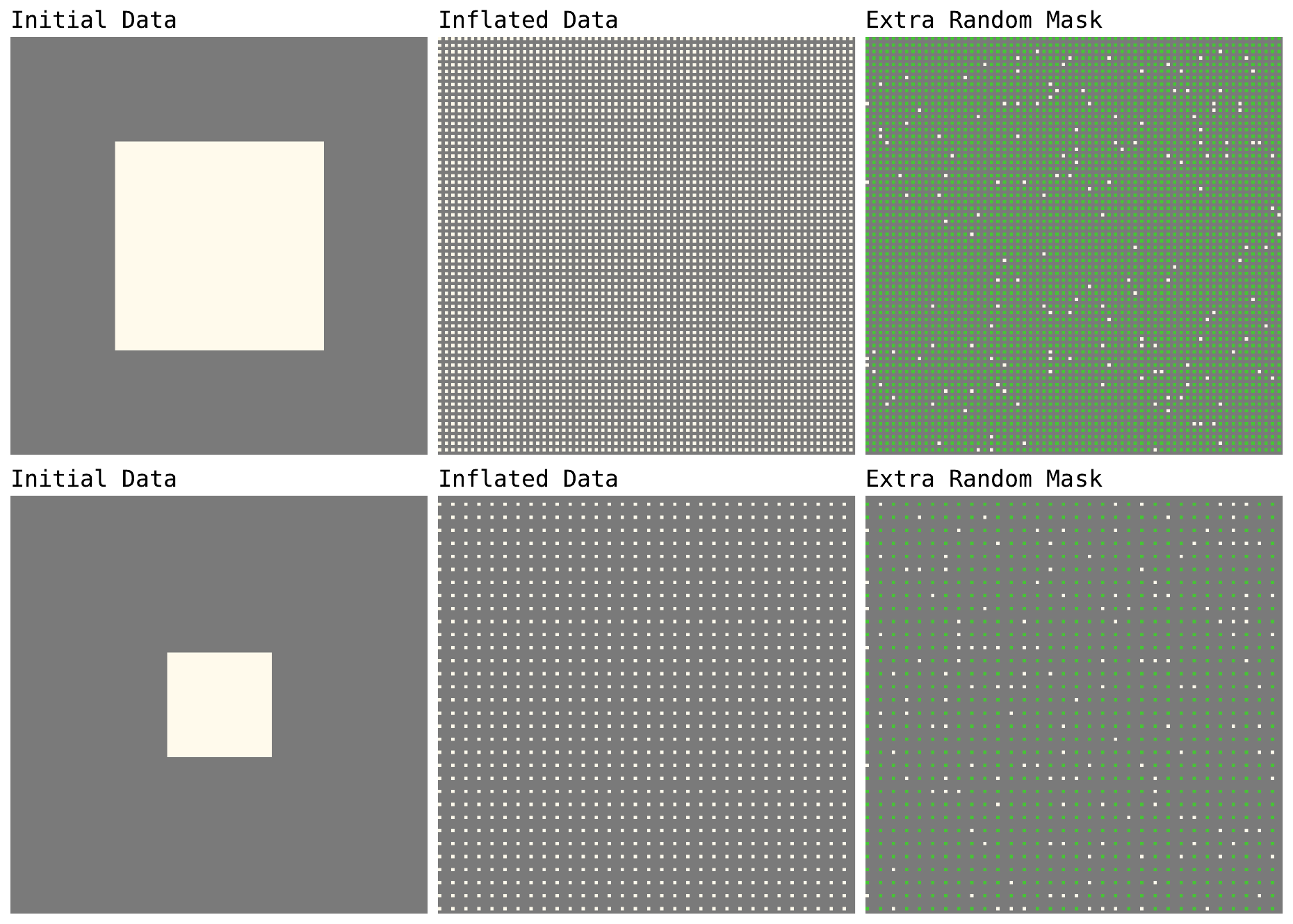}
    \caption{\textbf{Low-Resolution Observations as Partial Measurements for Super-Resolution.} A low-resolution measurement can be interpreted as a partially observed high-resolution field by inflating (upsampling) the observed values onto a higher-resolution (possibly shifted) grid. However, the inflated observations lie on a regular checkerboard-like lattice; if supervised only on these fixed locations, a model can learn to be accurate on the lattice while predicting unreliably elsewhere. We therefore apply an additional random mask over the inflated grid. Crucially, this mask must be sufficiently strong to hide the lattice structure, so the supervised locations no longer reveal a fixed checkerboard pattern and the model is encouraged to produce coherent high-resolution predictions across the domain. Ambient Physics can then learn a generative prior over high-resolution fields directly from low-resolution observations and perform super-resolution (see \autoref{tab:superres}).}
    \label{fig:placeholder}
\end{figure}

\begin{table}[H]
    \centering
    \resizebox{\textwidth}{!}{%
    \begin{tabular}{ccccccccccc}
        \toprule
        \multirow{2}{*}{Initial Resolution} & \multirow{2}{*}{Extra Masked}  & \multirow{2}{*}{Total Unobserved} & \multicolumn{2}{c}{Darcy Flow} & \multicolumn{2}{c}{Helmholtz} & \multicolumn{2}{c}{Navier-Stokes} & \multicolumn{2}{c}{Poisson} \\
        \cline{4-5}\cline{6-7}\cline{8-9}\cline{10-11}
         &  &  & Coefficients & Solutions & Coefficients & Solutions & Coefficients & Solutions & Coefficients & Solutions \\
        \midrule
        \multirow{3}{*}{\(64\times64\)} & 22.00\% & 97.00\% & 3.57\% & 2.20\% & 9.55\% & 19.34\% & 8.23\% & 2.28\% & 10.63\% & 37.20\% \\
         & 23.00\% & 98.00\% & \textbf{1.55\%} & 1.57\% & \textbf{8.67\%} & 6.04\% & \textbf{7.40\%} & \textbf{1.71\%} & \textbf{9.77\%} & 3.65\% \\
         & 24.00\% & 99.00\% & 2.26\% & \textbf{1.13\%} & 10.02\% & \textbf{1.60\%} & 9.32\% & 2.48\% & 11.39\% & \textbf{1.47\%} \\
         & 24.50\% & 99.50\% & 3.43\% & 1.36\% & 14.29\% & 3.06\% & 14.82\% & 6.28\% & 14.85\% & 1.59\% \\
        \midrule
        \multirow{3}{*}{\(32\times32\)} & 3.25\% & 97.00\% & 1.72\% & 5.31\% & 24.74\% & 16.22\% & 14.41\% & 5.40\% & 27.33\% & 23.96\% \\
         & 4.25\% & 98.00\% & 3.05\% & 3.17\% & 14.25\% & 6.00\% & \textbf{8.77\%} & 1.50\% & 14.12\% & 5.83\% \\
         & 5.25\% & 99.00\% & \textbf{2.39\%} & \textbf{1.82\%} & \textbf{10.82\%} & 2.34\% & 9.05\% & \textbf{1.46\%} & \textbf{11.27\%} & 2.29\% \\
         & 5.75\% & 99.50\% & 3.48\% & 1.91\% & 13.73\% & \textbf{1.92\%} & 14.38\% & 2.51\% & 13.89\% & \textbf{2.05\%} \\
        \bottomrule
    \end{tabular}
    }
    \caption{\textbf{Super-resolution ablation on additional random masking.} Relative $L_2$ error (\%) for coefficient/solution reconstruction after inflating low-resolution observations and applying an extra random mask. Intermediate masking (about $98\%$--$99\%$ total unobserved) performs best, consistent with the need to obscure the checkerboard observation lattice.}
    \label{tab:superres}
\end{table}

\end{document}